\documentclass{article}

\usepackage{booktabs} % For formal tables
\usepackage{multirow}
\usepackage{amsmath}
\usepackage{makecell}
\usepackage{subcaption}
\usepackage{authblk}
\usepackage{graphicx}
\usepackage{url}
\usepackage{tabularx}
\usepackage{ragged2e}

\usepackage[ruled]{algorithm2e} % For algorithms

\SetAlFnt{\small}
\SetAlCapFnt{\small}
\SetAlCapNameFnt{\small}
\SetAlCapHSkip{0pt}
\IncMargin{-\parindent}

\newcolumntype{C}{>{\Centering\arraybackslash}X}
\begin{document}
% Title portion
\title{TBD: Benchmarking and Analyzing Deep Neural Network Training}
\author[1]{\normalsize Hongyu Zhu}
\author[1]{Mohamed Akrout}
\author[1]{Bojian Zheng}
\author[1]{Andrew Pelegris}
\author[2]{\\Amar Phanishayee}
\author[1]{Bianca Schroeder}
\author[1]{Gennady Pekhimenko}
\affil[1]{University of Toronto}
\affil[2]{Microsoft Research}


%\keywords{GPU computation, deep learning, profiling}

% DO NOT use this command unless you want to change
% the default behavior
% \authorsaddresses{Authors' addresses: G.~Zhou, Computer Science
%   Department, College of William and Mary, 104 Jameson Rd,
%   Williamsburg, PA 23185, US, \path{gzhou@wm.edu}; V.~B\'eranger,
%   Inria Paris-Rocquencourt, Rocquencourt, France; A.~Patel, Rajiv
%   Gandhi University, Rono-Hills, Doimukh, Arunachal Pradesh, India;
%   H.~Chan, Tsinghua University, 30 Shuangqing Rd, Haidian Qu, Beijing
%   Shi, China; T.~Yan, Eaton Innovation Center, Prague, Czech Republic;
%   T.~He, C.~Huang, J.~A.~Stankovic University of Virginia, School of
%   Engineering Charlottesville, VA 22903, USA; T. F. Abdelzaher,
%   (Current address) NASA Ames Research Center, Moffett Field,
%   California 94035.}
\maketitle

\begin{abstract}
The recent popularity of deep neural networks (DNNs) has generated a lot of research interest in performing DNN-related computation efficiently.
However, the primary focus is usually very narrow and limited to (i) inference -- i.e. how to efficiently execute already trained models
and (ii) image classification networks as the primary benchmark for evaluation.

Our primary goal in this work is to break this myopic view by (i) proposing a
new benchmark for DNN \textit{training}, called
\texttt{TBD}\footnote{\textbf{TBD} is short for \textbf{T}raining
\textbf{B}enchmark for \textbf{D}NNs}, that uses a representative set of DNN
models that cover a wide range of machine learning applications: image
classification, machine translation, speech recognition, object detection, adversarial networks,
reinforcement learning, and (ii) by performing an extensive performance
analysis of training these different applications on three major deep learning
frameworks (TensorFlow, MXNet, CNTK) across
different hardware configurations (single-GPU, multi-GPU, and multi-machine).
\texttt{TBD} currently covers six major application domains and eight different
state-of-the-art models.
We present a new toolchain for performance analysis for these models 
that combines the targeted usage of existing performance
analysis tools, careful selection of new and existing metrics and
methodologies to analyze the results, and utilization of
domain specific characteristics of DNN training. We also build a new set of tools
for memory profiling in all three major frameworks; much needed tools that
can finally shed some light on precisely how much memory is consumed by
different data structures (weights, activations, gradients, workspace) in
DNN training. By using our tools and methodologies, we make several important observations and
recommendations on where the future research and optimization of DNN training should be
focused.
\end{abstract}

\section{Introduction}
The availability of large datasets and powerful computing resources has enabled
a new type of artificial neural networks---deep neural networks
(DNNs~\cite{hinton2006fast, bengio2007greedy})---to solve hard problems such as
image classification, machine translation, and speech
processing~\cite{krizhevsky2012imagenet,he2016deep,amodei2016deep,hieber2017sockeye,wu2016google,vaswani2017attention}.
While this recent success of DNN-based learning algorithms has naturally
attracted a lot of attention, the primary focus of researchers especially in
the systems and computer architecture communities is usually on
\emph{inference}---i.e. how to efficiently execute already trained models, and
\emph{image classification} (which is used as the primary benchmark to evaluate
DNN computation efficiency).

While inference is arguably an important problem, we observe that efficiently
\emph{training} new models is becoming equally important as machine learning is
applied to an ever growing number of domains, e.g., speech
recognition~\cite{amodei2016deep,xiong2017microsoft}, machine
translation~\cite{bahdanau2014neural,luong2015effective,sutskever2014sequence},
automobile industry~\cite{bojarski2016end,huval2015empirical}, and
recommendation systems~\cite{covington2016deep,he2017neural}.
%% Moreover, these new applications employ new types of layers that differ from
%those used for image classification % (e.g., machine translation models
%usually use recurrent neural networks, RNN layers, % rather than convolutional
%layers~\cite{lecun1998gradient}).  % These new layer types have very different
%compute and % memory bandwidth characteristics, and frameworks optimized for %
%convolutions might not perform well on these new layers (e.g., RNNs).
But researchers currently lack comprehensive benchmarks and profiling tools for DNN training.  In
this paper, we present a new benchmark for DNN training, called \texttt{TBD},
that uses a representative set of DNN models covering a broad range of machine
learning applications: image classification, machine translation, speech
recognition, adversarial networks, reinforcement learning. \texttt{TBD}
also incorporates an analysis toolchain for performing detailed resource and performance profiling of these models,
including the first publicly available tool for profiling memory usage on major
DNN frameworks.
Using \texttt{TBD} we perform a detailed performance analysis on how these different
applications behave on three DNN training frameworks
(TensorFlow~\cite{abadi2016tensorflow}, MXNet~\cite{chen2015mxnet},
CNTK~\cite{yu2014introduction}) across different hardware configurations
(single-GPU, multi-GPU, and multi-machine) gaining some interesting insights.

\begin{table*}
	\centering
	\begin{tabularx}{\textwidth}{c|C|C}
	%\hline
	 & Image Classification Only & Broader (include non-CNN workloads) \\
\hline Training &
\cite{chilimbi2014project}\cite{de2017understanding}\cite{ding2017c}\cite{hsieh2017gaia}\cite{judd2016stripes}\cite{kim2016neurocube}\cite{rhu2016vdnn}\cite{song2017pipelayer}\cite{venkataramani2017scaledeep}
&
\cite{abadi2016tensorflow}\cite{bojnordi2016memristive}\cite{ji2016neutrams}\cite{li2014scaling}\cite{park2017scale}\cite{pei2017deepxplore}\cite{xiao2017tux2}
\\ \hline

	Inference &
\cite{albericio2017bit}\cite{albericio2016cnvlutin}\cite{alwani2016fused}\cite{chen2016eyeriss}\cite{chi2016prime}\cite{ding2017c}\cite{du2015shidiannao}\cite{gao2017tetris}\cite{judd2016stripes} \cite{likamwa2016redeye}\cite{lu2017flexflow}\cite{parashar2017scnn}\cite{ren2017sc}\cite{shafiee2016isaac}\cite{sharma2016high}\cite{shen2016maximizing}\cite{song2017pipelayer} \cite{yu2017scalpel}\cite{zhang2016cambricon}
&
\cite{abadi2016tensorflow}\cite{du2015neuromorphic}\cite{han2016eie}\cite{hauswald2015djinn}\cite{jouppi2017datacenter}\cite{park2017scale}
\\
	%\hline
    \end{tabularx} 
\vspace{0.1cm}
\caption{The table above shows a categorization of major
computer architecture and systems conference papers (SOSP, OSDI, NSDI, MICRO,
ISCA, HPCA, ASPLOS) since 2014. These papers are grouped by their focus along
two dimensions: Training versus Inference and Algorithmic Breadth. There are
more papers which optimize inference over training (25 vs. 16, 4 papers aim
for both training and inference).  Similarly more papers use image
classification as the \emph{only} application for evaluation (26 vs. 11).}
\label{Related_work_summary} \end{table*}

\texttt{TBD}'s benchmark suite and
analysis toolchain is driven by the motivation to address three main
challenges:

\noindent\textbf{1. Training differs significantly from inference.} The
algorithmic differences between training and inference lead to many differences
in requirements for the underlying systems and hardware architecture.  First,
\emph{backward pass} and \emph{weight updates}, operations unique to training,
need to save/stash a large number of intermediate results in GPU memory, e.g.,
outputs of the inner layers called \emph{feature maps} or
activations~\cite{rhu2016vdnn}.  This puts significant pressure on the memory
subsystem of modern DNN accelerators (usually GPUs) -- in some cases the model
might need tens of gigabytes of main memory~\cite{rhu2016vdnn}.  In contrast,
the memory footprint of inference is significantly smaller, in the order of
tens of megabytes~\cite{han2015deep}, and the major memory consumers are model
weights rather than feature maps.  Second, training usually proceeds in waves
of \emph{mini-batches}, a set of inputs grouped and processed in
parallel~\cite{goyal2017accurate,you2017imagenet}.  Mini-batching helps in
avoiding both overfitting and under utilization of GPU's compute parallelism.
Thus, throughput is the primary performance metric of concern in training.
Compared to training, inference is computationally less taxing and is latency
sensitive.

\noindent\textbf{2. Workload diversity.} Deep learning has achieved
state-of-the-art results in a very broad range of application domains.
Yet most existing evaluations of DNN performance remain narrowly focused
on just image classification as their benchmark application, and
convolutional neural networks (CNNs) remain the most widely-used models for
systems/architecture researchers (Table~\ref{Related_work_summary}).  As a
result, many important non-CNN models have not received much attention, with only a handful of
papers evaluating non-CNNs such as recurrent neural
networks~\cite{abadi2016tensorflow,jouppi2017datacenter,hauswald2015djinn}.
Papers that cover unsupervised learning or deep reinforcement learning are
extremely rare.  The computational characteristics of image classification
models are very different from these networks, thus motivating a need for a
broader benchmark suite for DNN training.  Furthermore, given the rapid pace of
innovation across the realms of algorithms, systems, and hardware related to
deep learning, such benchmarks risk being quickly obsolete if they don't change
with time.

\noindent\textbf{3. Identifying bottlenecks.} It is not obvious
which hardware resource is the critical bottleneck that typically limits training throughput,
as there are multiple plausible candidates.
Typical convolutional neural networks (CNNs) are usually computationally intensive, making
\emph{computation} one of the primary bottlenecks in single GPU training.
Efficiently using modern GPUs (or other hardware accelerators) requires
training with large mini-batch sizes. Unfortunately, as we will show later in
Section~\ref{sec:performance}, for some workloads (e.g., RNNs, LSTMs) this
requirement can not be satisfied due to capacity limitations of GPU \emph{main
memory} (usually 8--16GBs).  Training DNNs in a distributed environment with
multiple GPUs and multiple machines, brings with it yet another group of
potential bottlenecks, \emph{network and interconnect bandwidths}, as training
requires fast communication between many CPUs and GPUs (see
Section~\ref{sec:performance-distributed}).
Even for a specific model, implementation and hardware setup pinpointing whether
performance is bounded by computation, memory, or communication is not
easy due to limitations of existing profiling tools. Commonly used tools (e.g., vTune~\cite{reinders2005vtune},
nvprof~\cite{nvprof}, etc.) have no domain-specific knowledge about the
algorithm logic, can only capture some low-level information within their own
scopes, and usually cannot perform analysis on full application executions with
huge working set sizes.  Furthermore, no tools for memory profiling are
currently available for any of the major DNN frameworks.

Our paper makes the following contributions.
\begin{itemize}%[topsep=0pt,wide=0pt,itemsep=0pt]
\item \textbf{\texttt{TBD}, a new benchmark suite.} We
create a new benchmark suite for DNN training that currently covers \emph{six}
major application domains and \emph{eight} different state-of-the-art models.
The applications in this suite are selected based on extensive conversations
with ML developers and users from both industry and academia. For all
application domains we select recent models capable of delivering
state-of-the-art results. We will open-source our benchmarks suite later this
year and intend to continually expand it with new applications and models based
on feedback and support from the community.
\item \textbf{Tools to enable end-to-end performance analysis.} We develop a
toolchain for end-to-end analysis of DNN training.  To perform such analysis,
we perform piecewise profiling by targeting specific parts of training using
existing performance analysis tools, and then merge and analyze them using
domain-specific knowledge of DNN training.  As part of the toolchain we also
built new memory profiling tools for the three major DNN frameworks we
considered: TensorFlow~\cite{abadi2016tensorflow}, MXNet~\cite{chen2015mxnet}, and
CNTK~\cite{yu2014introduction}.  Our memory profilers can pinpoint how much
memory is consumed by different data structures during training (weights,
activations, gradients, workspace etc.), thus enabling developer to make easy
data-driven decisions for memory optimizations.

\item \textbf{Findings and Recommendations.} Using our benchmark suite and
analysis tools, we make several important observations and recommendations on
where the future research and optimization of DNNs should be focused.
We include a few examples here:
(1) We find that the training of state-of-the-art RNN models is not as
efficient as for image classification models, because GPU utilization
for RNN models is 2--3$\times$ lower than for most other benchmark models.
(2) We find that GPU memory is often not utilized efficiently, the strategy of exhausting GPU memory capacity with large mini-batch provides limited benefits for a wide range of models.
%CPU cores are poorly utilized in DNN training with an averageutilization for most benchmark models being less than 15\%.
(3) We also find that the feature
maps, the output of the DNN intermediate layers, consume 70--90\%
of the total memory footprint for all our benchmark
models. This is a significant contrast to inference, where footprint is dominated
by the weights.
These observations suggest several interesting research directions, including
efficient RNN layer implementations and memory footprint reduction optimizations
with the focus on feature maps.
%to improve memory consumption of DNN training should
%focus on feature maps with high priority.

\end{itemize}

The \texttt{TBD} benchmark suite and the accompanying measurement toolchain,
and insights derived from them will aid researchers and practitioners
in computer systems, computer architecture, and machine learning to determine where to
target their optimizations efforts within each level in the DNN training stack:
(i) applications and their corresponding models,
(ii) currently used libraries (e.g., cuDNN), and (iii) hardware that is used to
train these models. We also hope that our paper will instigate additional follow-up work
within the Sigmetrics community aimed at providing DNN research with a more rigorous foundation rooted
in measurements and benchmarking.

In the rest of this paper, we first provide some background on DNN training, both single-GPU
and distributed training (with multiple GPUs and multiple machines) in Section~2. We then
present our methodology, explaining which DNN models we selected to be included in our  benchmark suite and why,
and describing our measurement framework and tools to analyze the performance of these models (Section~\ref{sec:method}).
We then use our benchmark and measurement framework to derive observations and insights about these models' performance and resource characteristics
in Section~\ref{sec:results}.
We conclude the paper with a description of related work in Section~\ref{sec:related} and a summary of our
work in Section~\ref{sec:summary}.

\section{Background}

\subsection{Deep Neural Network Training and Inference}
\label{background:training}

\begin{figure}
    \centering
    \includegraphics[width=\textwidth]{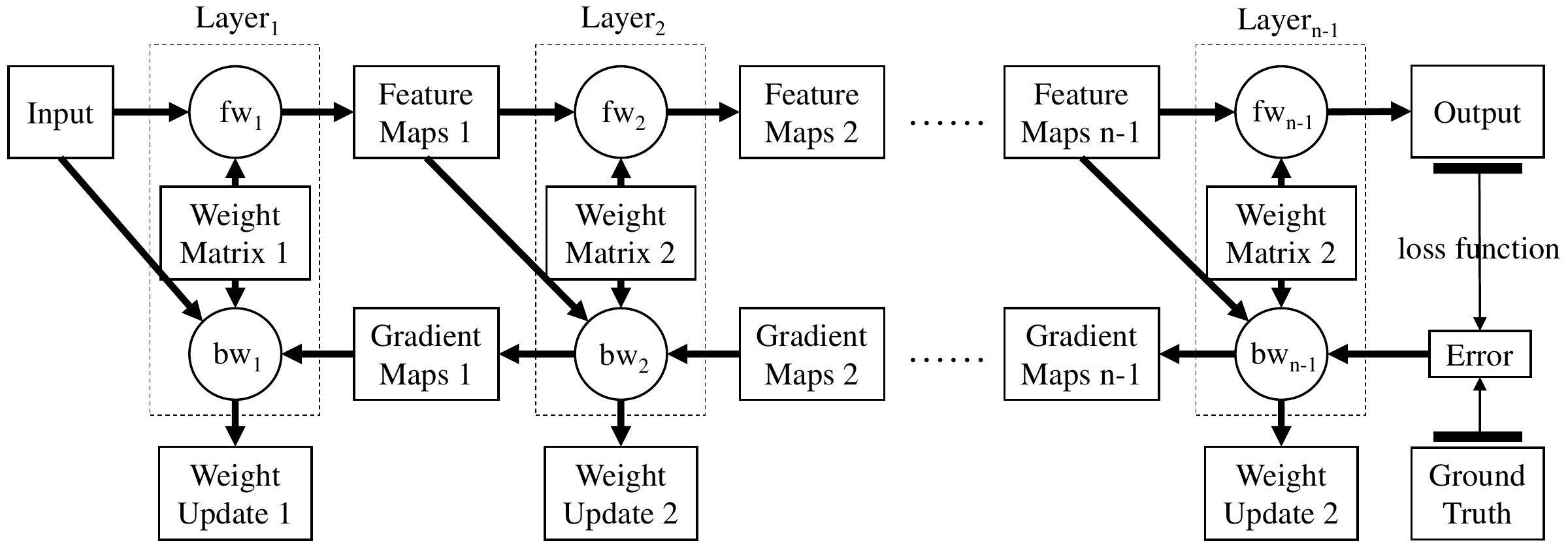}
    \caption{Feed-forward and Back-propagation}
    \label{fig:FFBP}
\end{figure}

A neural network can be seen as a function which takes data samples as inputs,
and outputs certain properties of the input samples (Figure~\ref{fig:FFBP}).
Neural networks are made up of a series of layers of neurons.
Neurons across layers are connected, and layers can be of different types such as
fully-connected, convolutional, pooling, recurrent, etc.
While the edges connecting neurons across layers are weighted, each layer can be considered to have its own set of \textit{weights}.
Each layer applies a mathematical transformation to its input.
For example, a fully-connected layer multiplies intermediate results computed by its preceding/upstream layer (input) by its weight matrix,
adds a bias vector, and applies a non-linear function (e.g., sigmoid) to the result;
this result is then used as the input to its following/downstream layer.
The intermediate results generated by each layer are often called \textit{feature maps}.
Feature maps closer to the output layer generally represent higher order features of the data samples.
This entire layer-wise computation procedure from input data samples to output is called \textit{inference}.

A neural network needs to be trained before it can detect meaningful
properties corresponding to input data samples. The goal of \textit{training} is to
find proper weight values for each layer so that the network as a whole can produce desired
outputs. Training a neural network is an iterative algorithm, where each
iteration consists of a \textit{forward} pass and a \textit{backward} pass. The
forward pass is computationally similar to inference.
For a network that is not fully trained, the inference results might be very different from ground truths labels.
A \textit{loss function} measures the difference between the
predicted value in the forward pass and the ground truth. Similar to the forward pass, computation in the backward pass
also proceeds layer-wise, but in an opposite direction. Each layer uses errors from its
downstream layers and feature maps generated in the forward pass to compute not only
errors to its upstream layers according to the chain rule~\cite{rumelhart1986learning} but also gradients
of its internal weights. The gradients are then used for updating the weights.
This process is known as the \textit{gradient descent} algorithm, used widely to train neural networks.

As modern training dataset are extremely large, it is
expensive to use the entire set of the training data in each iteration.
Instead, a training iteration randomly samples a \textit{mini-batch} from the training
data, and uses this mini-batch as input. The randomly sampled mini-batch is a
stochastic approximation to the full batch. This algorithm is called
\emph{stochastic gradient descent} (SGD)~\cite{lecun1998gradient}. The size of
the mini-batch is a crucial parameter which greatly affects both the training performance
and the memory footprint.

\subsection{GPUs and Distributed Training via Data Parallelism}

While the theoretical foundations of neural networks have a long history, it is only relatively recently the people realized the power of deep neural networks. This is because to fully
train a neural network on a CPU is extremely time-consuming~\cite{shi2016benchmarking}. The first successful deep neural network~\cite{krizhevsky2012imagenet}
that beat all competitors in image classification task in 2012, was trained using two GTX 580 GPUs~\cite{gtx580} in six days
instead of months of training on CPUs. One factor that greatly limits the size of the network is the amount of
tolerable training time. Since then, almost all advanced deep learning
models are trained using either GPUs or some other type of hardware accelerators~\cite{jouppi2017datacenter,guan2017fpga}.

One way to further speed up
the neural network training is to parallelize the training procedure and deploy
the parallelized procedure in a distributed environment. A simple and
effective way to do so is called \textit{data parallelism}~\cite{dean2012large}. It lets each worker train a single network replica.
In an iteration, the input mini-batch is partitioned into $n$ subsets, one for each worker.
Each worker then takes this subset of the mini-batch, performs the forward and backward passes respectively,
and exchanges weight updates with all other workers.

Another way to parallelize the computation is by using \textit{model parallelism}~\cite{wang2014minerva},
an approach used when the model's working set is too large to fit in the memory of a single worker.
Model parallel training splits the workload of training a complete model across the workers;
each worker trains only a part of the network.
This approach requires careful workload partitioning to achieve even load-balancing and low communication overheads.
The quality of workload partitioning in model parallelism depends highly on DNN architecture.
Unlike model parallelism, data parallelism is simpler to get right and is the predominant method of parallel training.
In this paper we limit our attention to data parallel distributed training.

\subsection{DNN Frameworks and Low-level Libraries}

DNN frameworks and low-level libraries are designed to simplify the life of
ML programmers and to help them to efficiently utilize existing complex hardware.
A DNN framework (e.g., TensorFlow or MXNet) usually provides users
with compact numpy/matlab-like matrix APIs to define the computation logic, or
a configuration format, that helps ML programmers to specify the topology
of their DNNs layer-by-layer.
The programming APIs are usually bounded with the popular high-level
programming languages such as Python, Scala, and R. A framework transforms the user
program or configuration file into an internal intermediate representation
(e.g., dataflow graph
representation~\cite{abadi2016tensorflow,chen2015mxnet,bergstra2010theano}),
which is a basis for backend execution including data transfers, memory
allocations, and low-level CPU function calls or GPU kernel\footnote{A GPU kernel is a
routine that is executed by an array of CUDA threads on GPU
cores.} invocations. The invoked low-level functions are usually provided by
libraries such as
cuDNN~\cite{chetlur2014cudnn}, cuBLAS~\cite{cublas}, MKL~\cite{wang2014intel}, and Eigen~\cite{eigen}.
These libraries provide efficient implementations of basic vector and
multi-dimension matrix operations (some operations are NN-specific such as
convolutions or poolings) in C/C++ (for CPU) or CUDA (for GPU). The performance of
these libraries will directly affect the overall training performance.

\section{Methodology} \label{sec:method}

\renewcommand{\thefootnote}{\alph{footnote}}

\begin{table*}
    \begin{tabularx}{\textwidth}{ |p{1.6cm}|p{1.9cm}|p{1.1cm}|p{1.3cm}|p{1.7cm}|p{1.9cm}| }
    \hline
    Application & Model & Number of Layers & Dominant Layer & Frameworks & Dataset \\ \hline
    Image classification & ResNet-50 \cite{krizhevsky2012imagenet} & 50 (152 max) & CONV & TensorFlow, MXNet, CNTK & ImageNet1K \cite{ILSVRC15} \\
     & Inception-v3 \cite{szegedy2016rethinking} & 42 &  &  & \\ \hline
    Machine translation & Seq2Seq \cite{sutskever2014sequence} & 5 & LSTM & TensorFlow, MXNet & IWSLT15~\cite{cettolo2015iwslt} \\
     & Transformer \cite{vaswani2017attention} & 12 & Attention & TensorFlow & \\ \hline
    Object detection & Faster R-CNN \cite{ren2015faster} & 101\footnotemark[1] & CONV & TensorFlow, MXNet & Pascal VOC 2007 \cite{everingham2010pascal} \\ \hline
    Speech recognition & Deep Speech 2 \cite{amodei2016deep} & 9\footnotemark[2] & RNN & MXNet & LibriSpeech \cite{panayotov2015librispeech} \\ \hline
    Adversarial learning & WGAN \cite{gulrajani2017improved} & 14+14\footnotemark[3] & CONV & TensorFlow & Downsampled ImageNet \cite{chrabaszcz2017downsampled} \\ \hline
    Deep reinforcement learning & A3C \cite{mnih2016asynchronous} & 4 & CONV & MXNet & Atari 2600 \\ \hline
    \end{tabularx}
    \caption{Overview of Benchmarks, including the models and datasets used, number and major layer types, and frameworks with available implementations.}
    \label{table:benchmark}
\end{table*}

\begin{table*}
    \begin{tabularx}{\textwidth}{ |C|C|C|C| }
    \hline
    Dataset & Number of Samples & Size & Special \\ \hline
    ImageNet1K & 1.2million & 3x256x256 per image & N/A \\ \hline
    IWSLT15 & 133k & 20-30 words long per sentence & vocabulary size of 17188 \\ \hline
    Pascal VOC 2007 & 5011\footnotemark[4] & around 500x350 & 12608 annotated objects \\ \hline
    LibriSpeech & 280k & 1000 hours\footnotemark[5] & N/A \\ \hline
    Downsampled ImageNet & 1.2million & 3x64x64 per image & N/A \\ \hline
    Atari 2600 & N/A & 4x84x84 per image & N/A \\ \hline
    \end{tabularx}
    \caption{Training Datasets}
    \label{table:dataset}
\end{table*}

\subsection{Application and Model Selection}

Based on a careful survey of existing literature and in-depth discussions with
machine learning researchers and industry developers at several institutions
(Google, Microsoft, and Nvidia) we identified a diverse set of interesting
application domains, where deep learning has been emerging as the most promising solution:
{image classification, object detection, machine translation,
speech recognition, generative adversarial nets, and deep reinforcement
learning}. While this is the set of applications we will include with the first release
of our open-source benchmark suite, we expect to continuously expand it based on
community feedback and contributions and to keep up with advances of deep learning
in new application domains.

Table~\ref{table:benchmark} summarizes the models and datasets we chose to
represent the different application domains.  When selecting the models, our
emphasis has been on picking the most recent models capable of producing
state-of-the-art results (rather than for example classical models of
historical significance). The reasons are that these models are the most
likely to serve as building blocks or inspiration for the development of future
algorithms and also often use new types of layers, with new resource profiles,
that are not present in older models.  Moreover, the design of models is often
constrained by hardware limitations, which will have changed since the
introduction of older models.

%Table \ref{table:benchmark} shows our current model collection. Each application area has at least one state-of-the-art
%model and its corresponding dataset. Two areas (machine translation and image classification)
%have two models mostly because these models have either extremely different DNN architectures or types of layers
%used that leads to very interesting tradeoffs in their execution. For example, state-of-the-art models for machine
%translation were usually based on recurrent neural networks (RNNs) using LSTM cells. But recently (June 2017)
%a new model, called \emph{Transformer}~\cite{vaswani2017attention}, was released that uses \emph{attention} layers instead of RNNs
%, and is competitive with the best prior RNN-based models. As we will show later in Section~\ref{sec:performance}, Transofer model
%has very different performance characteristics from those of RNNs.

%One note is that implementing the benchmarks on frameworks is beyond the scope of this paper. As the algorithms are quickly evolving, there is sometimes not enough time for implementations to be available on all frameworks. In this paper we use the existing open-source implementations provided by either the framework developers on the official github repository, or third-party implementations only when official versions are not available.

\subsubsection{Image Classification}
Image classification is the archetypal deep learning application, as this was
the first domain where a deep neural network
(AlexNet~\cite{krizhevsky2012imagenet}) proved to be a watershed, beating all
prior traditional methods.  In our work, we use two very recent models,
Inception-v3~\cite{szegedy2016rethinking} and Resnet~\cite{he2016deep}, which
follow a structure similar to AlexNet's CNN model, but improve accuracy through
novel algorithm techniques that enable extremely deep networks.

%The basic structure of AlexNet is a stack of convolutional and pooling layers, followed by a three connected layers (if the output layer counts).
%This structure is adopted by subsequent successful CNN models (VGG~\cite{simonyan2014very}, GoogleNet~\cite{szegedy2015going}, ResNet~\cite{he2016deep}) with deeper convolutional stacks.

%In this paper we use Inception-v3~\cite{szegedy2016rethinking} and Resnet~\cite{he2016deep} for our image classification benchmark models because of their contributions and current popularity. These two models proposed novel algorithm techniques to allow the network becoming extremely deep.
%

\subsubsection{Object Detection}

Object detection applications, such as face detection, are another popular deep
learning application and can be thought of as an extension of image
classification, where an algorithm usually first breaks down an image into regions of
interest and then applies image classification to each region.  We choose to
include Faster R-CNN~\cite{ren2015faster}, which achieves state-of-the-art
results on the Pascal VOC datasets~\cite{everingham2010pascal}.
A training iteration consists of the
forward and backward passes of two networks (one for identifying regions and
one for classification), weight sharing and local fine-tuning.  The convolution
stack in a Faster R-CNN network is usually a standard image classification
network, in our work a 101-layer ResNet.

In the future, we plan to add YOLO9000~\cite{redmon2016yolo9000}, a network
recently proposed for the real-time detection of objects, to our benchmark
suite.  It can perform inference faster than Faster R-CNN, however at the point
of writing its accuracy is still lagging and its implementations on the various
frameworks is not quite mature enough yet.

\subsubsection{Machine Translation}

Unlike image processing, machine translation involves the analysis of
sequential data and typically relies on RNNs using LSTM cells as its core
algorithm.  We select \textit{NMT}\cite{wu2016google} and
\textit{Sockeye}\cite{hieber2017sockeye}, developed by the \textit{TensorFlow}
and \textit{Amazon Web Service} teams, respectively, as representative
RNN-based models in this area.
 %are selected to be the representative in this area, and \textit{NMT} also
 %claimed that they could achieve comparable results with the
 %Google-Neural-Machine-Translation (GNMT) \cite{wu2016google} system.
We also include an implementation of the recently introduced~\cite{vaswani2017attention} \textit{Transformer} model,
which achieves a new state-of-the-art in
translation quality using attention layers as an alternative to recurrent layers.
%Since it is maintained by the authors of \cite{vaswani2017attention}, the
%tensor2tensor implementation, written in Tensorflow, is selected to be the
%representative for this model.

\subsubsection{Speech Recognition}

\textit{Deep Speech 2}~\cite{amodei2016deep} is an end-to-end speech recognition model
from \textit{Baidu Research}.  It is able to accurately recognize both English and
Mandarin Chinese, two very distant languages, with a unified model architecture
and shows great potential for deployment in industry. The Deep Speech 2 model
contains two convolutional layers, plus seven regular recurrent layers or Gate
Recurrent Units (GRUs), different from the RNN models in machine translation
included in our benchmark suite, which use LSTM layers.

%We use LibriSpeech~\cite{panayotov2015librispeech} as the training dataset due to its availability. This dataset is an automatic speech recognition corpus of approximately 1000 hours of English speech.

\subsubsection{Generative Adversarial Networks}

A generative adversarial network (GAN) trains two networks, one generator
network and one discriminator network. The generator is trained to generate
data samples that mimic the real samples, and the discriminator is trained to
distinguish whether a data sample is genuine or synthesized.
GANs are used, for example, to synthetically generate photographs that look at least
superficially authentic to human observers.

While GANs are powerful generative models, training a GAN suffers from instability. The
WGAN~\cite{arjovsky2017wasserstein} is a milestone as it makes great progress
towards stable training. Recently Gulrajani et al.~\cite{gulrajani2017improved}
proposes an improvement based on the WGAN to enable stable training on a wide
range of GAN architectures. We include this model into our benchmark suite as
it is one of the leading DNN algorithms in the unsupervised learning area.

\subsubsection{Deep Reinforcement Learning}

Deep neural networks are also responsible for recent advances in reinforcement
learning, which have contributed to the creation of the first artificial agents
to achieve human-level performance across challenging domains, such as the game
of Go and various classical computer games.
%directly by interacting with their simulated environments.
We include the \textit{A3C} algorithm~\cite{mnih2016asynchronous} in our benchmark
suite, as it has become one of the most popular deep reinforcement learning
techniques, surpassing the DQN training algorithms~\cite{mnih2015human}, and works
in both single and distributed machine settings. A3C relies on asynchronously
updated policy and value function networks trained in parallel over several
processing threads.

\footnotetext[1]{We use the convolution stack of ResNet-101 to be the shared convolution stack between Region Proposal Network and the detection network.}
\footnotetext[2]{The official Deep Speech 2 model has 2 convolutional layers plus 7 RNN layers. Due to memory issue, we use the default MXNet configuration which has 5 RNN layers instead.}
\footnotetext[3]{The architecture for both the generator and discriminator of WGAN is a small CNN containing 4 residual blocks.}
\footnotetext[4]{We use the train+val set of Pascal VOC 2007 dataset.}
\footnotetext[5]{The entire LibriSpeech dataset consists of 3 subsets with 100 hours, 360 hours and 500 hours respectively. By default, the MXNet implementation uses the 100-hour subset as the training dataset.}

\renewcommand{\thefootnote}{\arabic{footnote}}

\subsection{Framework Selection}

There are many open-source DNN frameworks, such as TensorFlow~\cite{abadi2016tensorflow},
Theano~\cite{bergstra2010theano}, MXNet~\cite{chen2015mxnet},
CNTK~\cite{yu2014introduction}, Caffe~\cite{jia2014caffe},
Chainer~\cite{tokui2015chainer}, Torch~\cite{collobert2011torch7},
Keras~\cite{chollet2015keras}, PyTorch~\cite{paszke2017automatic}.
Each of them applies some generic high-level optimizations (e.g.,
exploiting model parallelism using dataflow computation, overlapping computation
with communication) and some unique optimizations of their own (e.g.,
different memory managers and memory allocation strategies, specific libraries
to perform efficient computation of certain DNN layer types). Most of these
frameworks share similar code structure, and provide either declarative or imperative
high-level APIs. The computation of forward and backward passes is
performed by either existing low-level libraries (e.g., cuBLAS, cuDNN, Eigen,
MKL, etc.) or using their own implementations. For the same neural network
model trained using different frameworks, the invoked GPU kernels (normally
the major part of the computation) are usually functionally the same. This provides
us with a basis to compare, select, and analyze the efficiency of different
frameworks.

As there is not one single framework that has emerged as the dominant leader in the field
and different framework-specific design choices and optimizations might lead to different results,
we include several frameworks in our work.
In particular, we choose TensorFlow~\cite{abadi2016tensorflow},
MXNet~\cite{chen2015mxnet}, and CNTK~\cite{yu2014introduction},
as all three platforms have a large number of active users,
are actively evolving, have many of the implementations for the models we
were interested in\footnote{Note that implementing a model on a new framework from scratch
is a highly complex task beyond the scope of our work. Hence
in this paper we use the existing open-source implementations provided by either the framework
developers on the official github repository, or third-party implementations when official versions are not available.
}, and support hardware acceleration using single and multiple GPUs.
%There are two reasons why we choose more than one frameworks. First, our analysis will be helpful for a wider range of ML practitioners and researchers. Second, our results are less biased by framework-specific designs and optimizations, and therefore more algorithm-related and fundamental.
%

%One
%note regarding the hardware accelerators is that, we use GPUs for all our
%experiments, as GPUs are the most commonly used accelerators for DNN training,
%and GPU integration is available for all frameworks.

\subsection{Training Benchmark Models}

\begin{figure*} [h!]
    \centering
    \begin{subfigure}[t]{0.30\textwidth}
        \centering
        \includegraphics[width=\textwidth]{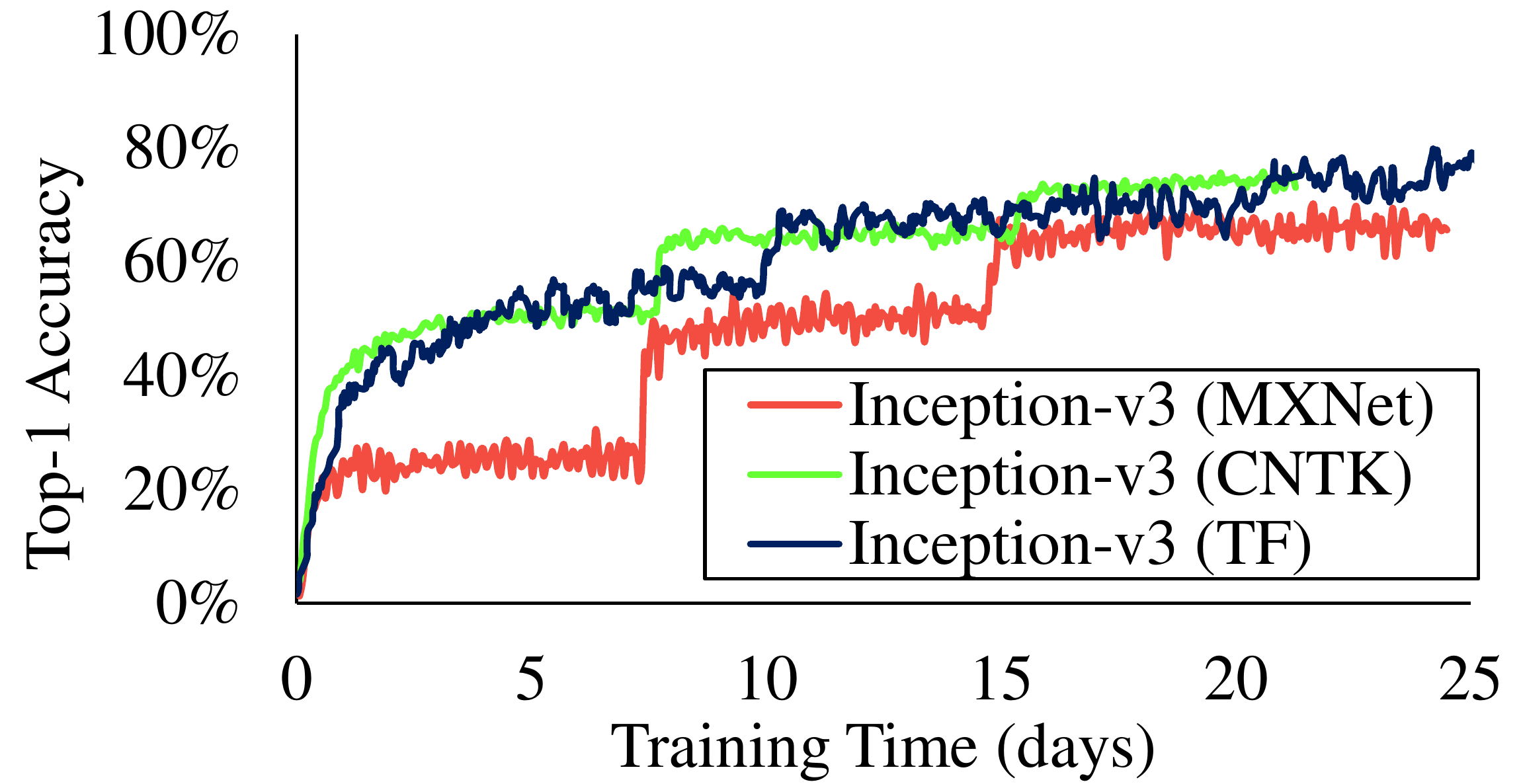}
        \caption{Inception-v3}
        \label{fig:training_inception}
    \end{subfigure}%
    \hspace{0.1cm}
    \begin{subfigure}[t]{0.30\textwidth}
        \centering
        \includegraphics[width=\textwidth]{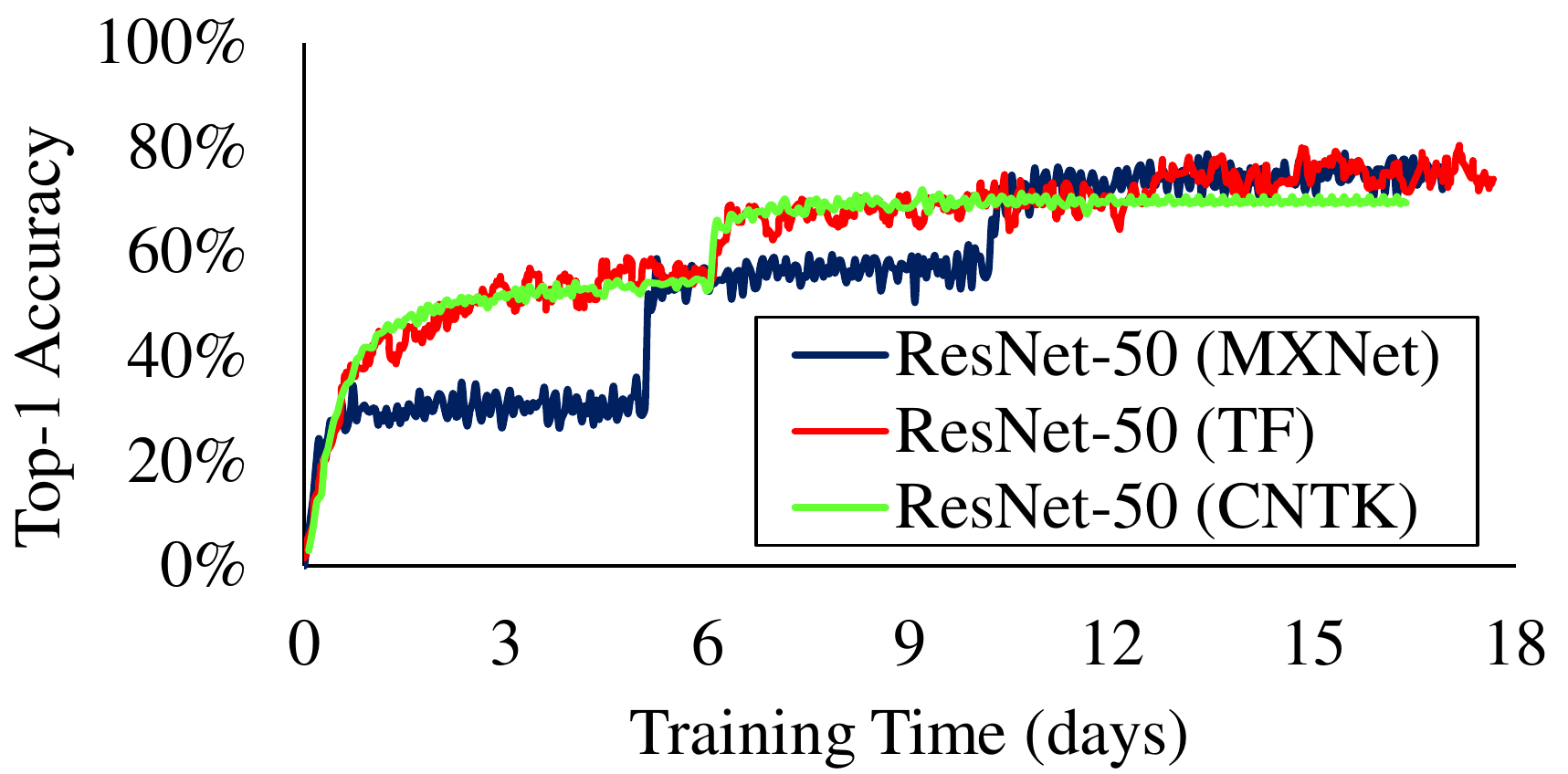}
        \caption{ResNet-50}
        \label{fig:training_resnet}
    \end{subfigure}%
    \hspace{0.1cm}
    \begin{subfigure}[t]{0.30\textwidth}
        \centering
        \includegraphics[width=\textwidth]{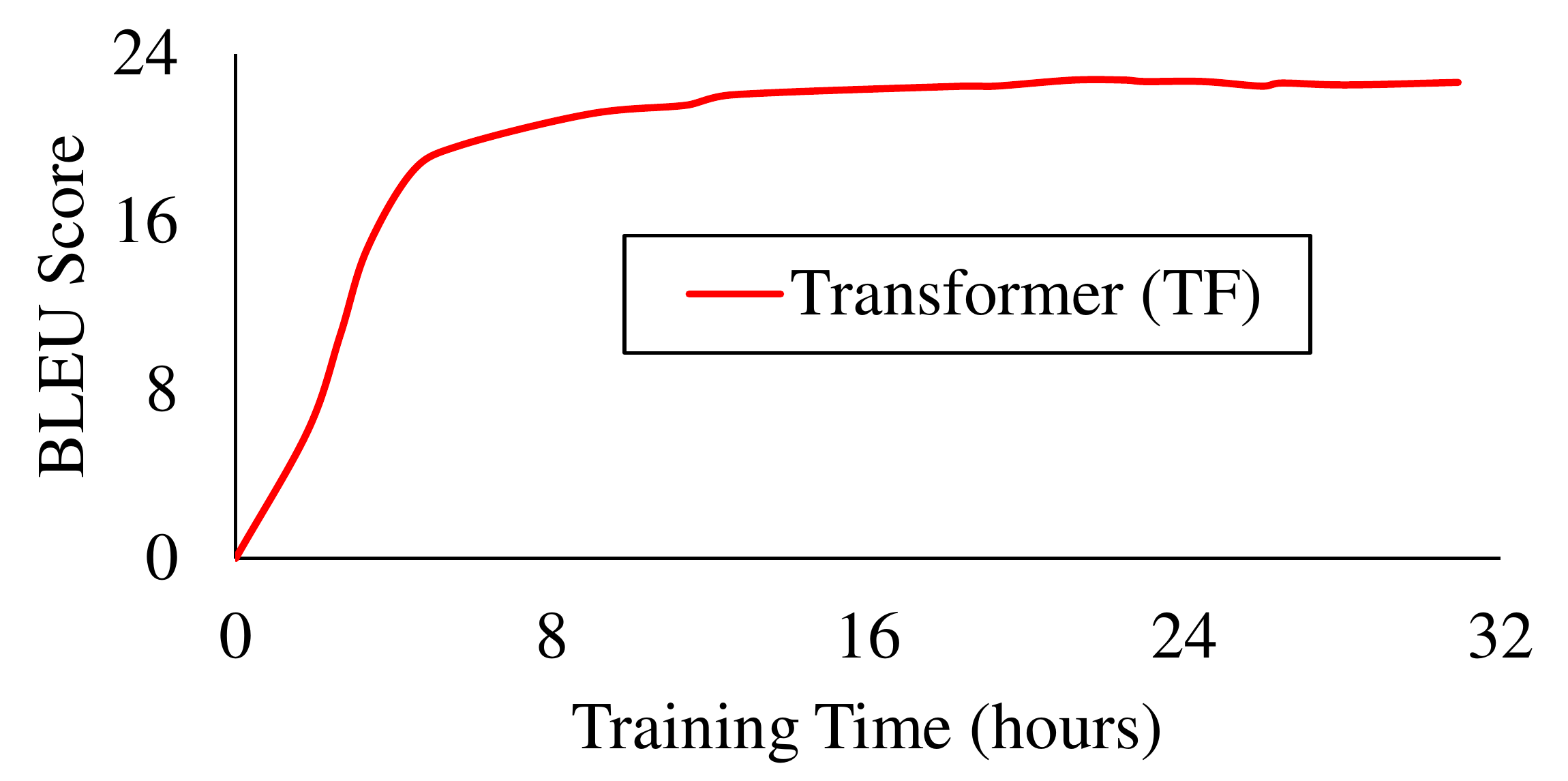}
        \caption{Transformer}
        \label{fig:training_transformer}
    \end{subfigure}%

    \begin{subfigure}[t]{0.30\textwidth}
        \centering
        \includegraphics[width=\textwidth]{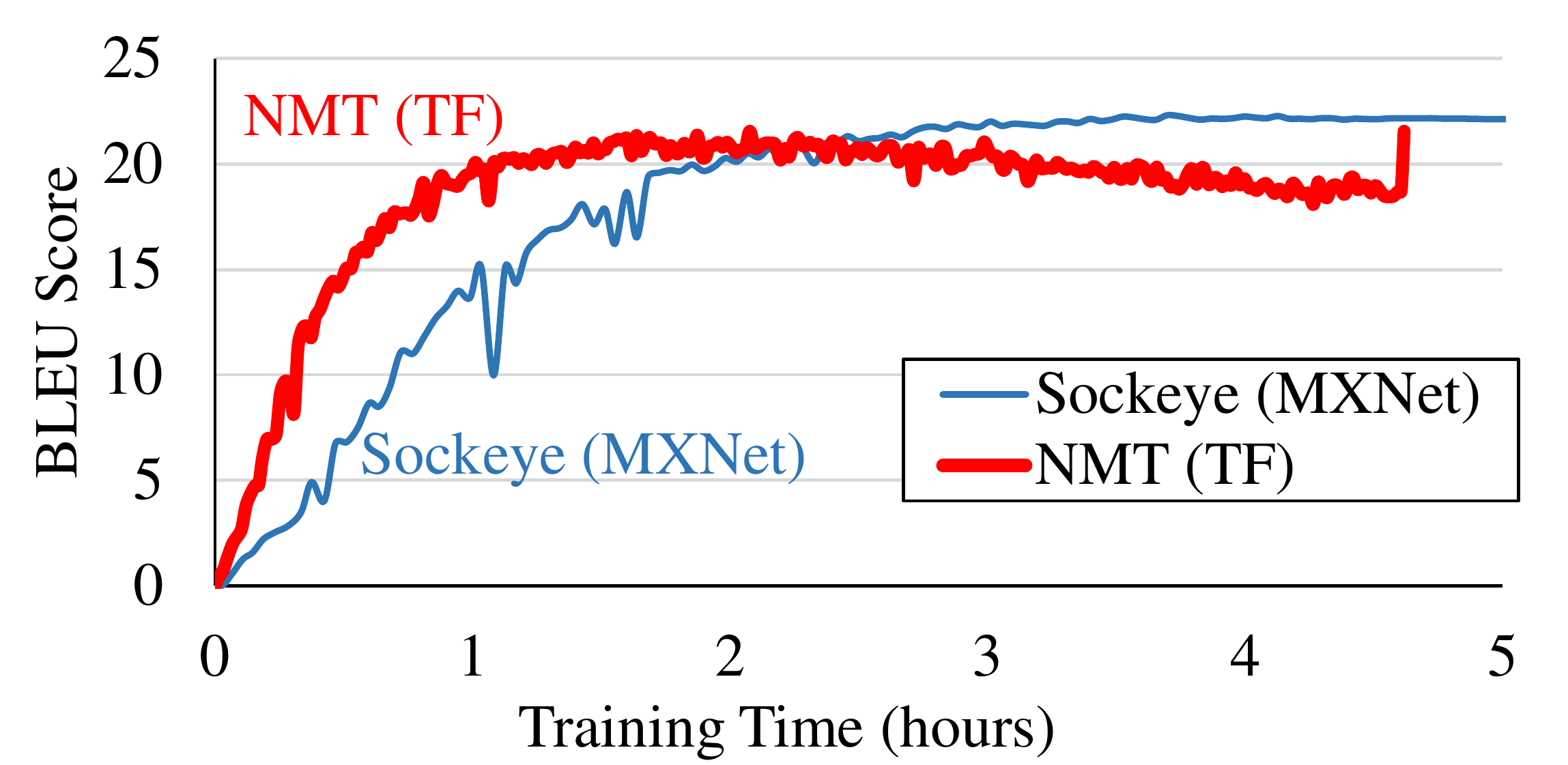}
        \caption{Seq2Seq}
        \label{fig:training_seq2seq}
    \end{subfigure}%
    \hspace{0.1cm}
    \begin{subfigure}[t]{0.30\textwidth}
        \centering
        \includegraphics[width=\textwidth]{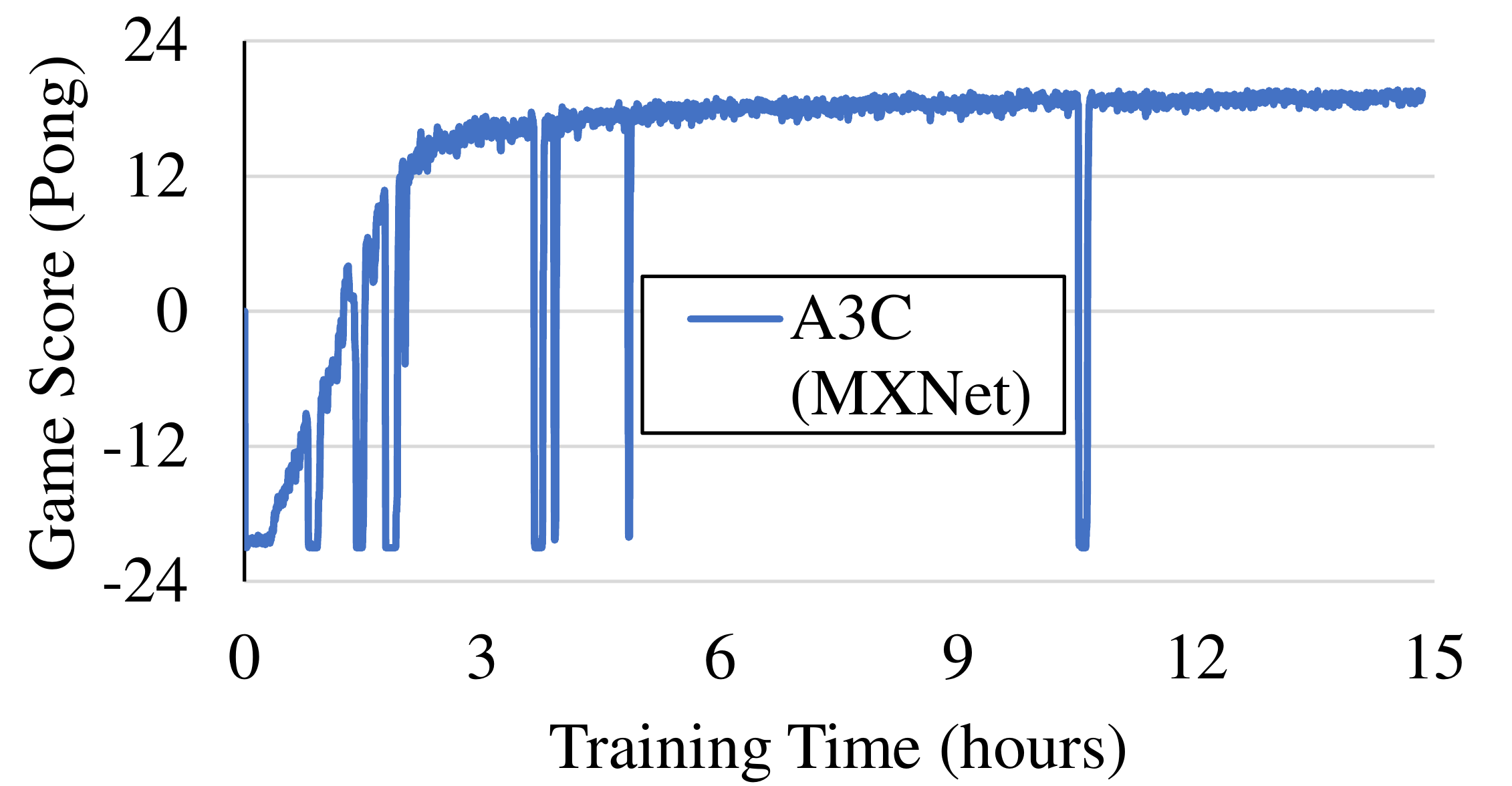}
        \caption{A3C}
        \label{fig:training_a3c}
    \end{subfigure}
    \caption{The model accuracy during the training for different models.}
    \label{fig:training}
\end{figure*}

To ensure that the results we obtain from our measurements are representative
we need to verify that the training process for each model results in classification accuracy comparable to state of the art results
published in the literature.
To achieve this, we train the benchmark models in our suite
until they converge to some expected accuracy rate (based on results
from the literature).
% to verify that
%the implementations we choose can be trained properly.

Figure~\ref{fig:training} shows the classification accuracy observed over time for
five representative models in our benchmark suite, \emph{Inception-v3}, \emph{ResNet-50}, \emph{Seq2Seq}, \emph{Transformer}, and \emph{A3C},
when trained on the single Quadro P4000 GPU hardware configuration described in Section~\ref{sec:results}.
We observe that the training outcome of all models matches results in the literature.
For the two image classification models (\emph{Inception-v3} and
\emph{ResNet-50}) the Top-1 classification accuracy reaches
75--80\% and the the Top-5\footnote{In the Top-5 classification the classifier
can select up to 5 top prediction choices, rather than just 1.}
accuracy is above 90\%,
both in agreement with previously reported results for these models ~\cite{he2016deep}.
The accuracy of the machine
translation models is measured using the BLEU
score~\cite{papineni2002bleu} metric, and we trained our model to achieve
a BLEU score of around 20.
For reinforcement learning, since the models are
generally evaluated by Atari games, the accuracy of the A3C model is directly
reflected by the score of the corresponding game. The A3C curve we show in this figure
is from the Atari Pong game and matches previously reported results for that game (19--20)~\cite{mnih2016asynchronous}.
The training curve shape for different implementations of the same model on different frameworks can vary,
but most of them usually converge to similar accuracy at the end of training.

\subsection{Performance Analysis Framework and Tools}

In this section we describe our analysis toolchain. This toolchain is
designed to help us understand for each of the benchmarks, where the training
time goes, how well the hardware resources are utilized and how to efficiently improve training performance.

\begin{figure} \centering
\includegraphics[width=\textwidth]{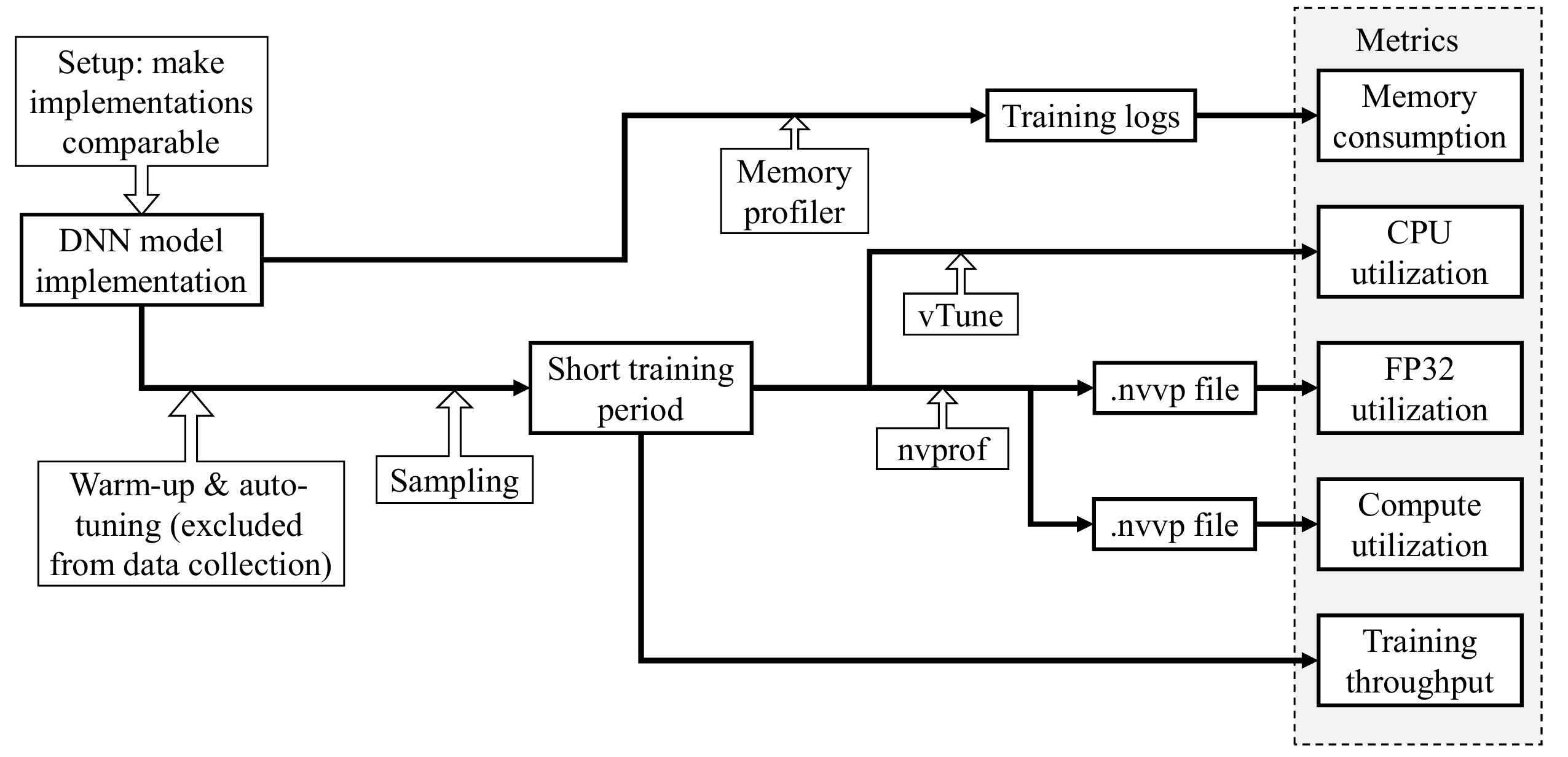}
\caption{Analysis Pipeline}
\label{fig:pipeline}
\end{figure}

\subsubsection{Making implementations comparable across frameworks}

Implementations of the same model on different frameworks might vary in a few aspects that can impact
performance profiling results.
For example, different implementations might have hard-coded values for key hyper-parameters (e.g.,
learning rate, momentum, dropout rate, weight decay) in their code. To make sure that benchmarking identifies
model-specific performance characteristics, rather than just implementation-specific details, we first adapt
implementations of the same model to make them comparable across platforms.
Besides making sure that all implementations run using the same model hyper-parameters, we also
ensure that they define the same network, i.e. the same types and sizes of corresponding layers and layers
are connected in the same way. Moreover, we make sure that the key properties of the training algorithm are
the same across implementations. This is important for models, such as Faster R-CNN~\cite{ren2015faster}, where there are
four different ways in which the training algorithm can share the internal weights.

\subsubsection{Accurate and time-efficient profiling via sampling}

The training of a deep neural network can take days or even weeks  making it impractical to profile
the entire training process. Fortunately, as the training process is an iterative algorithm and almost all
the iterations follow the same computation logic, we find that accurate results can be obtained via sampling
only for a short training period (on the order of minutes) out of the full training run.
In our experiments, we sample 50-1000 iterations and collect the metrics of interest based on these iterations.

To obtain representative results, care must be taken when choosing the sample interval to
ensure that the training process has reached stable state. Upon startup, a typical training procedure first goes through
a warm-up phase (initializing for example the data flow graph, allocating memory and loading data) and then spends some
time auto-tuning various parameters (e.g., system hyper-parameters, such as matrix multiplication algorithms, workspace size).
Only after that the system enters the stable training phase for the remainder of the execution.
While systems do not explicitly indicate when they enter the stable training phase, our experiments show that the warm-up and
auto-tuning phase can be easily identified in measurements. We see that throughput stabilizes after several hundred iterations (a few thousand iterations in the
case of Faster R-CNN). The sample time interval is then chosen after throughput has stabilized.

%Care must be taken when choosing the interval for sampling
%When sampling iterations care
%The number of sampled iterations depend on the metric we collect and the model we profile.
%This sampling enables us to capture performance characteristics of a full training process using a short training period of only minutes.
%For the sampled period to be representative, the training throughput of these iterations should be able
%to represent the entire training process, which needs to be ensured by the next step.

\subsubsection{Relevant metrics}

Below we describe the metrics we collect as part of the profiling process.

\noindent$\bullet$ \emph{Throughput:} Advances in deep neural networks have been tightly coupled to the availability of compute resources capable of efficiently processing
large training data sets. As such, a key metric when evaluating training efficiency is the number of input data samples that is being
processed per second. We refer to this metric as {\em throughput}. Throughput is particularly relevant in the case of DNN training, since training, unlike inference,
is not latency sensitive.

For the speech recognition model we slightly modify our definition of throughput.
Due to the large variations in lengths among the audio data samples, we use the total duration of audio files processed per second instead of the number of files.
The lengths of data samples also varies for machine translation models, but the throughput of these models is still stable so we use the throughput determined by simple counting for them.

\noindent$\bullet$\emph{GPU Compute Utilization:} The GPU is the workhorse behind DNN training, as it is the unit responsible for executing the key operations involved in DNN training (broken down into basic operations such as vector and matrix operations).
Therefore, for optimal throughput, the GPU should be busy all the time.
Low utilization indicates that throughput is limited by other resources, such as CPU or data communication,
and further improvement can be achieved by overlapping CPU runtime or data communication with GPU execution.

We define GPU Compute Utilization as the fraction of time that the GPU is busy (i.e. at least one of its typically many
cores is active):
\begin{equation} \label{formula:occupation} \text{GPU utilization}
= \frac{\text{GPU active time} \times 100}{\text{total elapsed time}} \%  \end{equation}

\noindent$\bullet$ \emph{FP32 utilization:} We also look at GPU utilization from a different angle, measuring how effectively the GPU's resources are being utilized
{\em while the GPU is active}. More specifically, the training of DNNs is typically performed using single-precision floating point operations (FP32),
so a key metric is how well the GPU's compute potential for doing floating point operations is utilized.
We compare the number of FP32 instructions the GPU actually executes while
it is active to the maximal number of FP32 instructions it can theoretically execute during this time,
to determine what percentage of its floating point capacity is utilized.
More precisely, if a GPU's theoretical peak capacity across all its cores is $FLOPS_{peak}$ single-precision floating point operations per second, we
observe the actual number of floating point operations executed during a period of $T$ seconds that the GPU is active, to compute {\em FP32 utilization} as
follows:

\begin{equation} \label{formula:utilization} \text{FP32 utilization} =
 \frac {\text{actual flop count during T} \times 100}{FLOPS_{peak} \times T} \%
\end{equation}

%Gupta et al.~\cite{gupta2015deep} shows that using
%half precision is feasible for training small networks ~\cite{lecun1998gradient,hinton2012improving} on small datasets (MNIST, CIFAR-10).}
%of \emph{nvprof}.

%We therefore define a metric we refer to as {\em FP32 utilization} as follows:
%
%\begin{equation} \label{formula:utilization} \text{overall GPU utilization} =
%\frac{\sum_{k} {\text{$Time_k$}\times \text{$Utilization_k$}}}{\sum_{k}{\text{$Time_k$}}} \% \end{equation}
%
%where $k$ is the index of kernels, $Time_k$ is the duration of kernel $k$, and $Utilization_k$ is the kernel $k$'s GPU utilization level reported by nvprof.

The FP32 utilization gives us a way to calculate the theoretical upper bound of performance improvements one could achieve by a better implementation.
For example, an FP32 utilization of 50\% indicates that we can increase throughput by up to 2x if we manage to increase the FP32 utilization up to 100\%.

In addition to looking at the aggregate FP32 utilization across all cores, we also measure the per-core FP32 utilization for individual kernels,
to identify the kernels with long duration, but low utilization. These kernels should be optimized with high priority.

%Modern GPUs (e.g., Pascal P100~\cite{pascalp100}, Titan Xp~\cite{titanxp}
%contain thousands of cores per chip.
%To exploit the computation power of these cores, one must carefully design the implementation of
%kernels. An inefficient implementation can lead to bad performance of a kernel, and then affect the entire training performance.
%The efficiency of a kernel can be reflected by the utilization level, reported by the \emph{single\_precision\_fu\_utilization}
%metric\footnote{This metric is defined by the average instructions executed over the maximal instructions can be executed per cycle.

\noindent$\bullet$ \emph{CPU utilization:}
While most of the training is typically performed on the GPU, the CPU is also involved, for example, to execute the framework frontends, launch GPU kernels, and transfer the data between CPU and GPU. We report CPU utilization as the average utilization across all cores:
\begin{equation} \label{formula:occupation} \text{CPU utilization}
= \frac{\sum_{c}{\text{total active time of core c}} \times 100}{\text{CPU core count} \times \text{total elapsed time}} \%
\end{equation}

The ratio between the cumulative active time across all cores and total elapsed time is reported by \emph{vTune}, so CPU utilization can
be directly computed from there.

\noindent$\bullet$ \emph{Memory consumption:} In addition to compute cycles, the amount of available
physical memory has become a limiting factor in training large DNNs.
In order to optimize memory usage during DNN training, it is important
to understand where the memory goes, i.e. what data structures
occupy most of the memory. Unfortunately, there are
no open-source tools currently available for existing frameworks that can provide this analysis.
Hence we build our own memory profilers for
three main frameworks (TensorFlow, MXNet, and CNTK). \textit{We will open source
these tools together with our benchmarks, as we expect them to be useful to others in
developing and analyzing their models.}

%Performance is not the only metric that matters. The size of a neural network
%is also limited by the memory capacity of a GPU. This issue was pointed out
%early by Krizhevsky et al.~\cite{krizhevsky2012imagenet}, and was
%carefully studied recently by Rhu et al.~\cite{rhu2016vdnn}. Despite
%that memory capacity will increase as the advancement of architecture,
%resources will be quickly exhausted by larger DNN models and bigger data. The
%physical memory has become a limit to the capability of a single GPU training a
%large DNN.

When building our memory profiler we carefully inspect how the different
DNN frameworks in our benchmark allocate their memory and identify the data structures
that are the main consumers of memory.
We observe that most data structures are allocated before the training iterations start for these three frameworks.
Each of the data structures usually belongs to one of the three types: weights, weight gradients
and feature maps (similarly to prior works~\cite{rhu2016vdnn}). These data structures are
allocated statically. In addition, a framework might allocate some workspace as a temporary container for
intermediate results in a kernel function, which gives us another type of data structure.
The allocation of workspace can be either static, before the training iterations, or dynamic,
 during the training iterations. We observe that in MXNet, data structures other than
workspace are allocated during the training iterations (usually for the momentum computation)
as well. We assign these data structures to a new type called "dynamic". As memory can be
allocated and released during the training, we measure the memory consumption by the
maximal amount of memory ever allocated for each type.

%Rhu et al.~\cite{rhu2016vdnn} already shows the memory functionality break-down for some
%image classification models. In this work, we extend this analysis to a
%much wider benchmark suite.

%\input{sections/benchmark}
\section{Evaluation}
\label{sec:results}
In this section, we use the methodology and framework described
in the previous section for a detailed performance evaluation and analysis of the models in our \texttt{TBD}
benchmark suite.

%The training datasets we use are described in Table~\ref{table:dataset} and are
%the ones used in real world training.

\subsection{Experimental Setup}
We use Ubuntu 16.04 OS, TensorFlow v1.3, MXNet v0.11.0, CNTK v2.0, with CUDA 8
and cuDNN 6.
%These frameworks have large amount of active users. Each of them is sponsored
%by big cooperations (Tensorflow by Google, CNTK by Microsoft, MXNet by
%Amazon), and therefore currently actively evolving.
All of our experiments are carried out on a 16-machine cluster, where each node
is equipped with a Xeon 28-core CPU and one to four NVidia Quadro P4000 GPUs.
Machines are connected with both Ethernet and high speed Infiniband (100
Gb/sec) network cards.
%The models are executed on Tensorflow v1.3, Mxnet v0.11.0, CNTK v2.0, with
%CUDA version 8 and cudnn version 6.

As different GPU models provide a tradeoff between cost, performance, area and power,
it is important to understand how different GPUs affect the key metrics in DNN training.
We therefore also repeat a subset of our experiments using a second type of GPU,
the NVidia TITAN Xp GPU.  Table~\ref{table:GPU-spec} compares the technical specifications of
the two GPUs in our work.
We show the comparative throughput and comparisons of our metrics between TITAN Xp and P4000 in
Section~\ref{sec:result-hardware}.

\begin{table} [h!]
\footnotesize	
\centering
{
    \begin{tabular}{ |c|c|c|c| }
    \hline
     & Titan Xp & Quadro P4000 & Intel Xeon E5-2680 \\ \hline %Intel(R) Xeon(R) CPU E5-2680 v4 \\ \hline
    Multiprocessors & 30 & 14 & \\ \hline
    Core Count & 3840 & 1792 & 28 \\ \hline
    Max Clock Rate (MHz) & 1582 & 1480 & 2900 \\ \hline
    Memory Size (GB) & 12 & 8 & 128 \\ \hline
    LLC Size (MB) & 3 & 2 & 35 \\ \hline
    Memory Bus Type & GDDR5X & GDDR5 & DDR4 \\ \hline
    Memory BW (GB/s) & 547.6 & 243 & 76.8 \\ \hline
    Bus Interafce & PCIe 3.0 & PCIe 3.0  & \\ \hline
    Memory Speed (MHz) & 5705 & 3802 & 2400 \\ \hline
%    \bottomrule
    \end{tabular}
}
    \caption{Hardware specifications}
    \label{table:GPU-spec}
\vspace{-0.5cm}
\end{table}

\subsection{Performance Analysis}
\label{sec:performance}
As  previously explained, our analysis will focus on a set of key metrics: throughput,
GPU and CPU compute utilization, FP32 utilization, as well as a memory consumption breakdown.

Since one of the aspects that makes our work unique is the breadth in application domains,
models and frameworks covered by our \texttt{TBD} benchmark suite we will pay particular attention
to how the above metrics vary across applications, models and frameworks.

Moreover, we will use our setup to study the effects of a key hyper-parameter,
the mini-batch size, on our metrics.
It has been shown that to achieve high training throughput with the power of
multiple GPUs using data parallelism, one must increase the mini-batch size,
and additional work needs to be done on model parameters such as learning rate
to preserve the training accuracy~\cite{goyal2017accurate,you2017imagenet}. In
the single-GPU case, it is often assumed that larger mini-batch size will translate
to higher GPU utilization, but the exact effects of varying mini-batch size
are not well understood.
In this work, we use our setup to quantify in detail
how mini-batch size affects key performance
metrics.

\begin{figure*} [h!]
    \centering
    \begin{subfigure}[t]{0.23\textwidth}
        \centering
        \includegraphics[width=\textwidth]{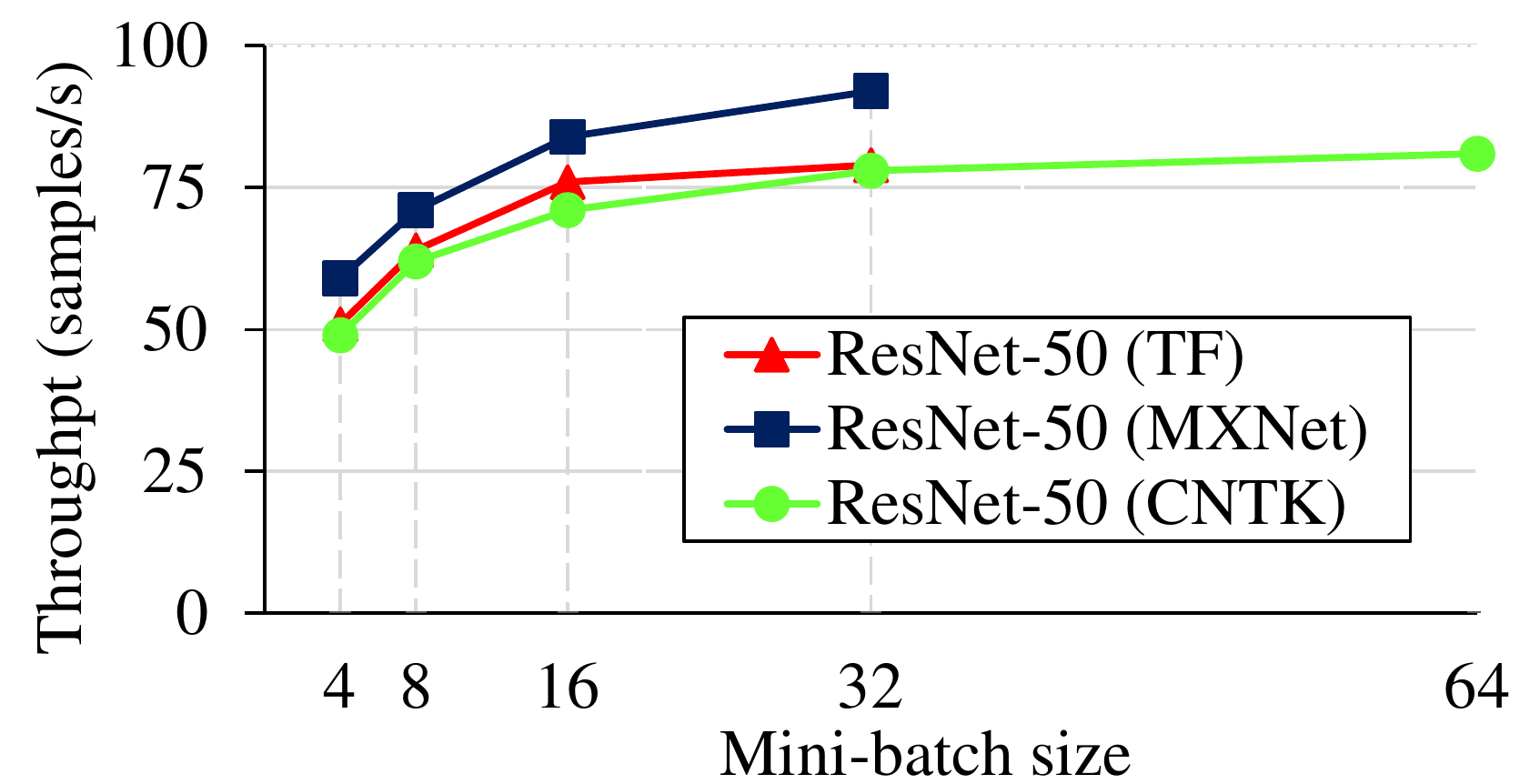}
        \caption{ResNet-50}
        \label{fig:throughput_resnet}
    \end{subfigure}%
    \hspace{0.1cm}
    \begin{subfigure}[t]{0.23\textwidth}
        \centering
        \includegraphics[width=\textwidth]{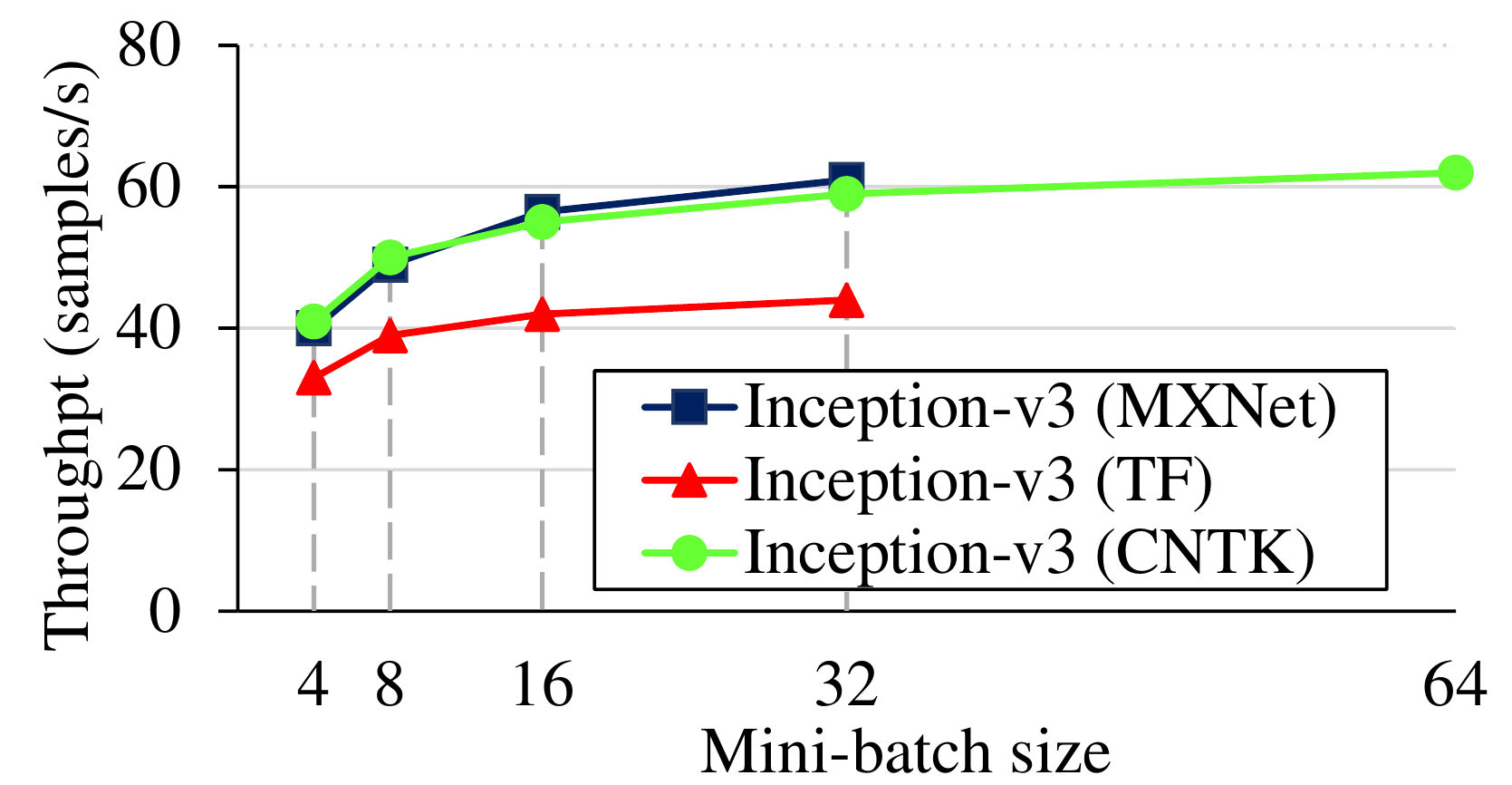}
        \caption{Inception-v3}
        \label{fig:throughput_inception}
    \end{subfigure}
    \begin{subfigure}[t]{0.23\textwidth}
        \centering
        \includegraphics[width=\textwidth]{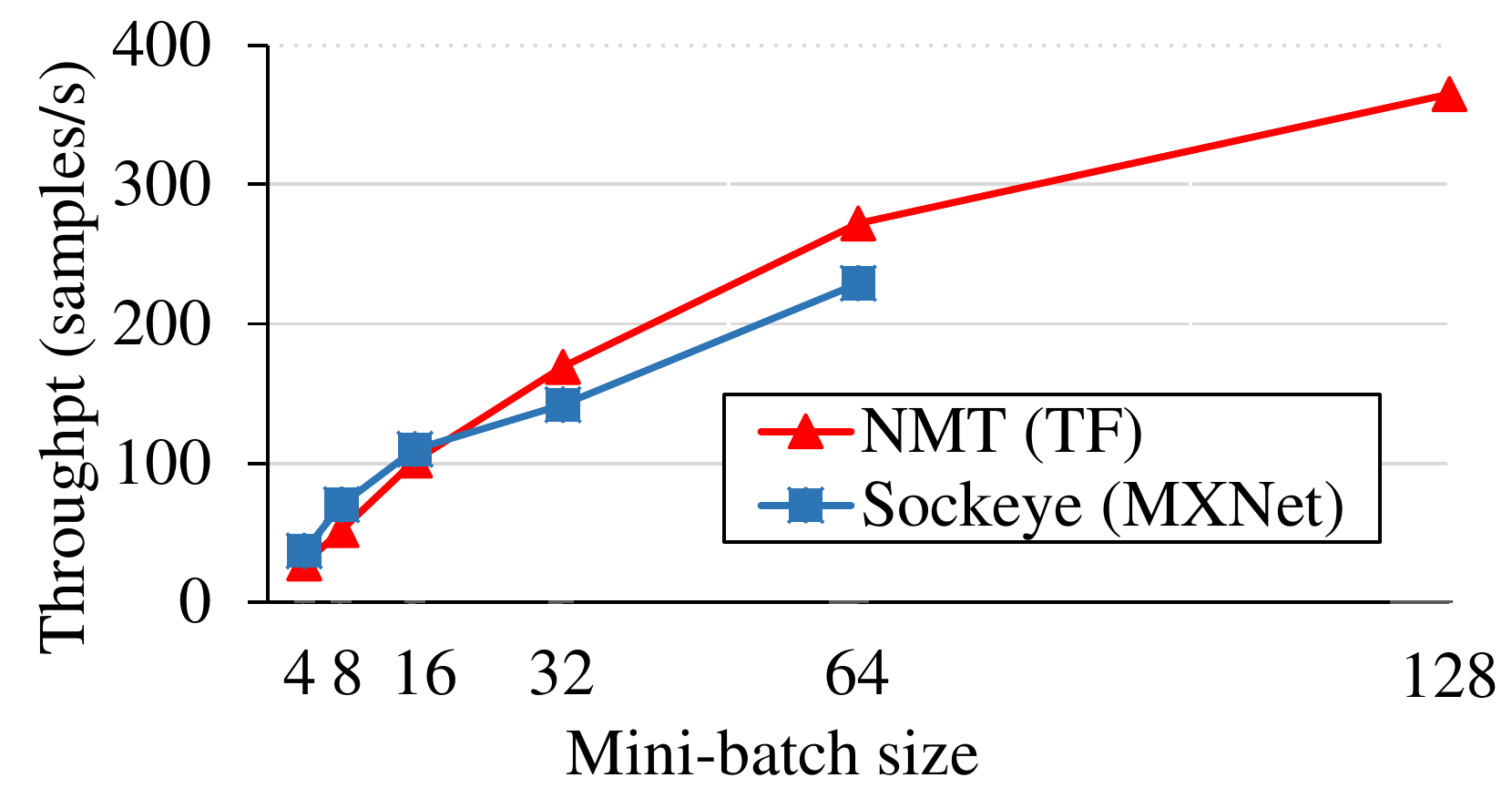}
        \caption{Seq2Seq}
        \label{fig:throughput_seq2seq}
    \end{subfigure}
    \hspace{0.1cm}
    \begin{subfigure}[t]{0.23\textwidth}
        \centering
        \includegraphics[width=\textwidth]{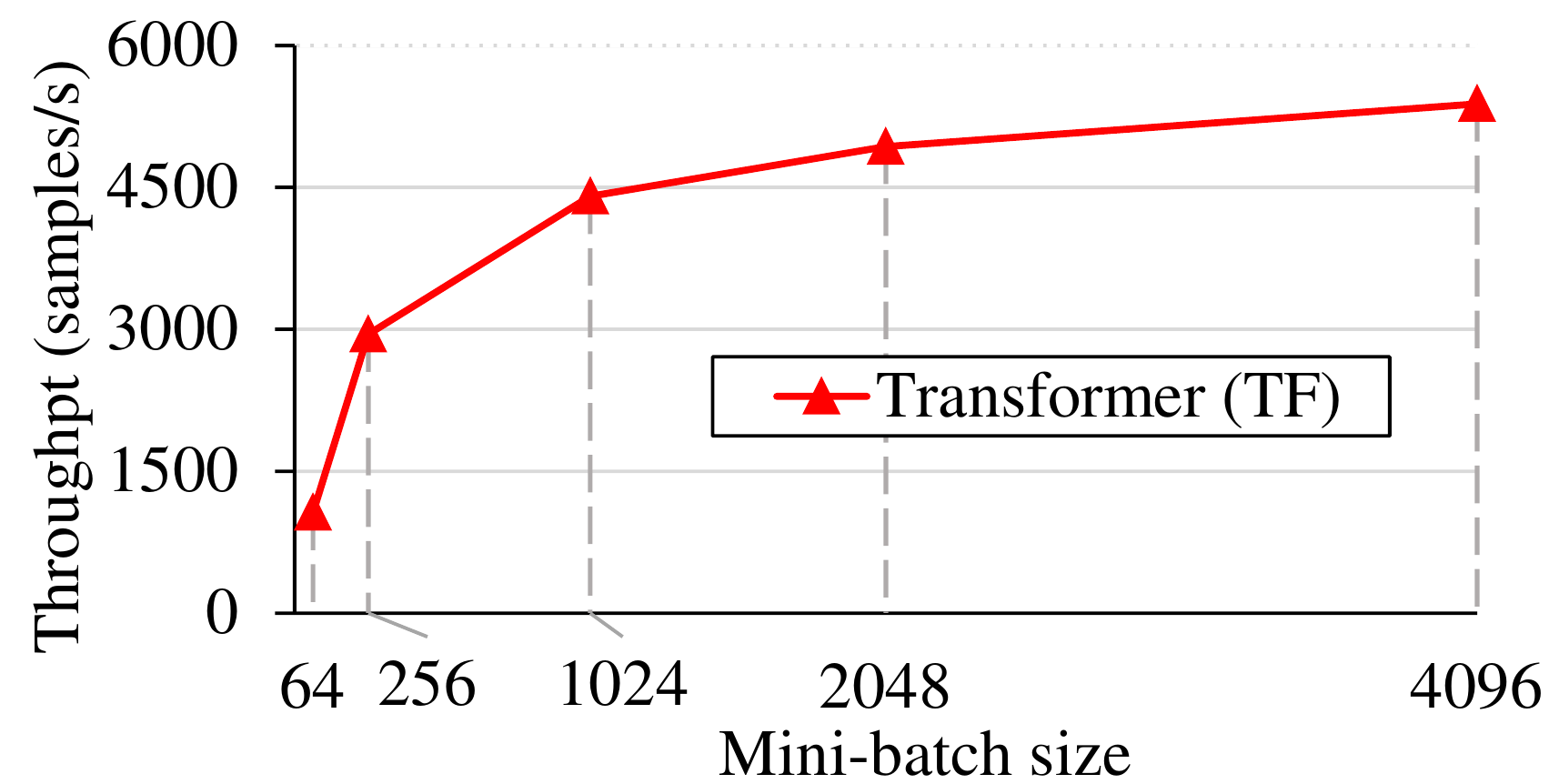}
        \caption{Transformer}
        \label{fig:throughput_transformer}
    \end{subfigure}

    \begin{subfigure}[t]{0.23\textwidth}
        \centering
        \includegraphics[width=\textwidth]{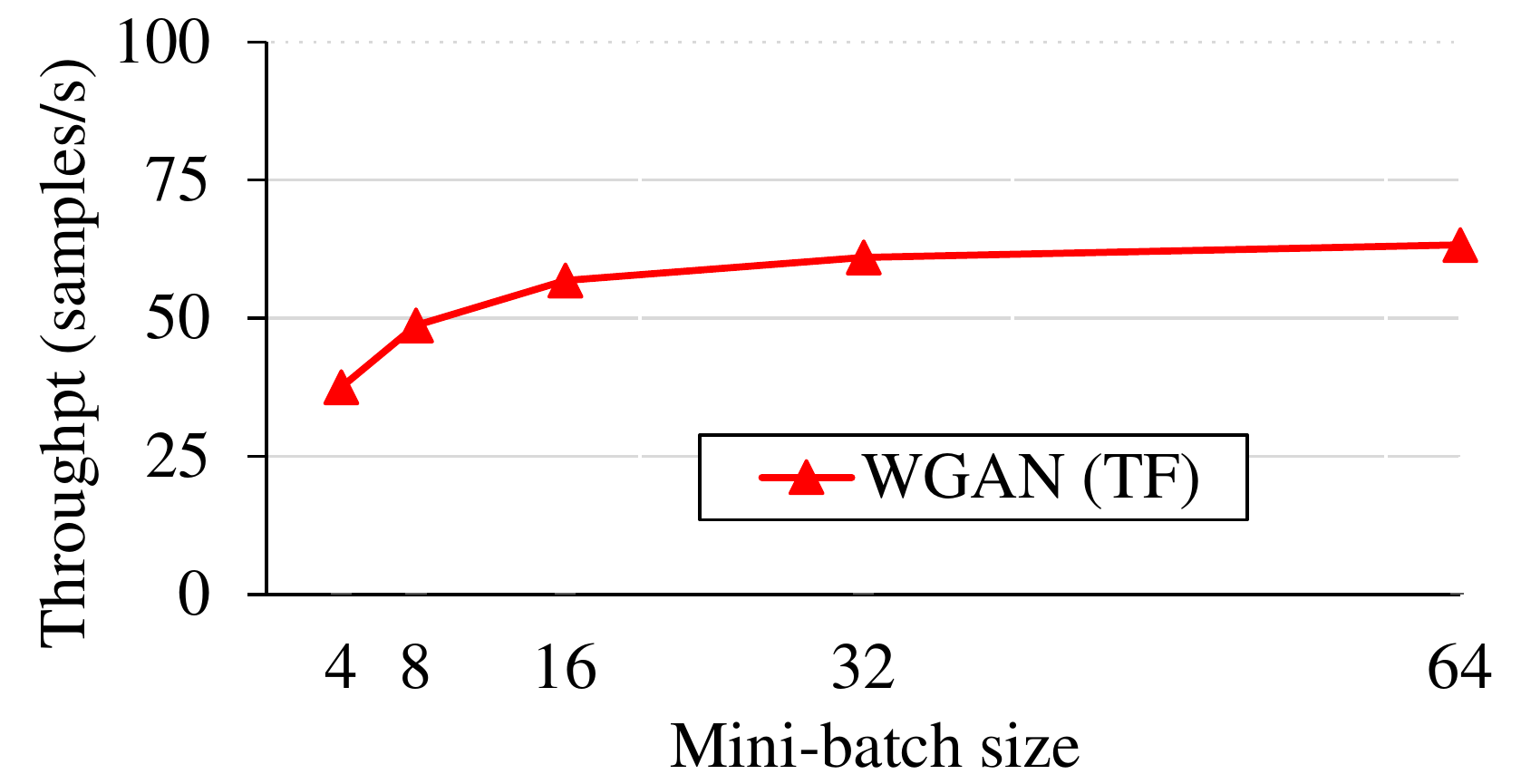}
        \caption{WGAN}
        \label{fig:throughput_wgan}
    \end{subfigure}
    \hspace{0.1cm}
    \begin{subfigure}[t]{0.23\textwidth}
        \centering
        \includegraphics[width=\textwidth]{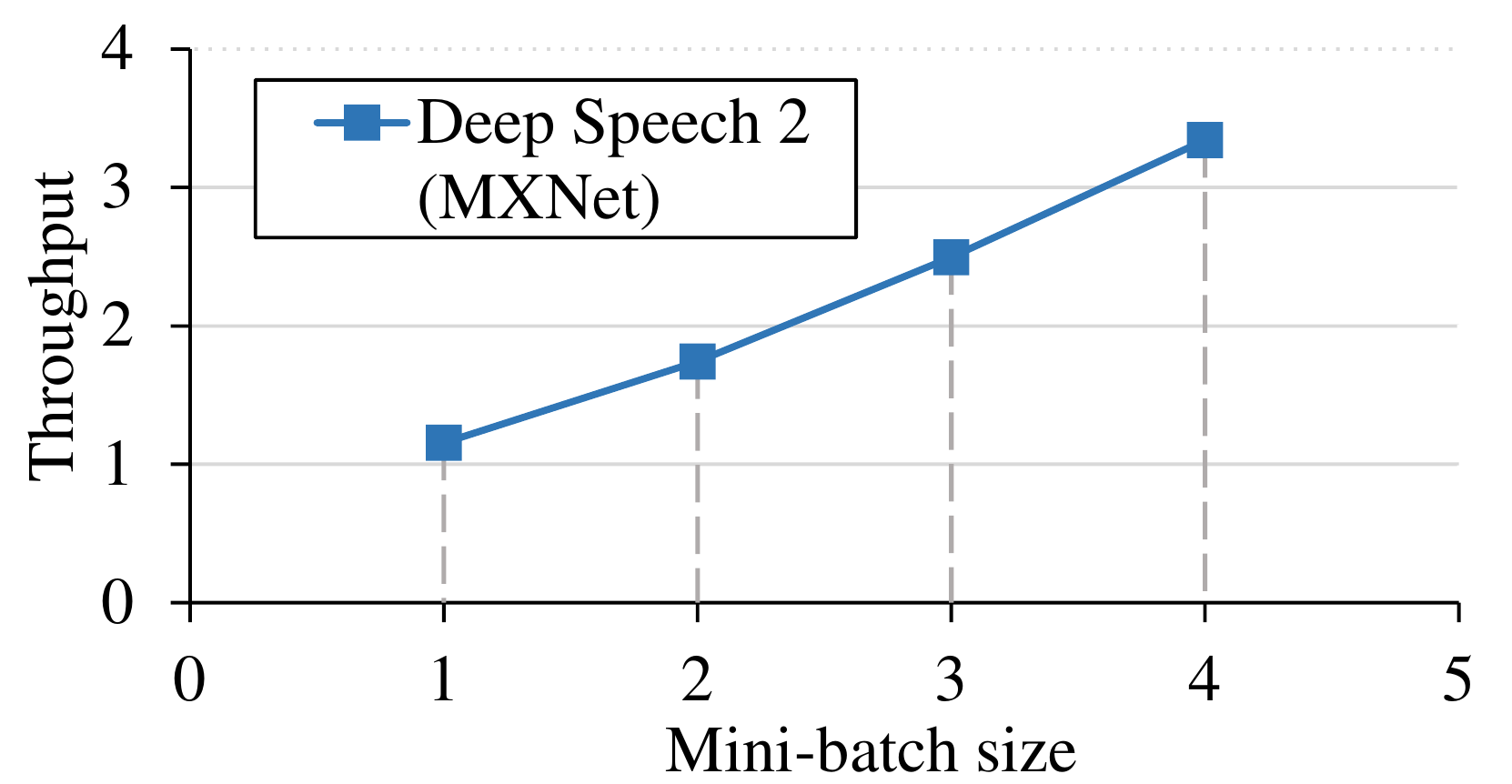}
        \caption{Deep Speech 2}
        \label{fig:throughput_ds2}
    \end{subfigure}
    \hspace{0.1cm}
    \begin{subfigure}[t]{0.23\textwidth}
        \centering
        \includegraphics[width=\textwidth]{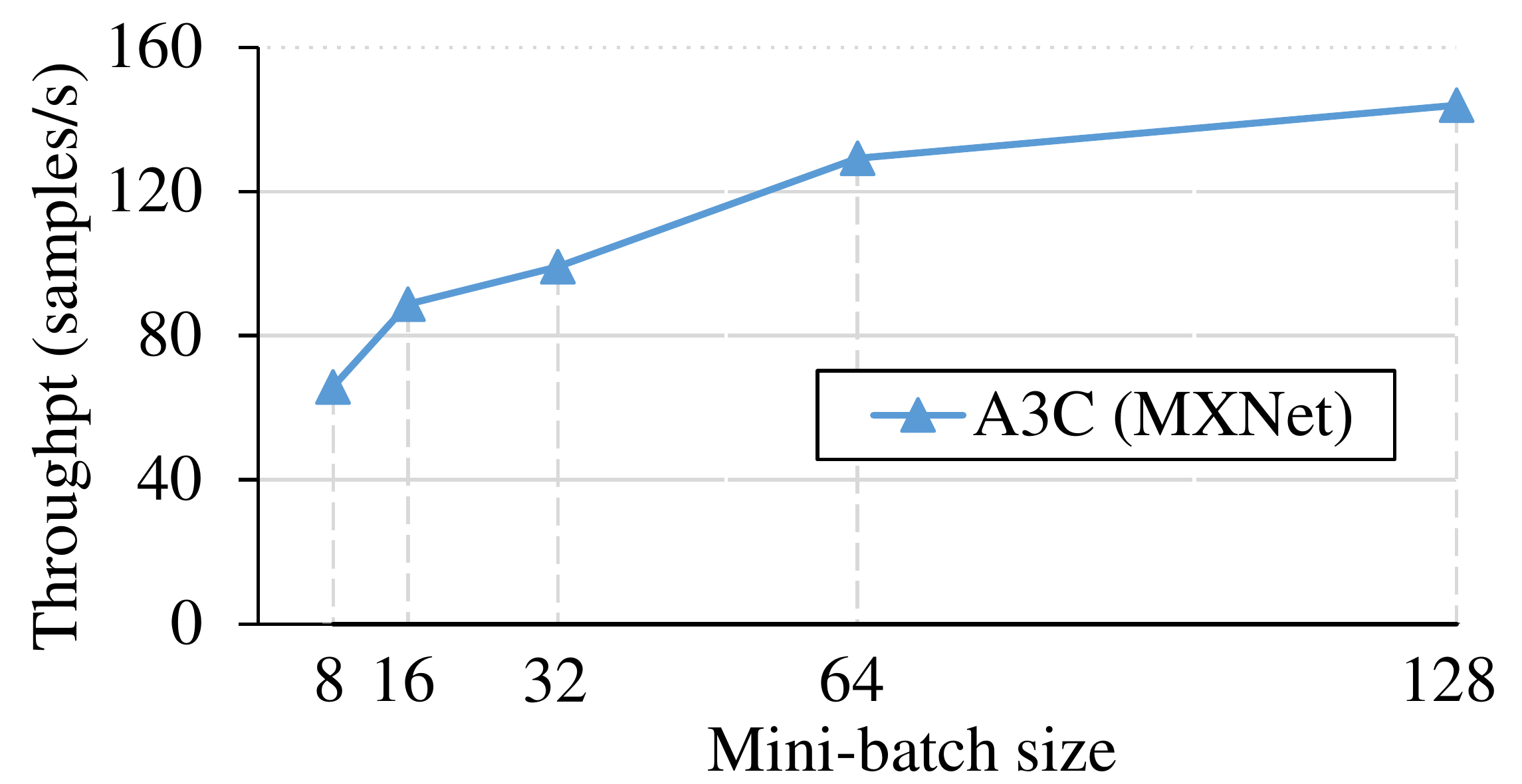}
        \caption{A3C}
        \label{fig:throughput_a3c}
    \end{subfigure}
    \caption{DNN training throughput for different models on multiple mini-batch sizes.}
    \label{fig:throughput}
\end{figure*}

\begin{figure*} [h!]
    \centering
    \begin{subfigure}[t]{0.23\textwidth}
        \centering
        \includegraphics[width=\textwidth]{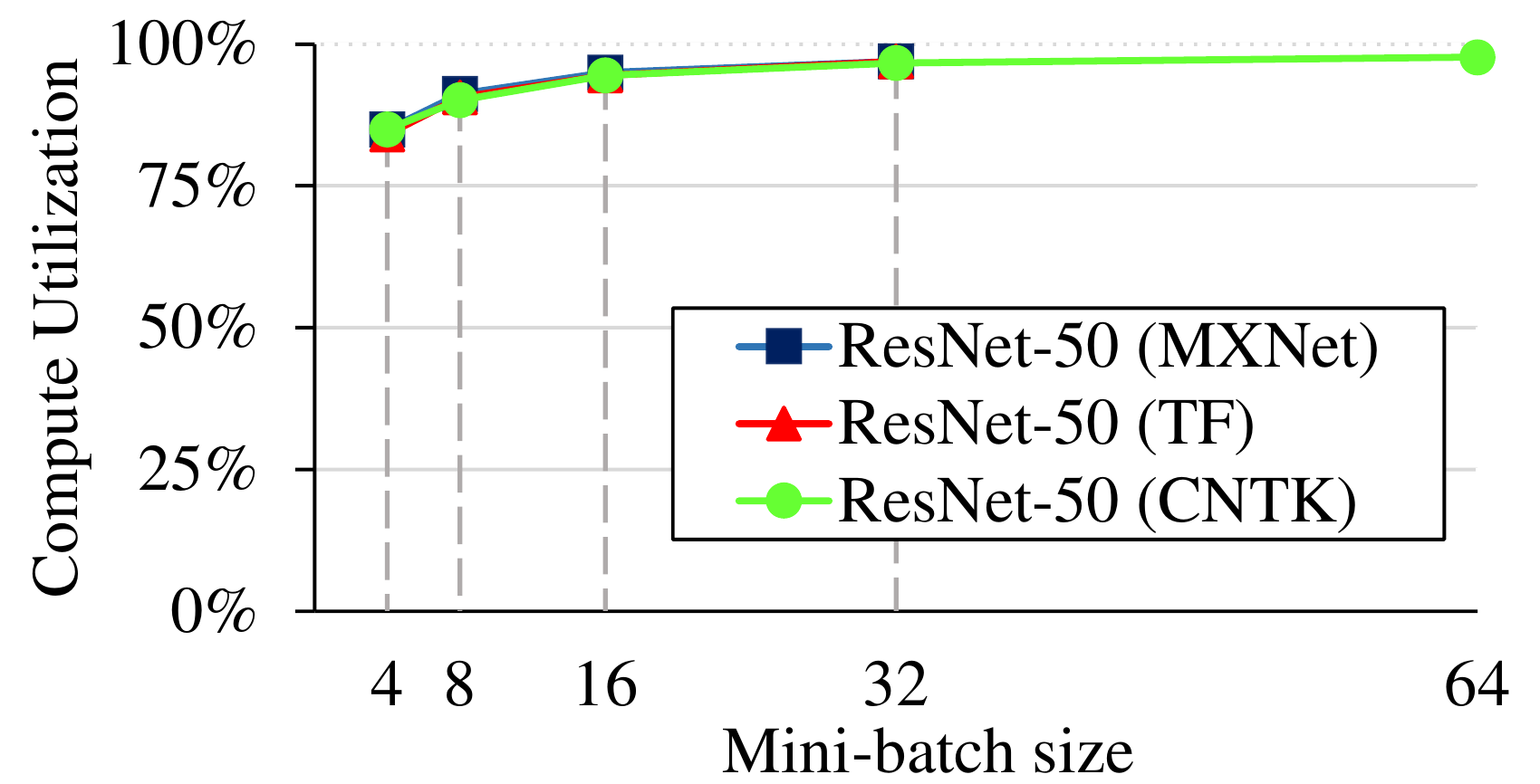}
        \caption{ResNet-50}
        \label{fig:occupation_resnet}
    \end{subfigure}%
    \hspace{0.1cm}
    \begin{subfigure}[t]{0.23\textwidth}
        \centering
        \includegraphics[width=\textwidth]{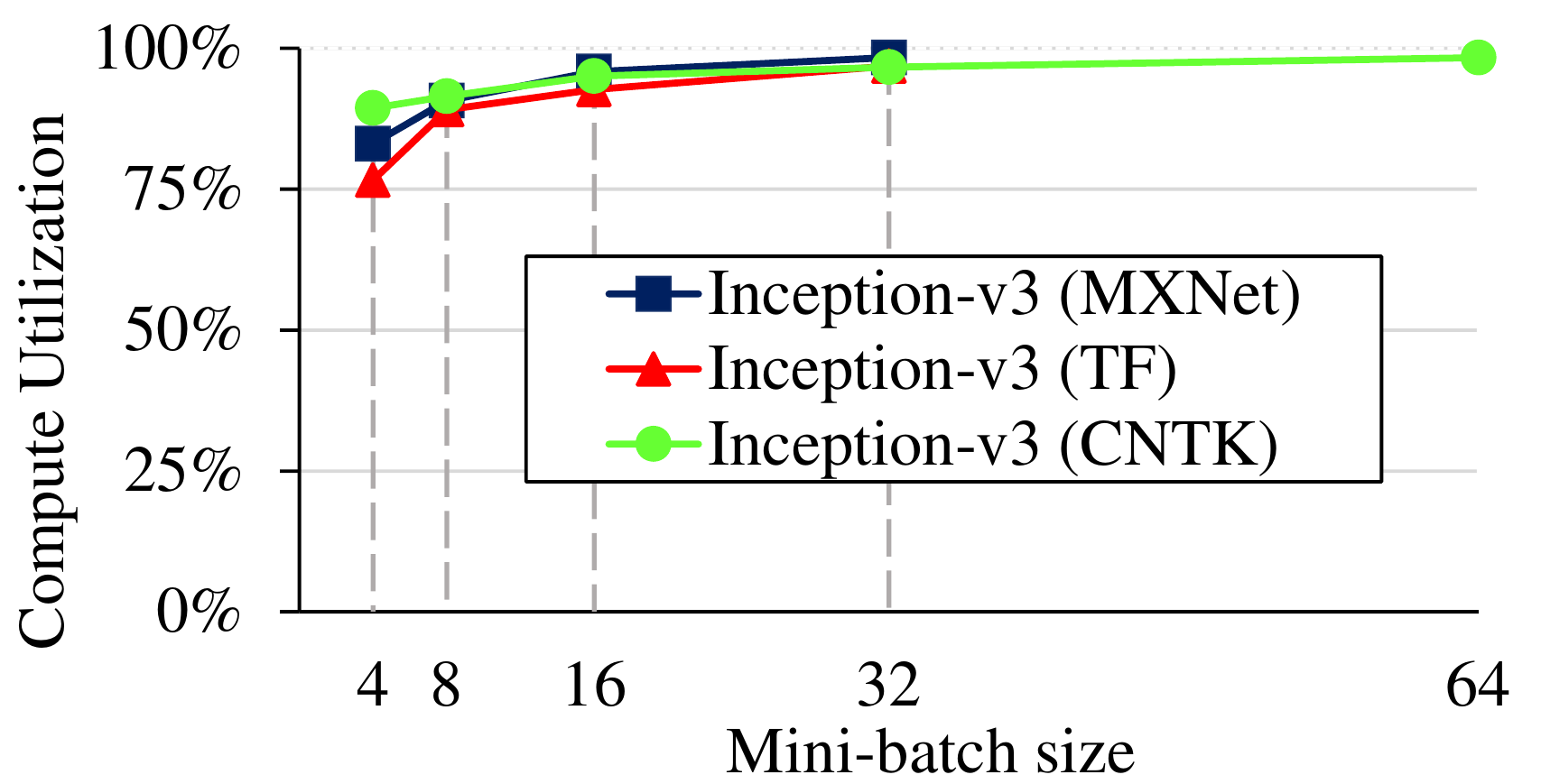}
        \caption{Inception-v3}
        \label{fig:occupation_inception}
    \end{subfigure}%
    \hspace{0.1cm}
    \begin{subfigure}[t]{0.23\textwidth}
        \centering
        \includegraphics[width=\textwidth]{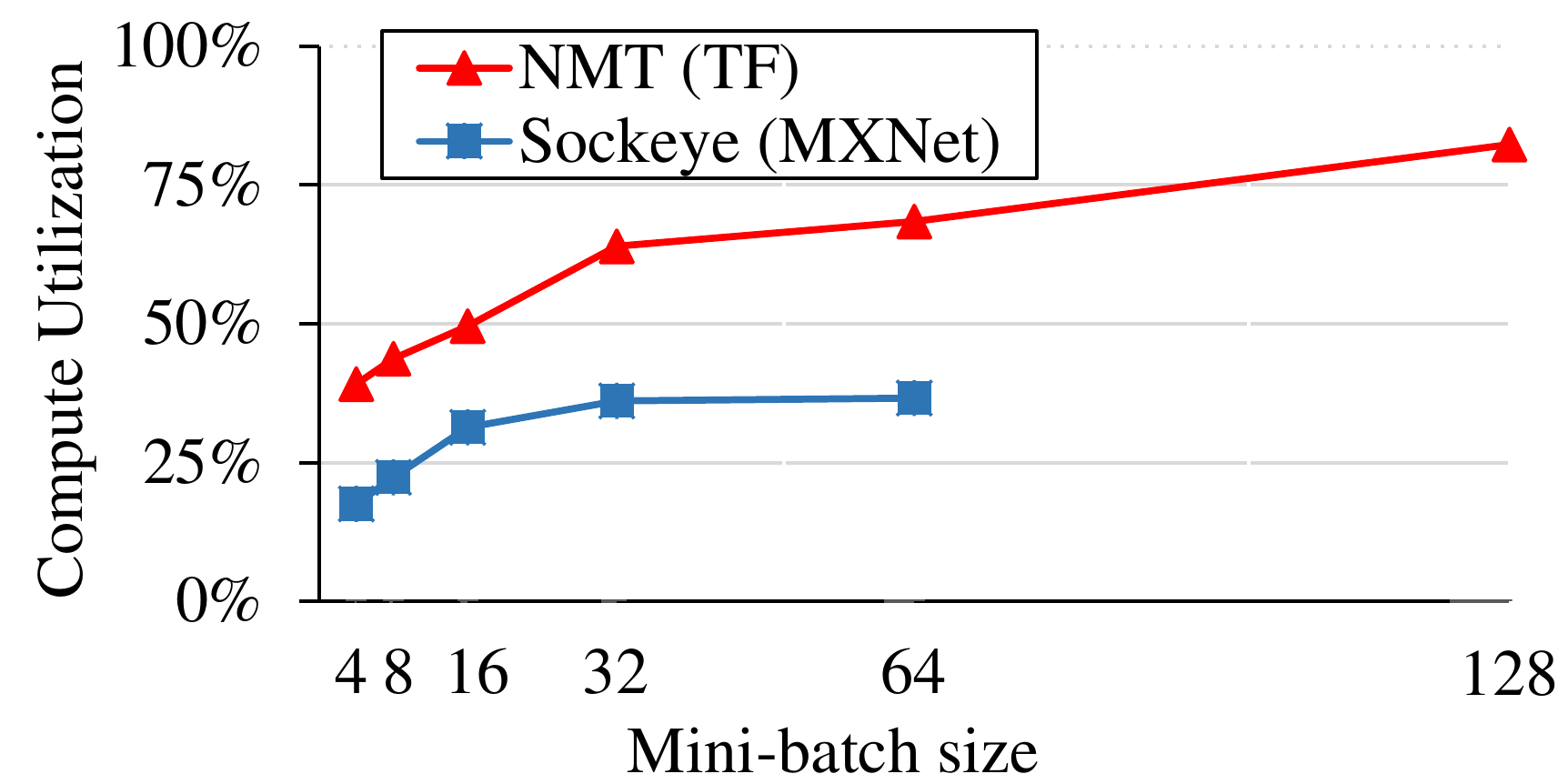}
        \caption{Seq2Seq}
        \label{fig:occupation_seq2seq}
    \end{subfigure}
    \hspace{0.1cm}
    \begin{subfigure}[t]{0.23\textwidth}
        \centering
        \includegraphics[width=\textwidth]{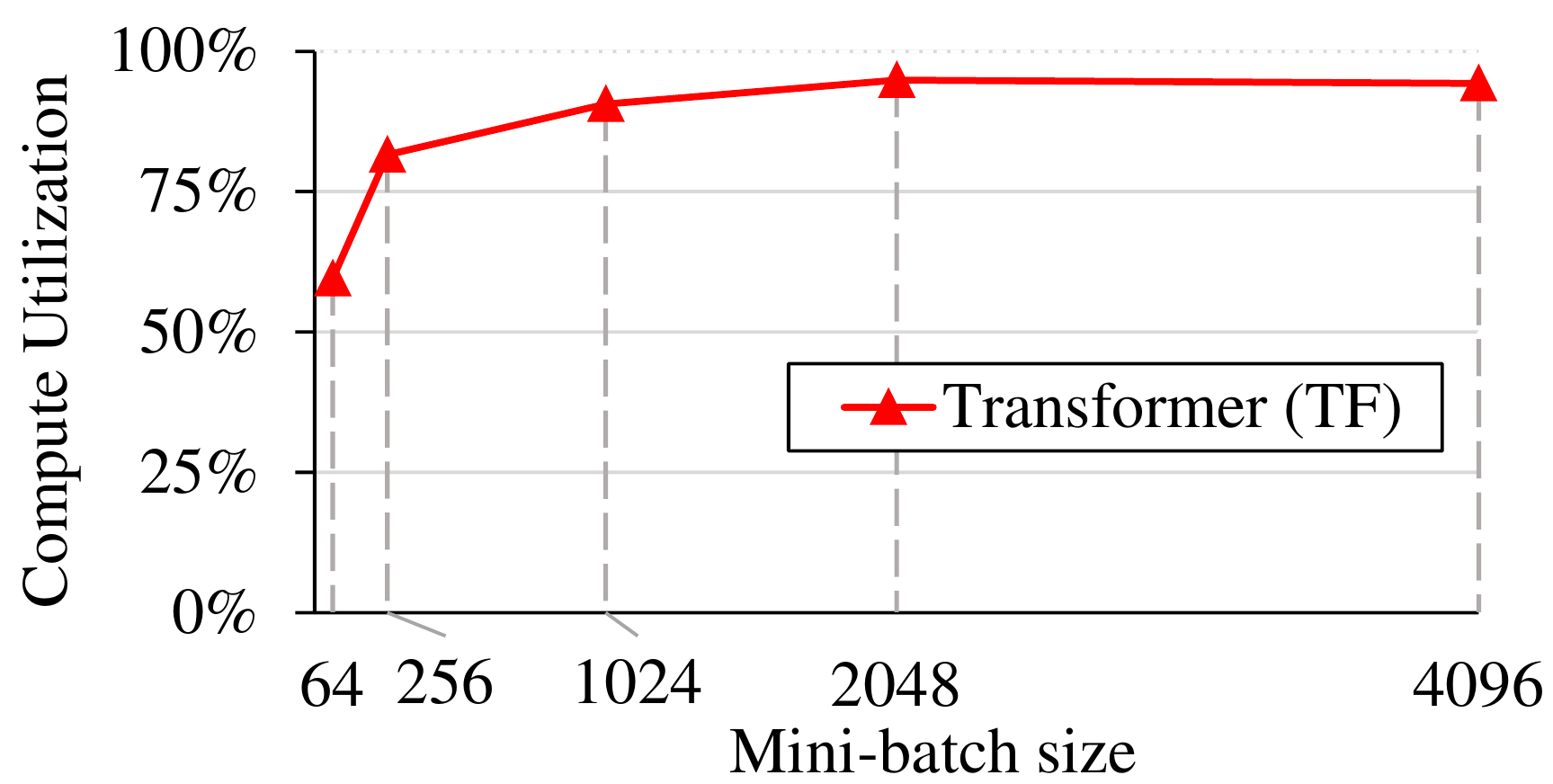}
        \caption{Transformer}
        \label{fig:occupation_transformer}
    \end{subfigure}

    \begin{subfigure}[t]{0.23\textwidth}
        \centering
        \includegraphics[width=\textwidth]{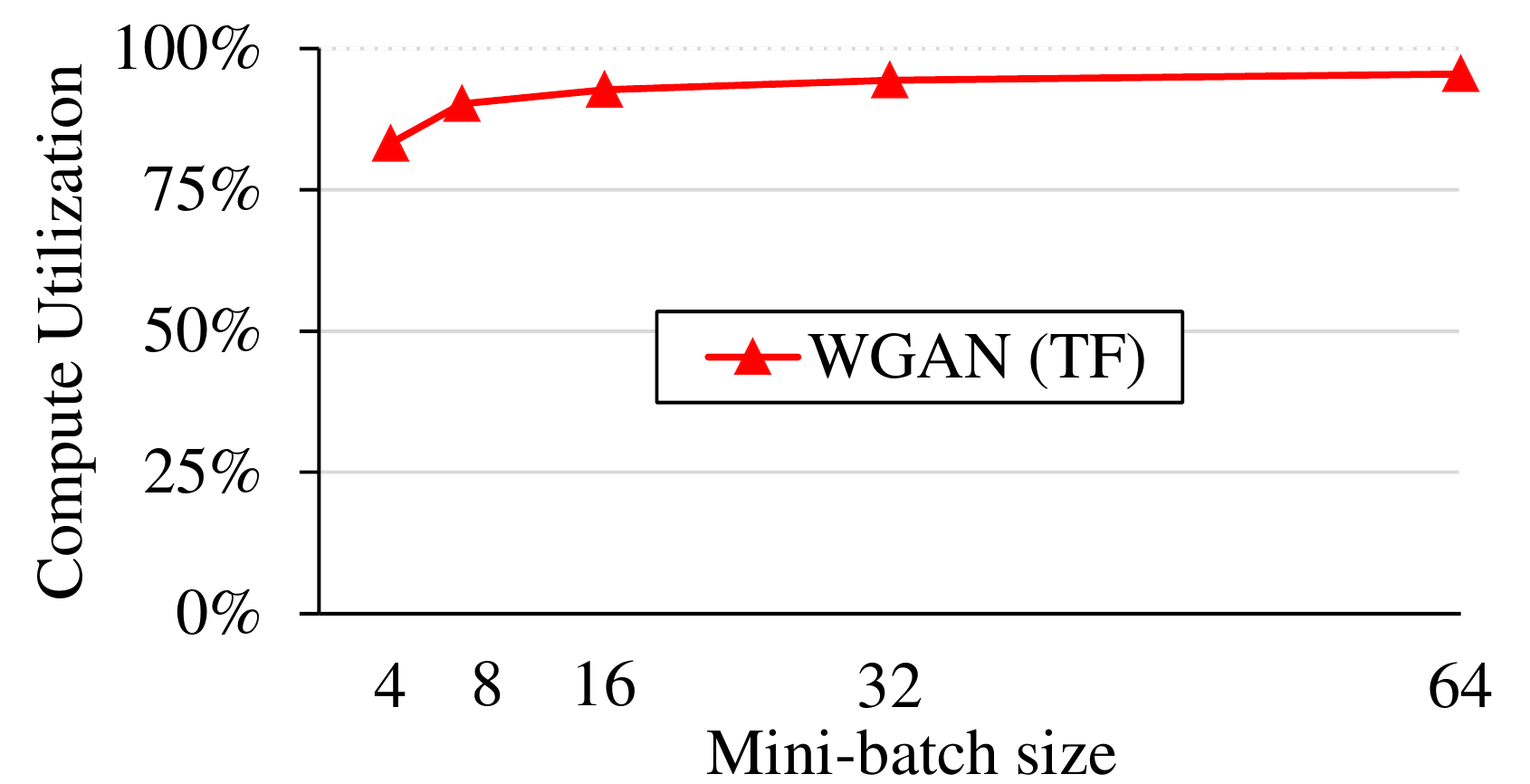}
        \caption{WGAN}
        \label{fig:occupation_wgan}
    \end{subfigure}
    \hspace{0.1cm}
    \begin{subfigure}[t]{0.23\textwidth}
        \centering
        \includegraphics[width=\textwidth]{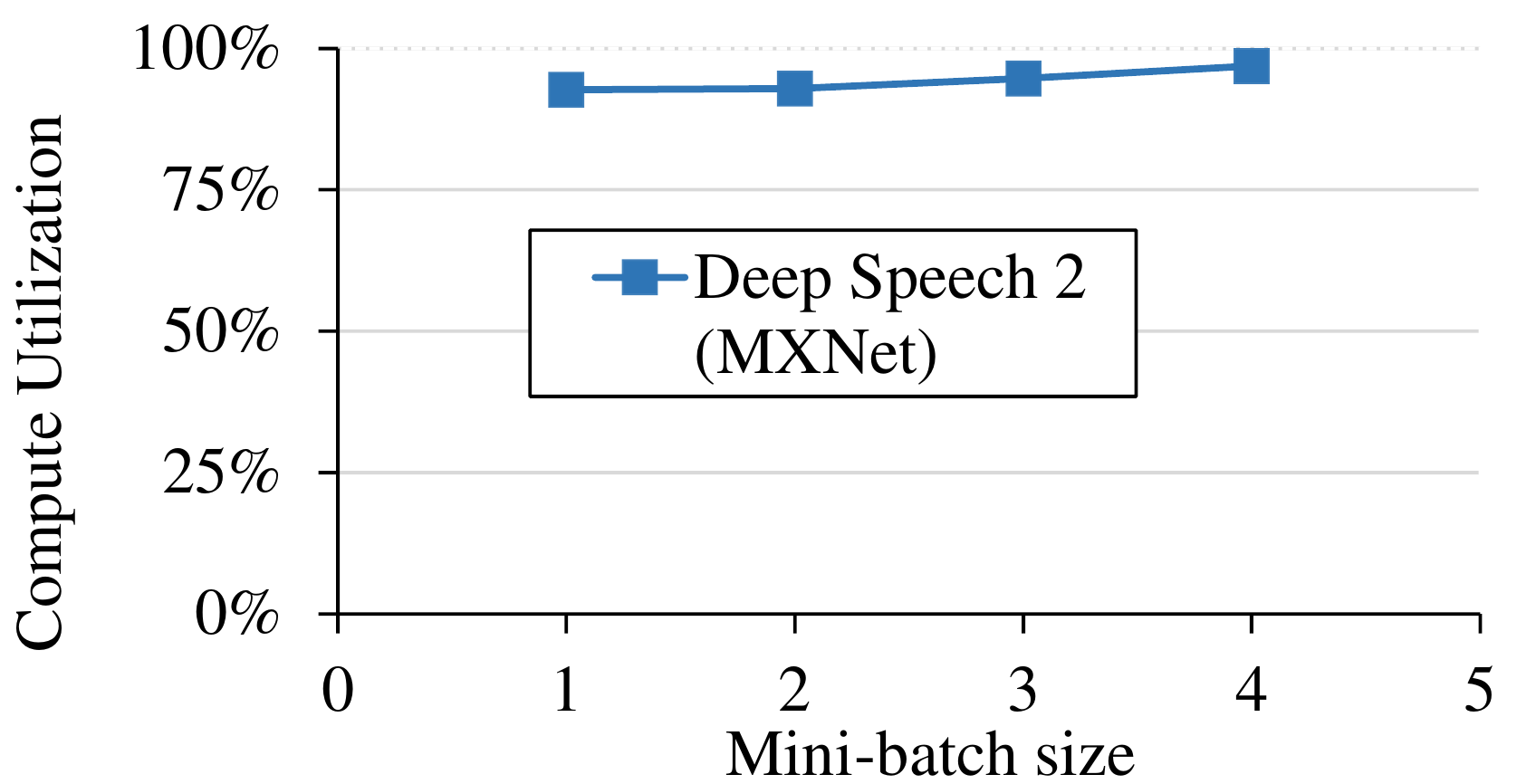}
        \caption{Deep Speech 2}
        \label{fig:occupation_ds2}
    \end{subfigure}
    \hspace{0.1cm}
    \begin{subfigure}[t]{0.23\textwidth}
        \centering
        \includegraphics[width=\textwidth]{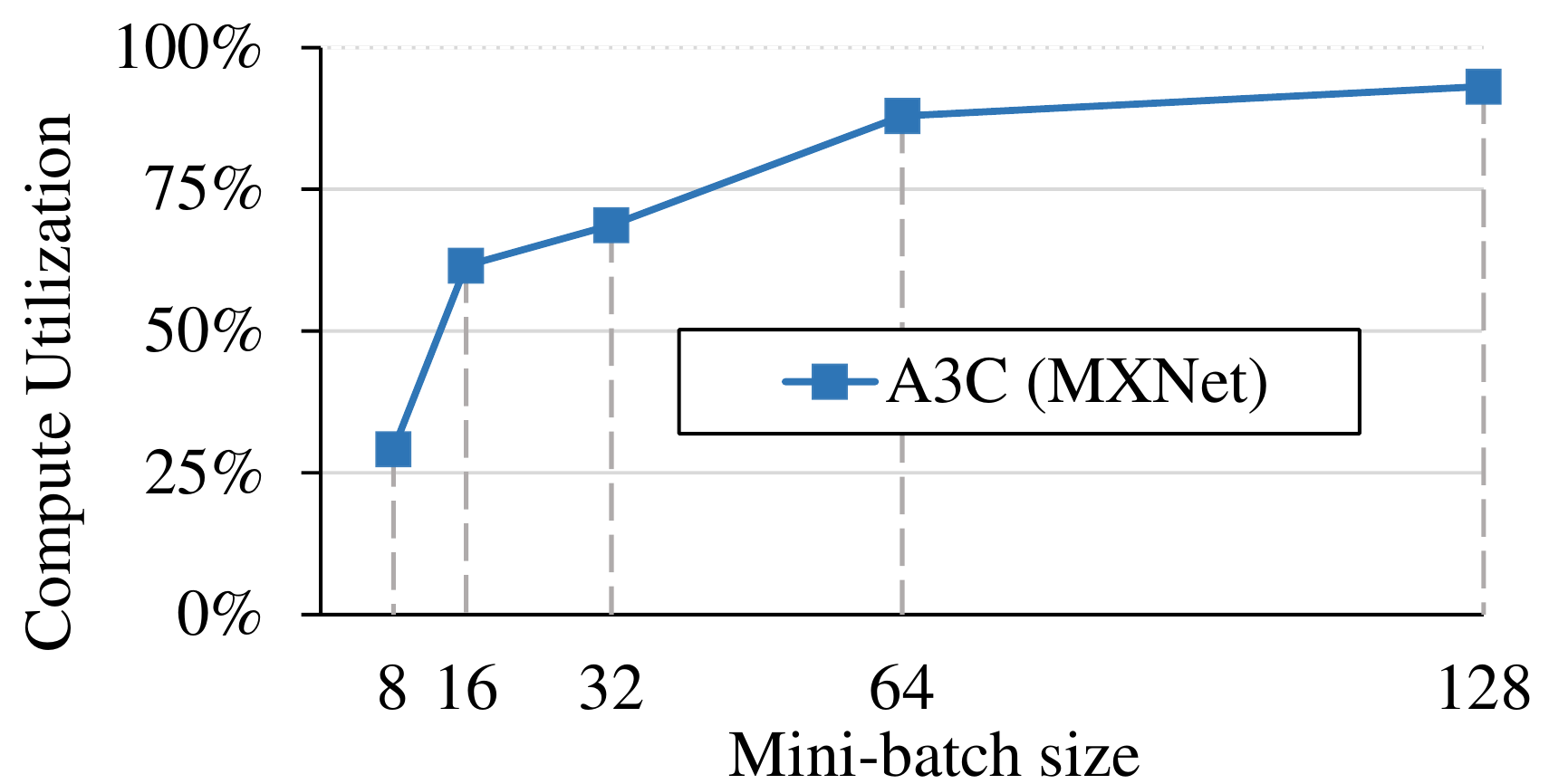}
        \caption{A3C}
        \label{fig:occupation_a3c}
    \end{subfigure}
    \caption{GPU compute utilization for different models on multiple mini-batch sizes.}
    \label{fig:occupation}
\vspace{-0.4cm}
\end{figure*}

\subsubsection{Throughput}

Figure~\ref{fig:throughput} shows the average training throughput for different
models from the \texttt{TBD} suite when varying the mini-batch size (the maximum mini-batch
size is limited by the GPU memory capacity).
For Faster R-CNN, the number of images processed per iteration is fixed to be
just one on a single GPU, hence we do not present a separate graph for Faster R-CNN.
Both TensorFlow and MXNet implementations achieve a throughput of 2.3 images per second for Faster R-CNN.
We make the following three observations from this figure.

\emph{Observation 1: Performance increases with the mini-batch size for all models.}
As we expected, the larger the
mini-batch size, the higher the throughput for all models we study.
% which makes
%mini-batch size one of the key hyper-parameters to achive high throughput.
We conclude that to achieve high training throughput on a single GPU, one should aim for a
reasonably high mini-batch size, especially for non-convolutions models.  We
explain this behavior as we analyze the GPU and FP32 utilization metrics
later in this section.

\emph{Observation 2: The performance of RNN-based models is not saturated within the
GPU's memory constraints.} The relative benefit of further increasing the mini-batch size
differs a lot between different applications. For example, for the
\emph{NMT} model
increasing mini-batch size from 64 to 128 increases training throughput by
25\%, and the training throughput of Deep Speech 2 scales almost linearly.
These two models' throughput (and hence performance) is essentially limited by
the GPU memory capacity and we do not see any saturation point for them while
increasing the mini-batch size.  In contrast, other models also benefit from
higher mini-batch size, but after certain \emph{saturation point} these benefits are
limited. For example, for the \emph{Inception-v3} model going from batch size of
16 to 32 has less than 10\% in throughput improvement for implementations on all three
frameworks.

%there is
%quite some room for the throughput to increase with larger mini-batch size
%(making these two models memory-bound), while the training throughput of other
%models is basically saturated within the permit of Quadro P4000 8GB memory
%TODO: we can't claim they are compute-bound, they can be memory BW bound or
% or something else -> (making them compute-bound).

\emph{Observation 3:  Application diversity is important when comparing
performance of different frameworks.} We find that the results when comparing performance of models on different frameworks
can greatly vary for different applications, and hence using a diverse set of applications in any comparisons of frameworks
is important.
For example, we observe that for image classification the \emph{MXNet} implementations of both models
(\emph{ResNet-50} and \emph{Inception-v3}) perform generally better
than the corresponding \emph{TensorFlow} implementations, but
at the same time, for machine translation the \emph{TensorFlow} implementation of \emph{Seq2Seq} (\emph{NMT}) performs
significalty better than its \emph{MXNet} counterpart (\emph{Sockeye}.
TensorFlow also utilizes the GPU memory better than MXNet for Seq2Seq models so that it can be trained
with a maximum mini-batch size of 128, while MXNet can only be trained with a maximum mini-batch of 64 (both limited by 8GB GPU memory).
For the same memory budget, it allows TensorFlow achieve higher throughput, 365 samples per second, vs.
229 samples per second for MXNet.
We conclude that there is indeed a signficant diversity on how different frameworks perform on different models,
making it extremely important to study a \emph{diverse} set of applications (and models) as we propose
in our benchmark pool.

\subsubsection{GPU Compute Utilization}

Figure~\ref{fig:occupation} shows the GPU compute utilization, the amount of time GPU is busy
running some kernels (as formally defined
by~\ref{formula:occupation} in Section~\ref{sec:method}) for different benchmarks
as we change the mini-batch size. Again, for \emph{Faster R-CNN}, only batch of one
is possible, and TensorFlow implementation achieves a relatively high compute utilization of
89.4\% and the \emph{MXNet} implementation achieves 90.3\%.
We make the following two observations from this figure.

\emph{Observation 4: The mini-batch size should be large enough to keep the GPU
busy.} Similar to our observation 1 about throughput, the larger the mini-batch
size, the longer the duration of individual GPU kernel functions
and the better the GPU compute utilization, as the GPU
spends more time doing computations rather than invoking and finishing small
kernels.
While large mini-batch sizes also increase the overhead of data transfers,
our results show that this overhead is usually efficiently parallelized with the computation.

\emph{Observation 5: The GPU compute utilization is low for LSTM-based models.}
Non-RNN models and \emph{Deep Speech 2} that uses regular RNN cells (not LSTM)
usually reach very high utilization with large batches, around 95\% or higher.
Unfortunately, LSTM-based models (\emph{NMT}, \emph{Sockeye})
cannot drive up GPU utilization significantly, even with maximim mini-batch sizes.
This means that, in general, these models do not utilize the available GPU hardware
resources well, and further research should be done in how to optimize LSTM cells
on GPUs. Moreover, it is important to notice that the low compute utilization problem is specific
to the layer type, but not the application -- the \emph{Transformer} model also used
in machine translation does not suffer from low compute utilization as it uses different (non-RNN)
layer called \emph{Attention}.
%, resulting in
%bad throughput saturation.
%In general, the compute-bound models have GPU
%occupation close to or around 95\%.

\subsubsection{GPU FP32 utilization}
\begin{figure*}
    \centering
    \begin{subfigure}[t]{0.23\textwidth}
        \centering
        \includegraphics[width=\textwidth]{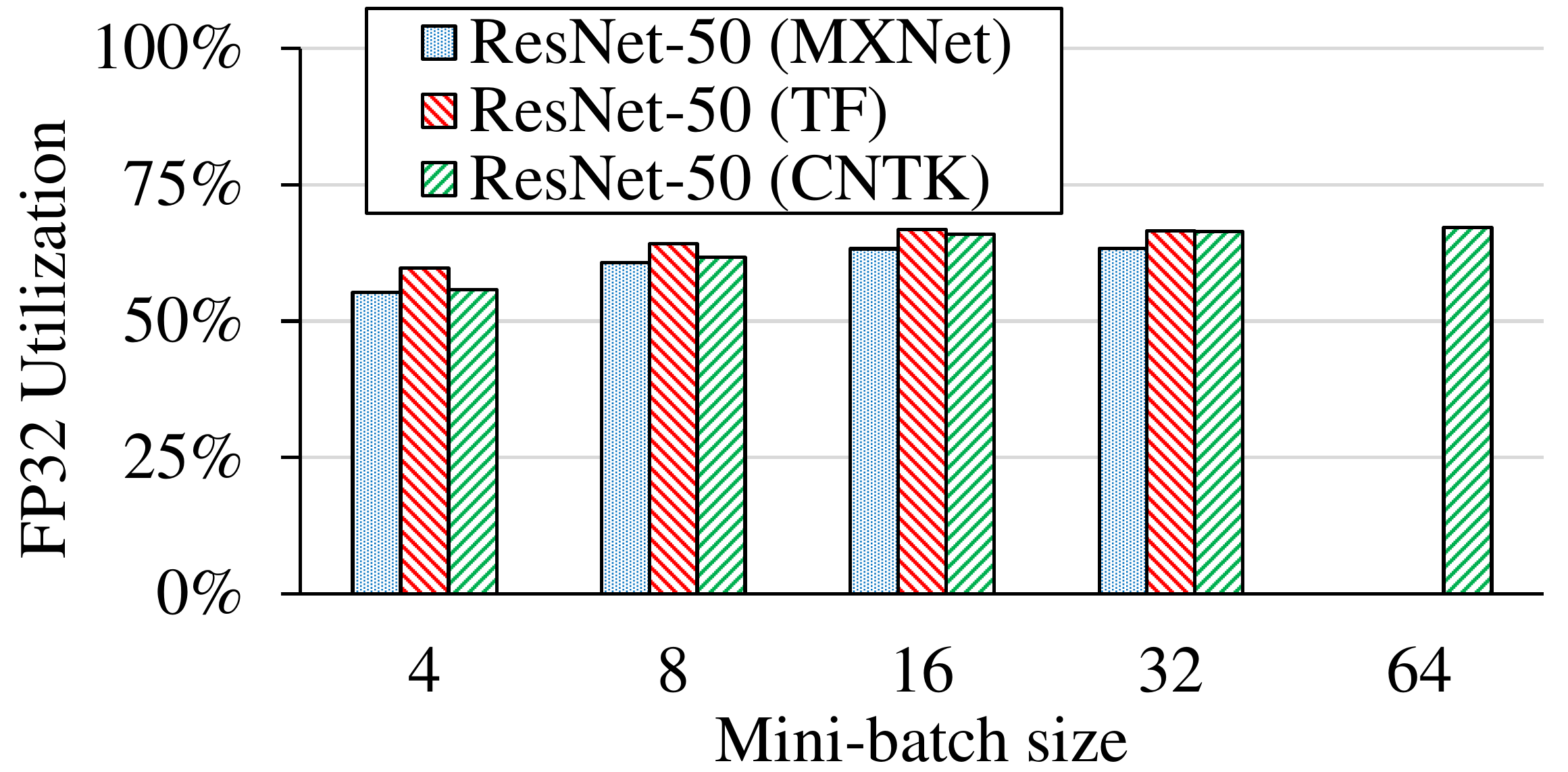}
        \caption{ResNet-50}
        \label{fig:utilization_resnet}
    \end{subfigure}%
    \hspace{0.1cm}
    \begin{subfigure}[t]{0.23\textwidth}
        \centering
        \includegraphics[width=\textwidth]{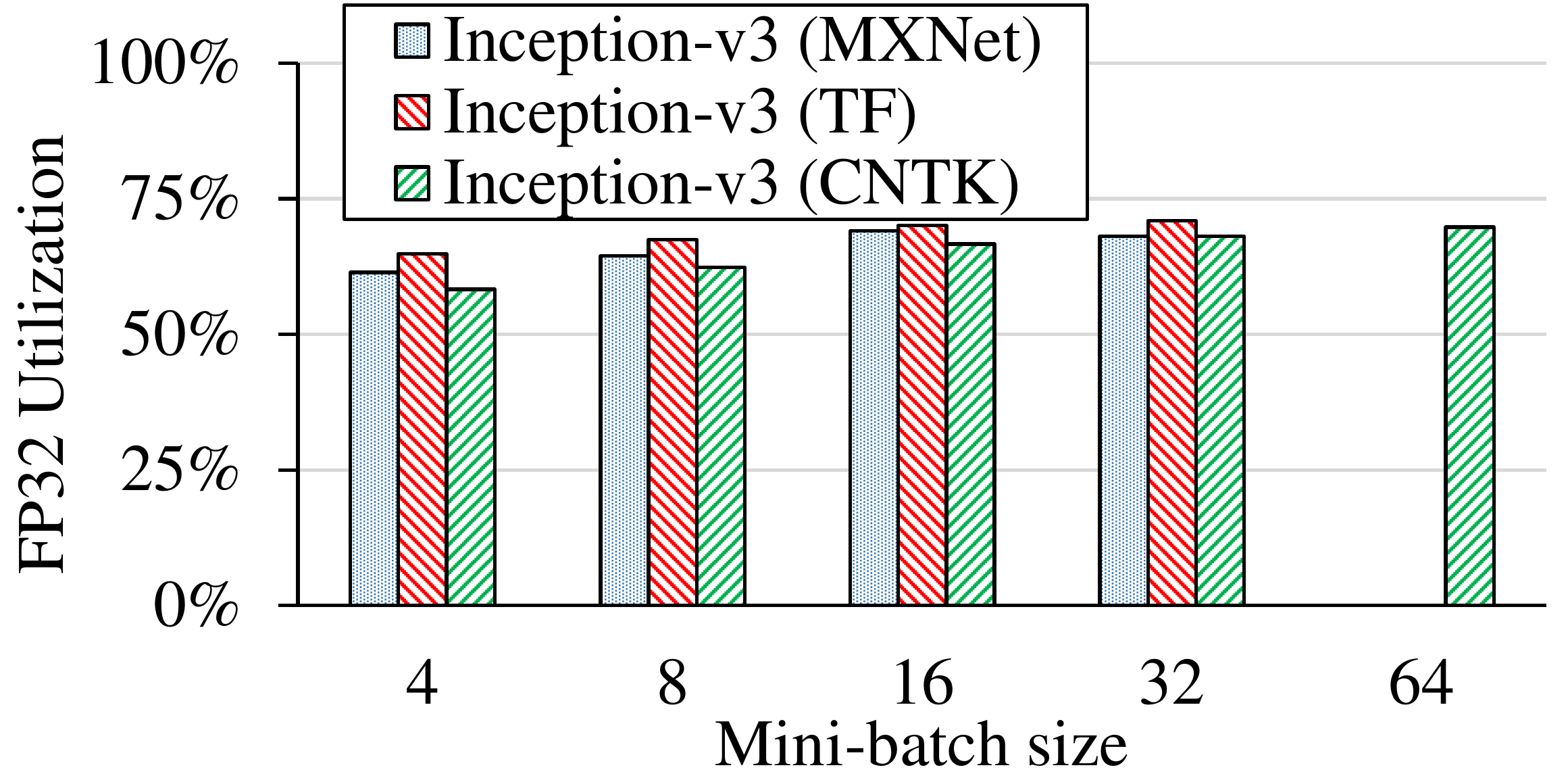}
        \caption{Inception-v3}
        \label{fig:utilization_inception}
    \end{subfigure}%
    \hspace{0.1cm}
    \begin{subfigure}[t]{0.23\textwidth}
        \centering
        \includegraphics[width=\textwidth]{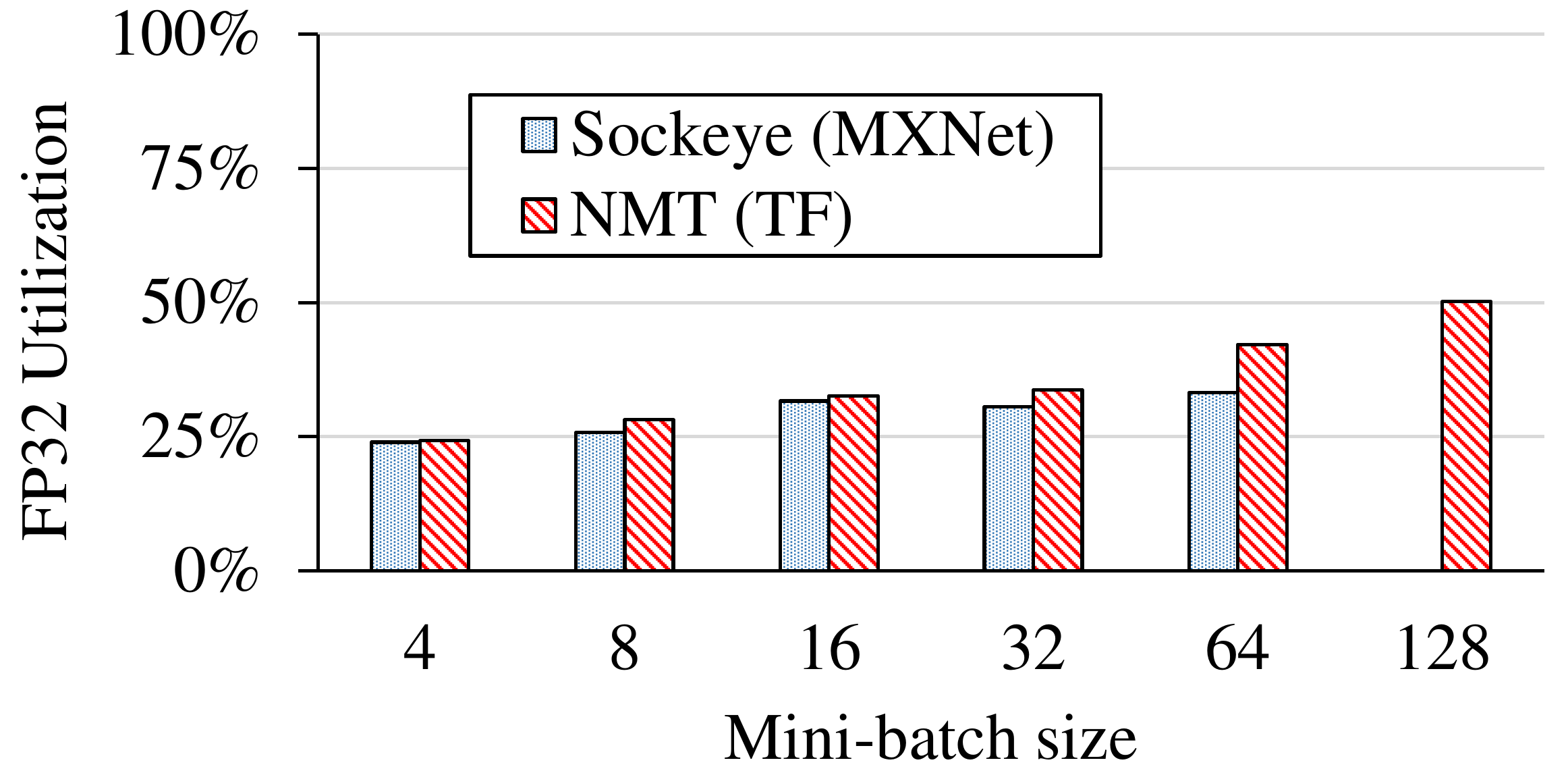}
        \caption{Seq2Seq}
        \label{fig:utilization_seq2seq}
    \end{subfigure}
    \hspace{0.1cm}
    \begin{subfigure}[t]{0.23\textwidth}
        \centering
        \includegraphics[width=\textwidth]{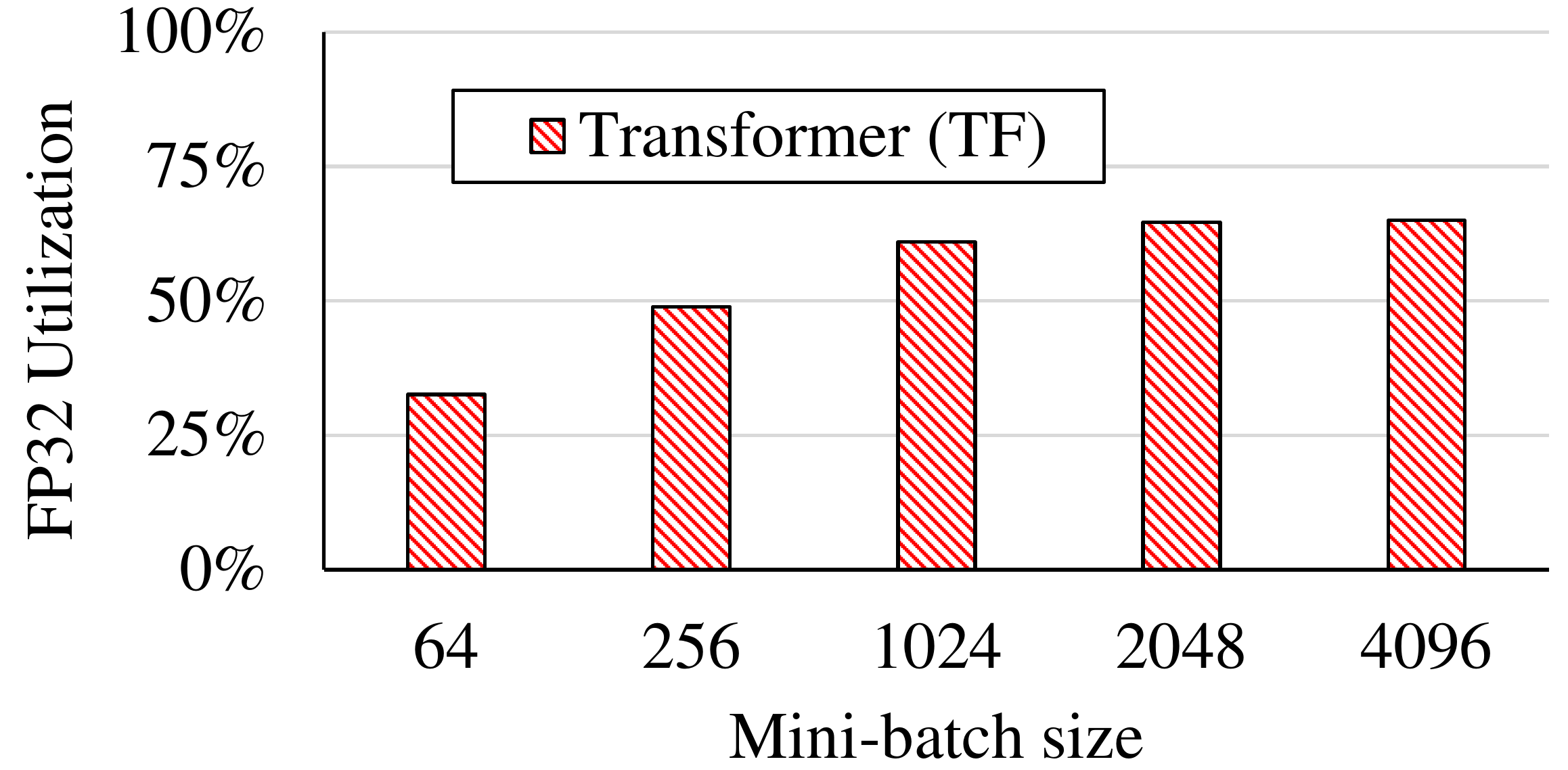}
        \caption{Transformer}
        \label{fig:utilization_transformer}
    \end{subfigure}

    \begin{subfigure}[t]{0.23\textwidth}
        \centering
        \includegraphics[width=\textwidth]{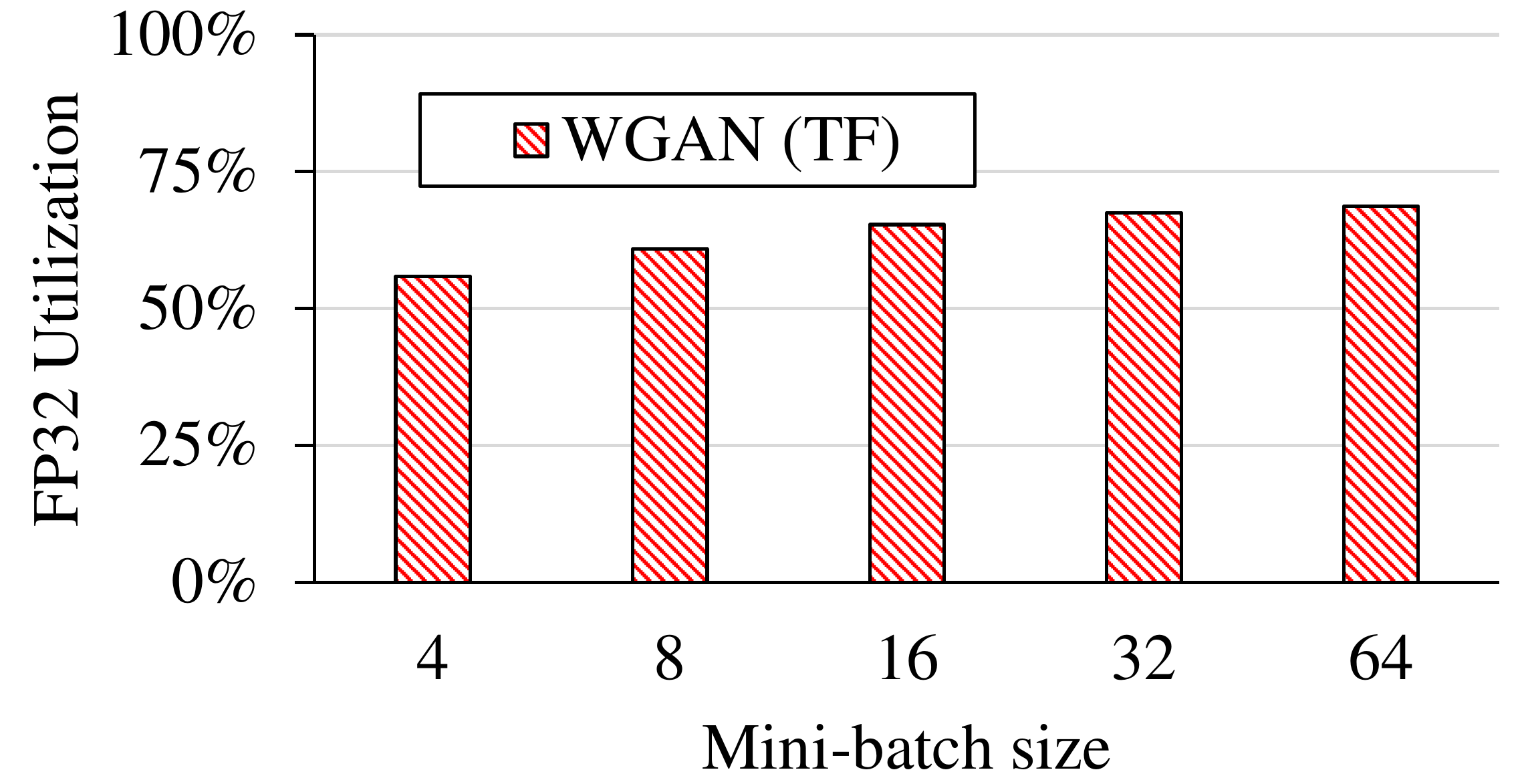}
        \caption{WGAN}
        \label{fig:utilization_wgan}
    \end{subfigure}
    \hspace{0.1cm}
    \begin{subfigure}[t]{0.23\textwidth}
        \centering
        \includegraphics[width=\textwidth]{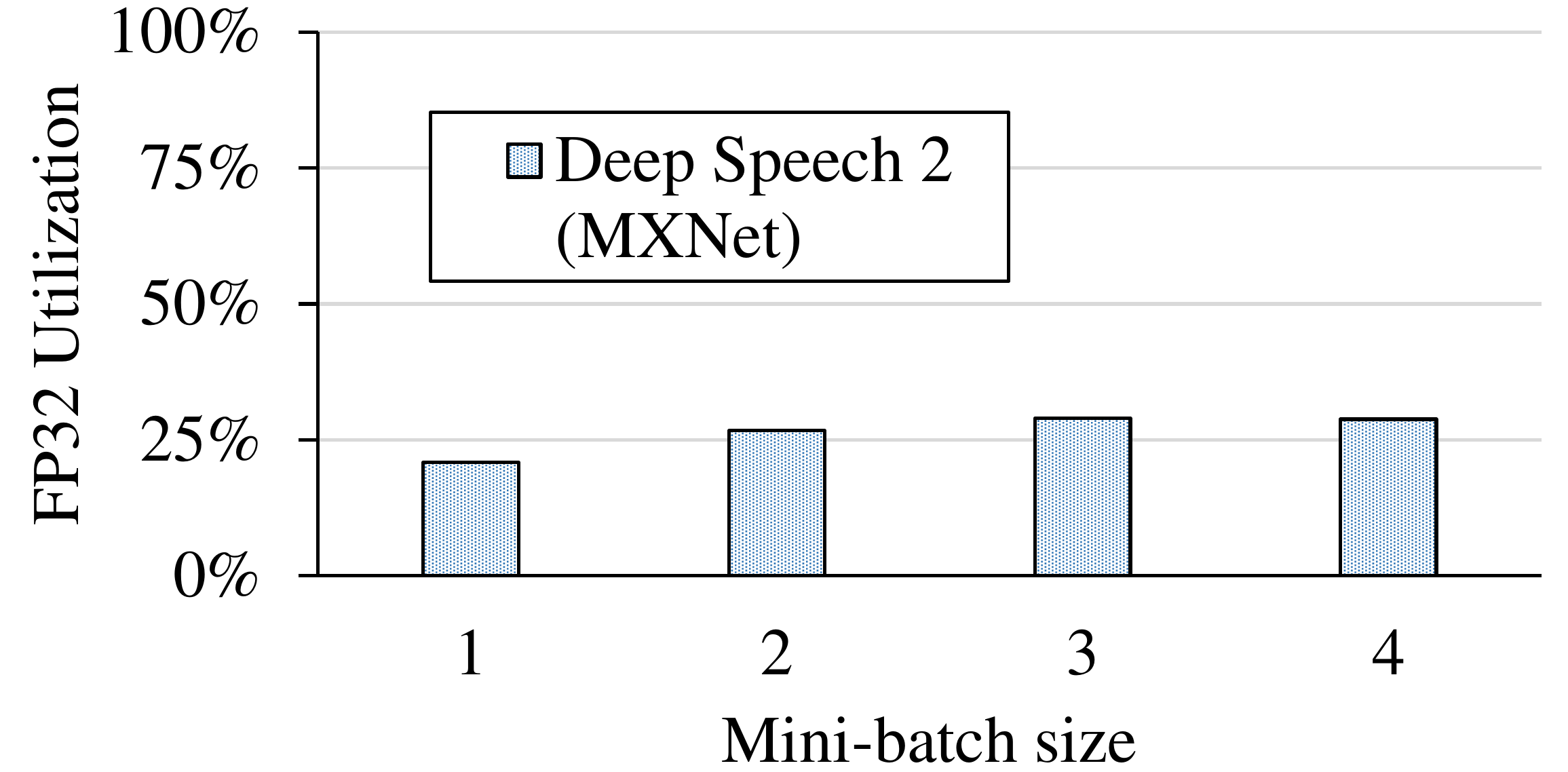}
        \caption{Deep Speech 2}
        \label{fig:utilization_ds2}
    \end{subfigure}
    \hspace{0.1cm}
    \begin{subfigure}[t]{0.23\textwidth}
        \centering
        \includegraphics[width=\textwidth]{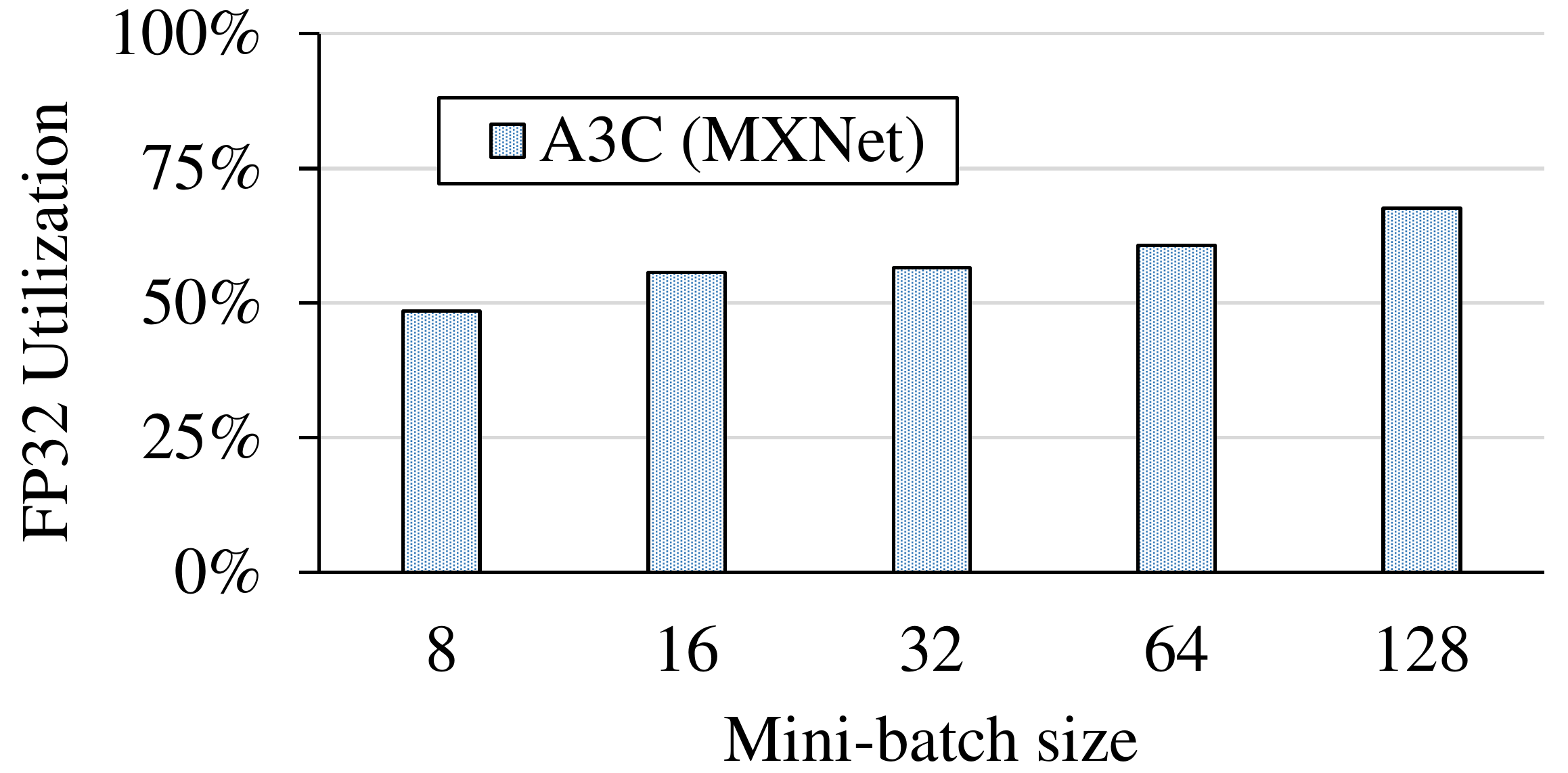}
        \caption{A3C}
        \label{fig:utilization_a3c}
    \end{subfigure}
    \caption{GPU FP32 utilization for different models on multiple mini-batch sizes.}
    \label{fig:utilization}
\end{figure*}

Figure~\ref{fig:utilization} shows the GPU FP32 utilization (formally defined
by~\ref{formula:utilization} in Section~\ref{sec:method})
for different benchmarks as we change the
mini-batch size (until memory capacity permits). For Faster R-CNN, the
MXNet/TensforFlow implementations achieve an average utilization of 70.9\%/58.9\%
correspondingly. We make three major observations from this figure.

\emph{Observation 6: The mini-batch size should be large enough to exploit the
FP32 computational power of GPU cores.} As expected, we observe that large mini-batch
sizes also improve GPU FP32 utilization for all benchmarks we study.
We conclude that both  the improved FP32
utilization (Observation 6) and GPU utilization (Observation 4) are key contributors
to the increases in overall throughput with the mini-batch size (Observation 1).

\emph{Observation 7: RNN-based models have low GPU FP32 utilization.}
Even with the maximum mini-batch size possible (on a single GPU),
the GPU FP32 utilization of the two RNN-based models (\emph{Seq2Seq}
and \emph{Deep Speech 2}, Figure~\ref{fig:utilization_seq2seq} and Figure~\ref{fig:utilization_ds2},
respectively) are much lower than for other non-RNN models. This clearly
indicates the potential of designing more efficient RNN layer implementations
used in TensforFlow and MXNet, and we believe further research should be
done to understand the sources of these inefficiences.
Together with Observation 5 (low GPU utilization for LSTM-based models) this observation
explains why in Observation 2 we do not observe throughput saturation for RNN-based
models even for very large mini-batches.
% and showing that there is great
%potential in better utilizing GPU cores.

\emph{Observation 8: There exists kernels with long duration, but low
FP32 utilization even for highly optimized models.} The previous observation might have
brought up the question why average FP32 utilizations are so low,
even for extremely optimized CNN models.
In this observation we provide an answer:
Different kernels vary greatly in their FP32 utilizations, and even optimized models
have long-running kernels with low utilization.
%which have been highly optimized, there is
%still more than 30\% GPU computation power remained unused.
Table~\ref{table:util-tf-below} and Table~\ref{table:util-mx-below} show the five most
important kernels with the FP32 utilization \emph{lower than average} (for \emph{ResNet-50}
model on TensorFlow and MXNet). We observe that the cuDNN batch normalization kernels
(have \emph{bn} part in their names) are the major source of inefficiency with FP32 utilizations
more than 20\% below the average. Note that this observation is true for implementations
on different frameworks.
% are the
%most important kernels that under-utilize the GPU cores.
If we want to get further progress in improving DNN training performance on GPUs, these
kernels are the top candidates for acceleration.
%There are also a lot of small kernels with low utilization, slowing down the entire training
%process.

\begin{table}
\centering
\footnotesize{
    \begin{tabular}{ |c|c|c| }
        \hline
        Duration & Utilization & Kernel Name \\ \hline
        8.36\% & 30.0\% & magma\_lds128\_sgemm\_kernel... \\ \hline
        5.53\% & 42.3\% & cudnn::detail::bn\_bw\_1C11\_kernel\_new... \\ \hline
        4.65\% & 46.3\% & cudnn::detail::bn\_fw\_tr\_1C11\_kernel\_new... \\ \hline
        3.12\% & 20.0\% & Eigen::internal::EigenMetaKernel... \\ \hline
        2.48\% & 40.0\% & tensorflow::BiasNHWCKernel... \\ \hline
    \end{tabular}
    \caption{Longest 5 kernels with utilization level below the average (ResNet-50, mini-batch size 32, TensorFlow)}
    \label{table:util-tf-below}
\vspace{-0.5cm}
}
\end{table}

\begin{table}
\centering
\footnotesize{
    \begin{tabular}{ |c|c|c| }
        \hline
        Duration & Utilization & Kernel Name \\ \hline
        9.43\% & 30.0\% & cudnn::detail::bn\_bw\_1C11\_kernel\_new... \\ \hline
        7.96\% & 42.3\% & cudnn::detail::bn\_fw\_tr\_1C11\_kernel\_new... \\ \hline
        5.14\% & 46.3\% & cudnn::detail::activation\_bw\_4d\_kernel... \\ \hline
        3.52\% & 20.0\% & cudnn::detail::activation\_fw\_4d\_kernel... \\ \hline
        2.85\% & 40.0\% & \_ZN5mxnet2op8mxnet\_op20mxnet\_generic\_kernel... \\ \hline
    \end{tabular}
    \caption{Longest 5 kernels with FP32 utilization  below the average (ResNet-50, mini-batch size 32, MXNet)}
    \label{table:util-mx-below}
}
\vspace{-0.5cm}
\end{table}

\subsubsection{CPU Utilization}
\begin{figure*} [h!]
    \centering
    \includegraphics[width=0.9\textwidth]{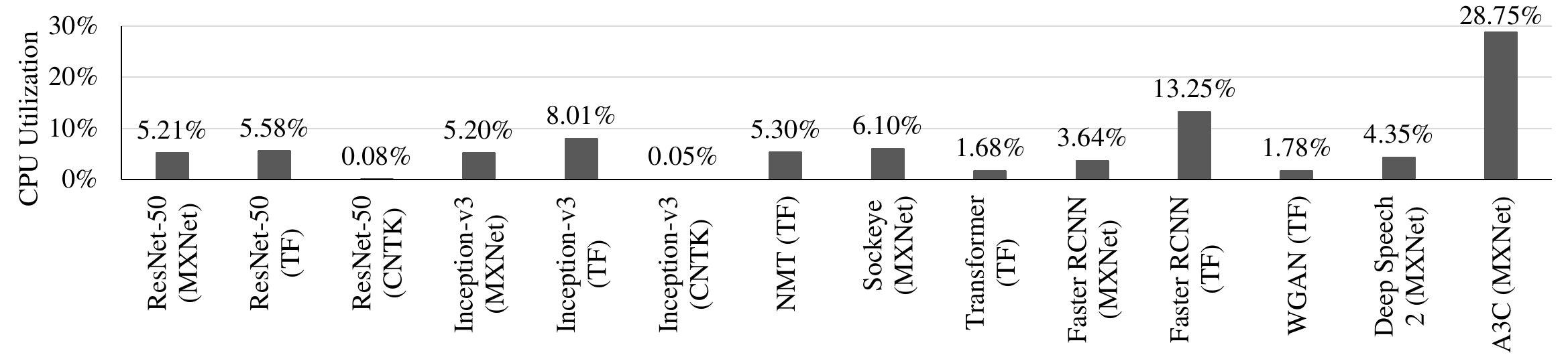}
    \caption{Average CPU utilization for different models.}
    \label{fig:cpu-utilization}
\end{figure*}

The results of our analysis of CPU utilization are presented in Figure~\ref{fig:cpu-utilization}.

\emph{Observation 9: CPU utilization is low in DNN training.} We observe that
for all our models the CPU utilization is very low, less than 15\% for all but one model,
less than 8\% for all but two models. For our CPU machines with 28 cores, this means
that on average less than 2 cores are usually busy. We believe that future
research should look into how to make CPUs more useful for DNN training. For example,
they can be used to compute layers that cannot benefit from the massive GPU compute
power, such as batch normalization.

\subsection{Hardware Sensitivity} \label{sec:result-hardware}
The results presented so far,  were based on experiments with the Quadro P4000 GPU.
In this section we are interested in seeing how the performance of DNN training will depend
on the hardware used. Toward this end we compare the training throughput, GPU
utilization, and FP32 utilization for several of our models: \emph{ResNet-50}, \emph{Inception-v3}, and
\emph{Seq2Seq} on P4000 GPU and the more powerful Titan Xp GPU.
For throughput comparison, we
normalized each model result to the throughput of less powerfull P4000 card. We make
the following observation from this figure.

\emph{Observation 10: More advanced GPUs should be accompanied by better systems
designs and more efficient low-level libraries.} Titan Xp usually helps improving the training throughput
(except for \emph{Sockeye}), however the computation power of
Titan Xp is not well-utilized. Both the GPU and the FP32 utilizations of Titan Xp
appear to be worse than those of P4000. Hence we conclude that although Titan Xp is more computationally powerful
(more multiprocessors, CUDA cores, and bandwidth, see Table~\ref{table:GPU-spec}),
the proper utilization of these resources requires a more careful design of existing
GPU kernel functions, libraries (e.g., cuDNN), and algorithms that can efficiently exploit
these resources.

\begin{figure} [h!]
    \centering
    \begin{subfigure}[t]{0.37\textwidth}
        \centering
        \includegraphics[width=\textwidth]{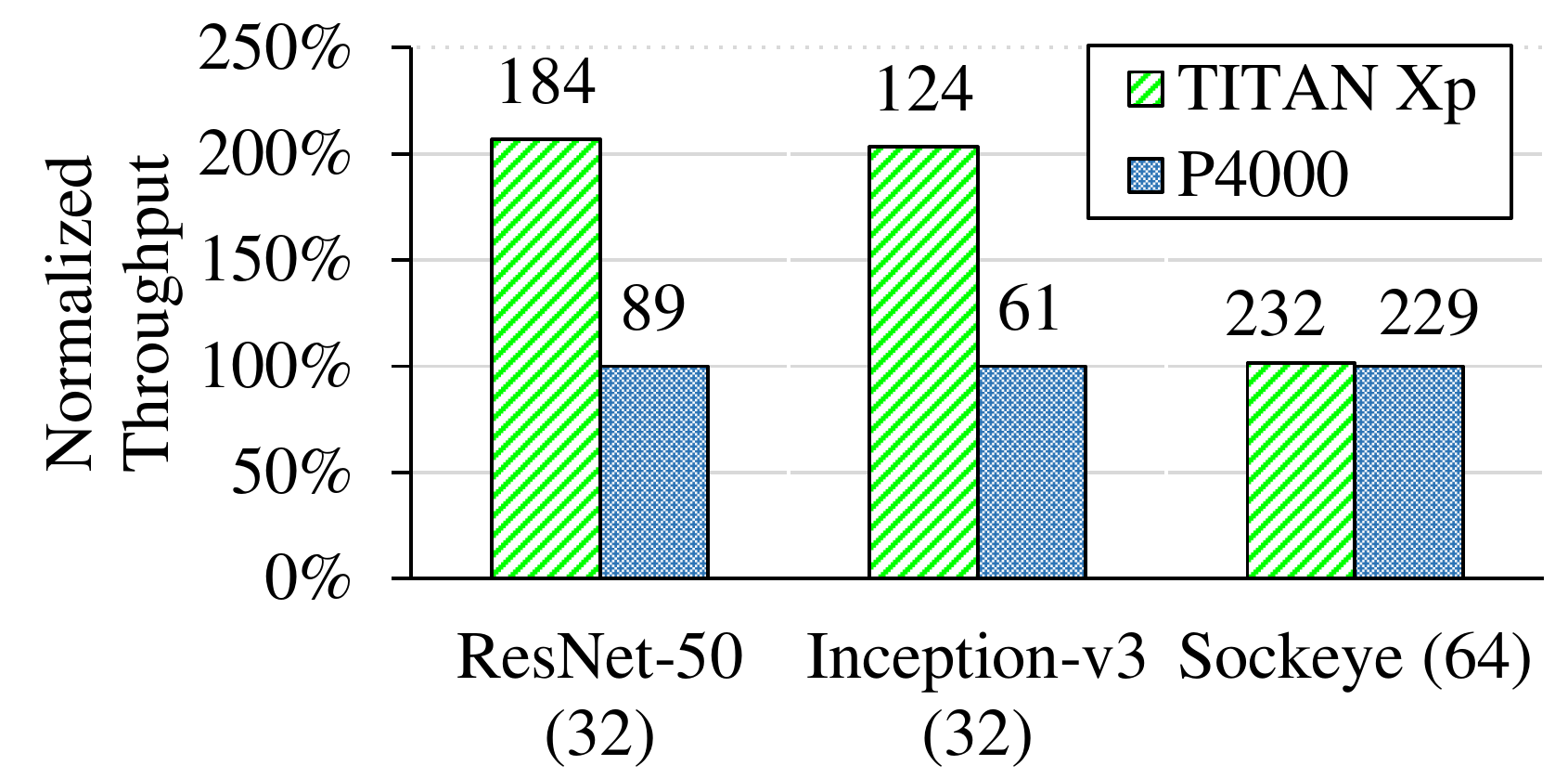}
        \caption{MXNet}
        \label{fig:throughput-compare-mx}
    \end{subfigure}%
    \hspace{0.1cm}
    \begin{subfigure}[t]{0.37\textwidth}
        \centering
        \includegraphics[width=\textwidth]{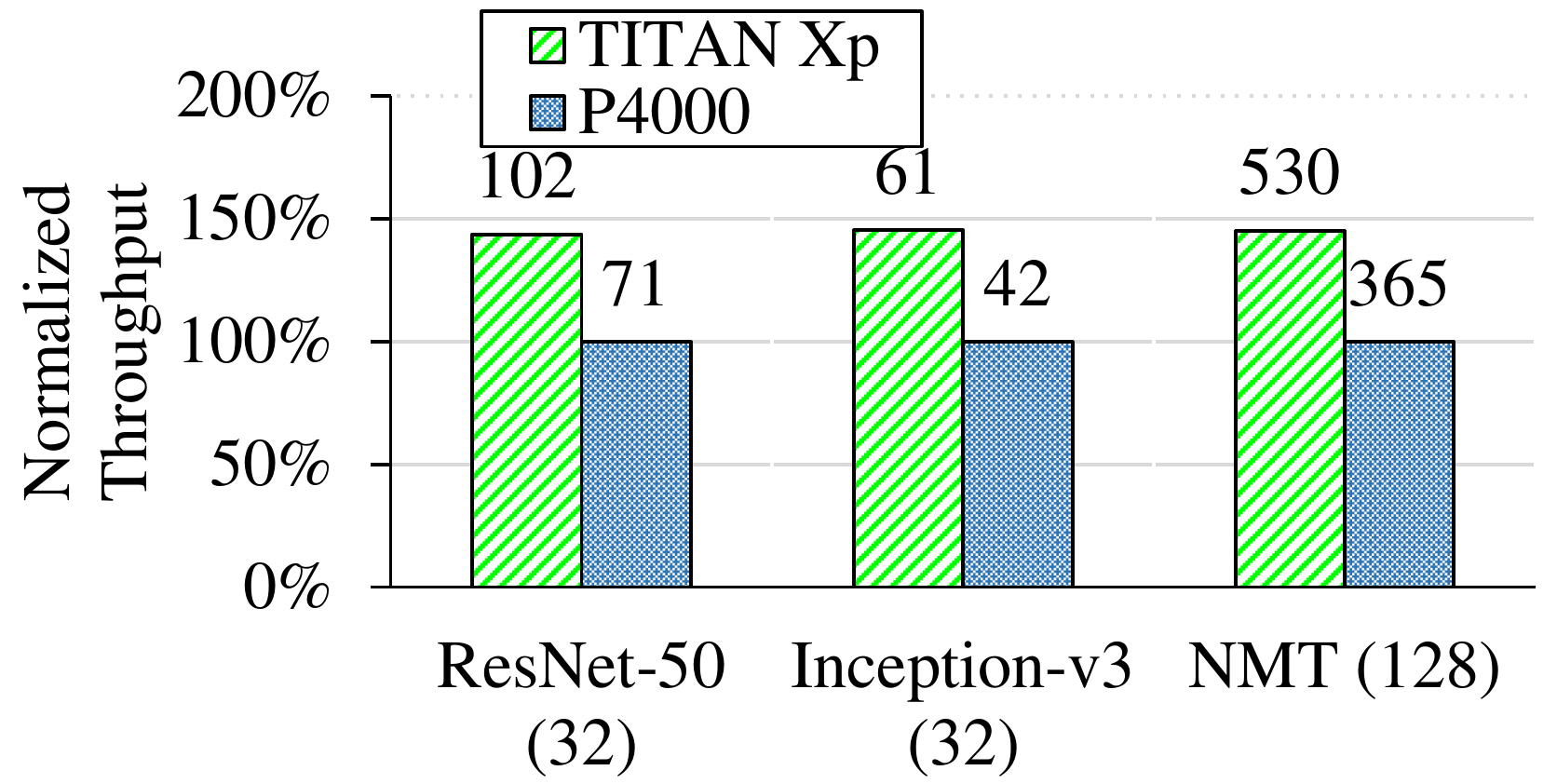}
        \caption{TensorFlow}
        \label{fig:throughput-compare-tf}
    \end{subfigure}%

    \begin{subfigure}[t]{0.37\textwidth}
        \centering
        \includegraphics[width=\textwidth]{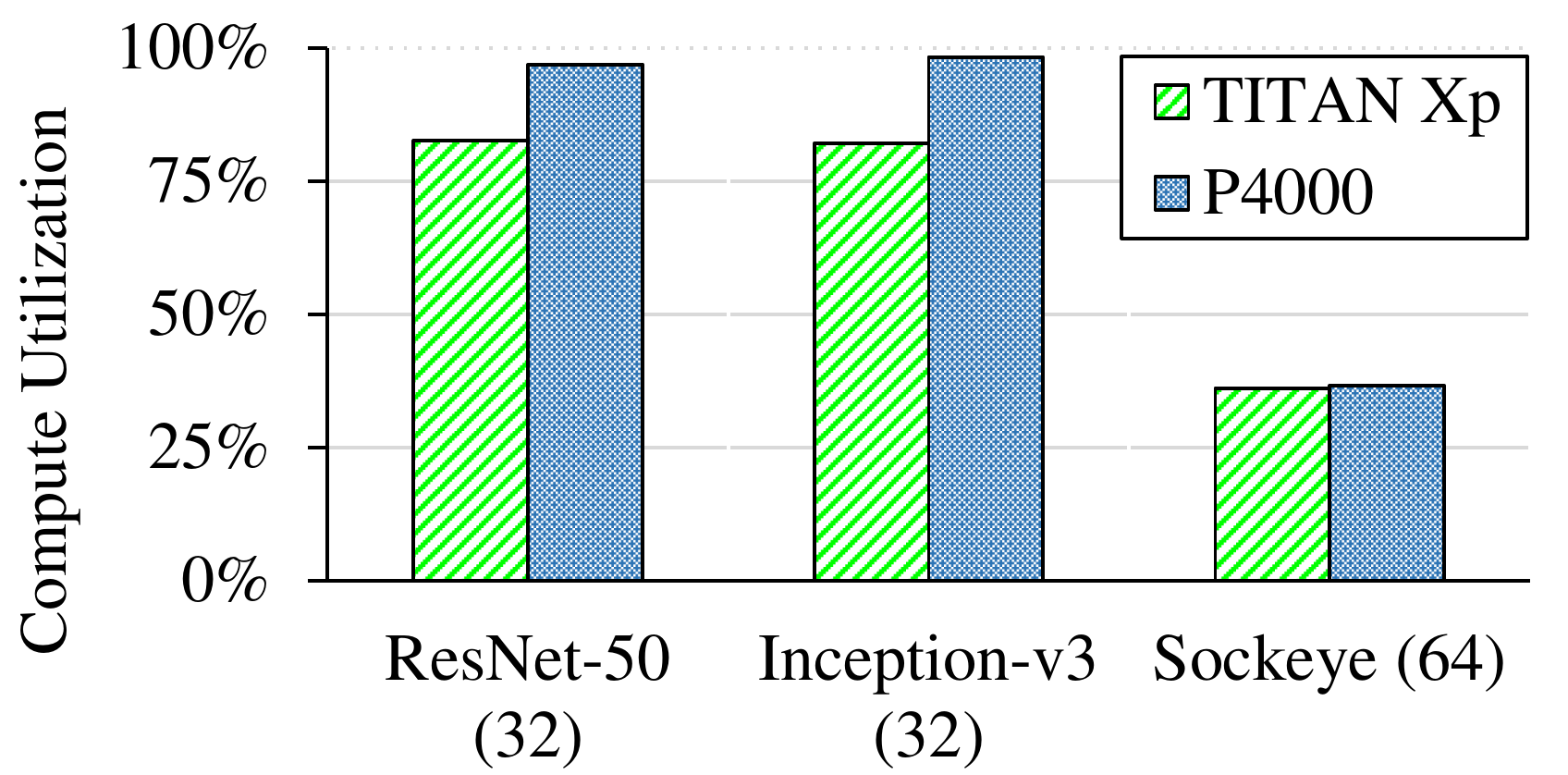}
        \caption{MXNet}
        \label{fig:occupation-compare-mx}
    \end{subfigure}%
    \hspace{0.1cm}
    \begin{subfigure}[t]{0.37\textwidth}
        \centering
        \includegraphics[width=\textwidth]{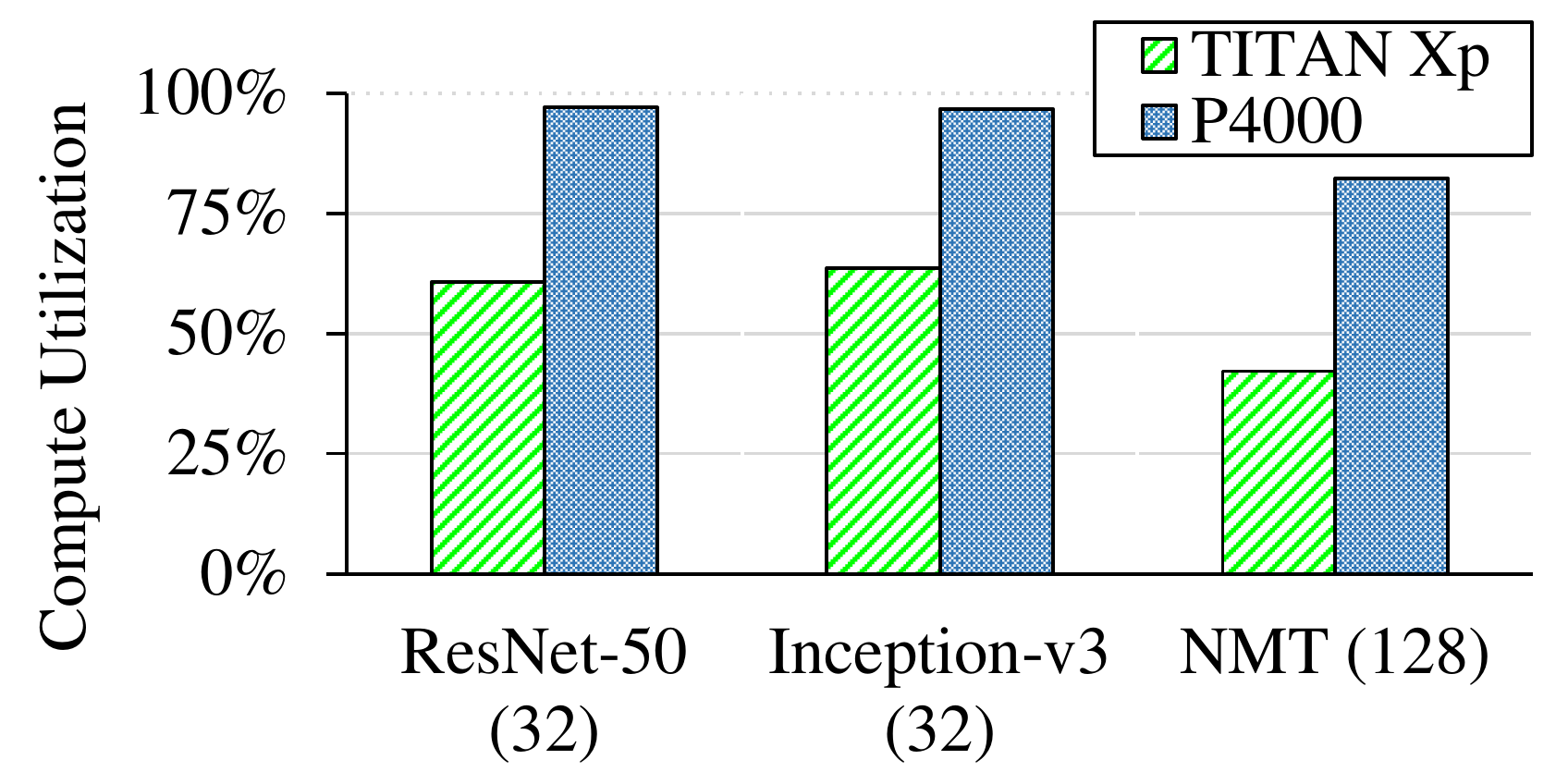}
        \caption{TensorFlow}
        \label{fig:occupation-compare-tf}
    \end{subfigure}

    \begin{subfigure}[t]{0.37\textwidth}
        \centering
        \includegraphics[width=\textwidth]{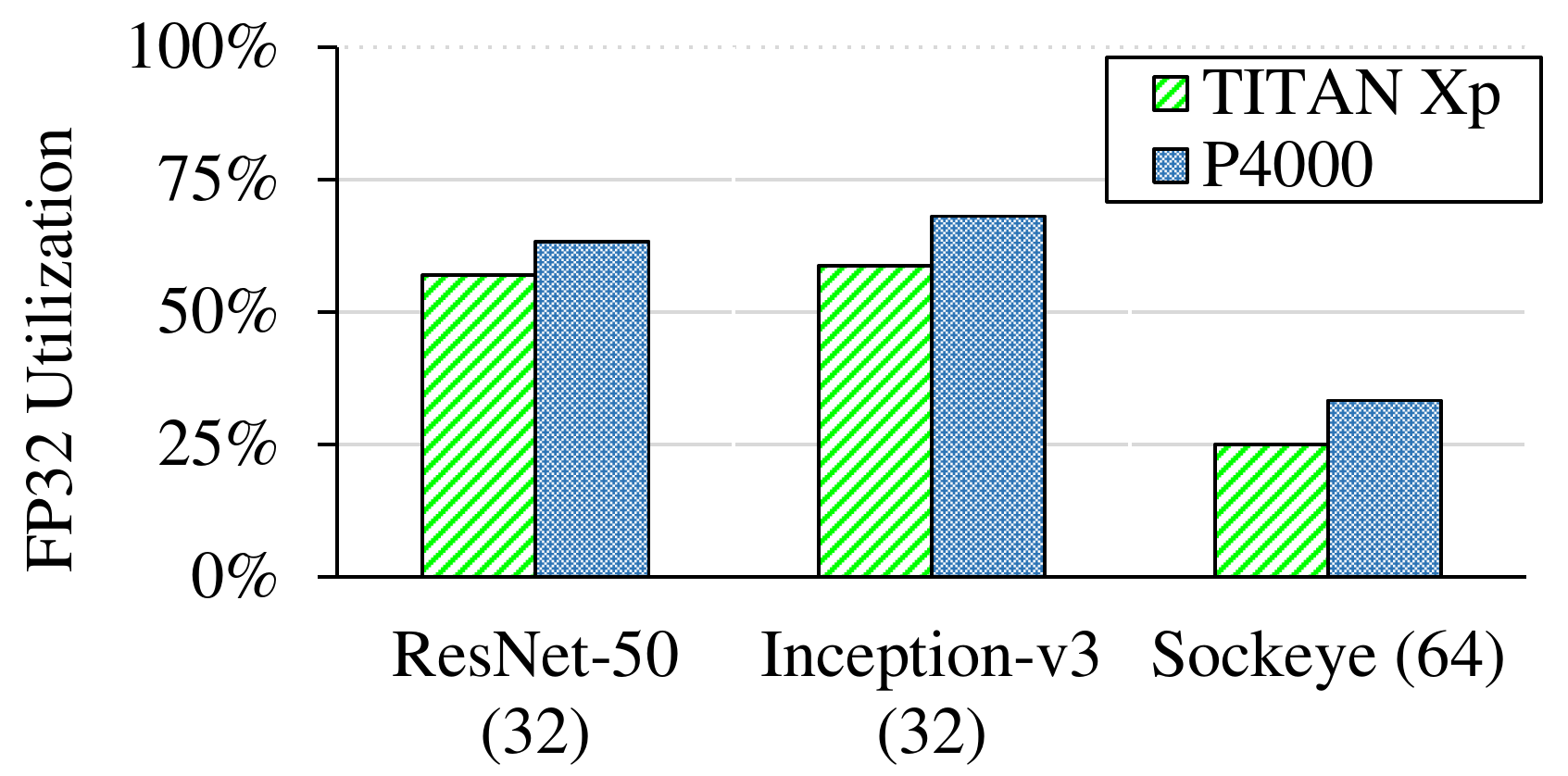}
        \caption{MXNet}
        \label{fig:utilization-compare-mx}
    \end{subfigure}%
    \hspace{0.1cm}
    \begin{subfigure}[t]{0.37\textwidth}
        \centering
        \includegraphics[width=\textwidth]{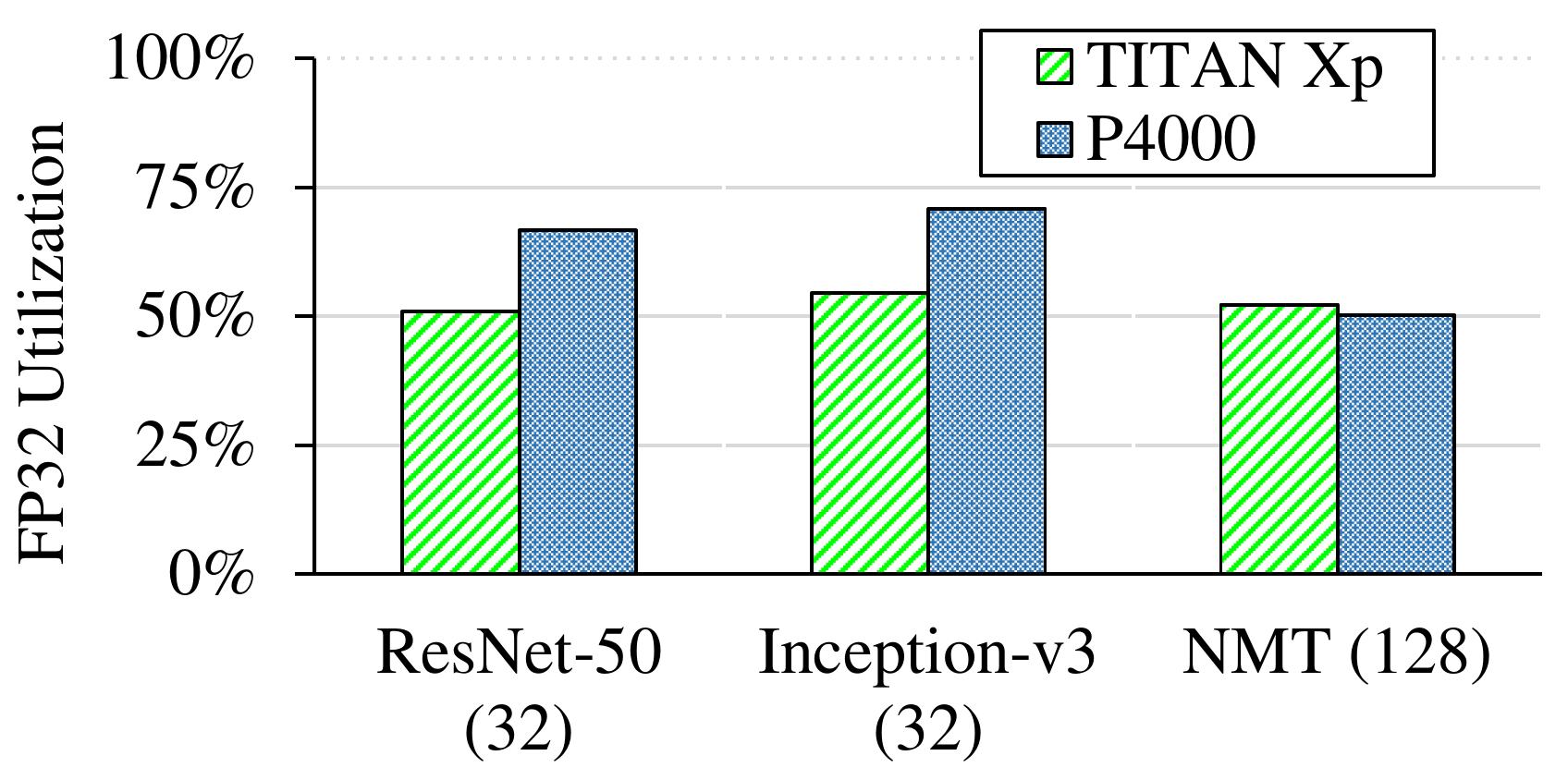}
        \caption{TensorFlow}
        \label{fig:utilization-compare-tf}
    \end{subfigure}
    \caption{Throughput, Compute Utilization, FP32 Utilization comparison between P4000 and Titan Xp for different benchmarks.}
    \label{fig:compare}
\vspace{-0.4cm}
\end{figure}

\subsection{Memory Profiling}

\begin{figure*} [h!]
    \centering
    \begin{subfigure}[t]{0.54\textwidth}
        \centering
        \includegraphics[width=\textwidth]{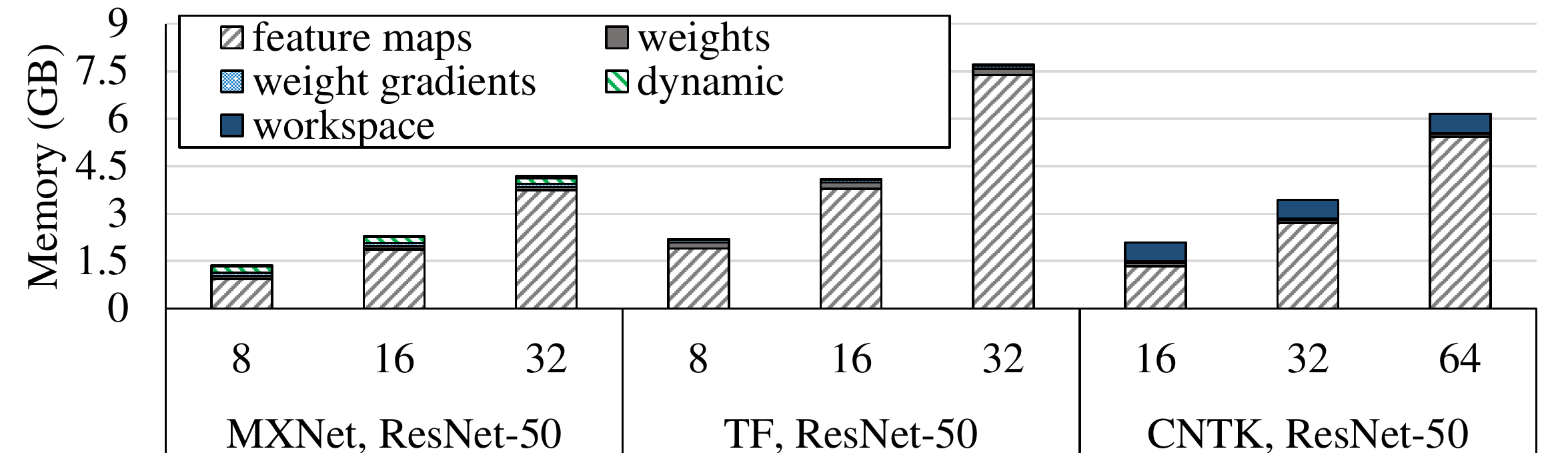}
        \caption{ResNet-50}
        \label{fig:memory_resnet}
    \end{subfigure}%
    \hspace{0.1cm}
    \begin{subfigure}[t]{0.40\textwidth}
        \centering
        \includegraphics[width=\textwidth]{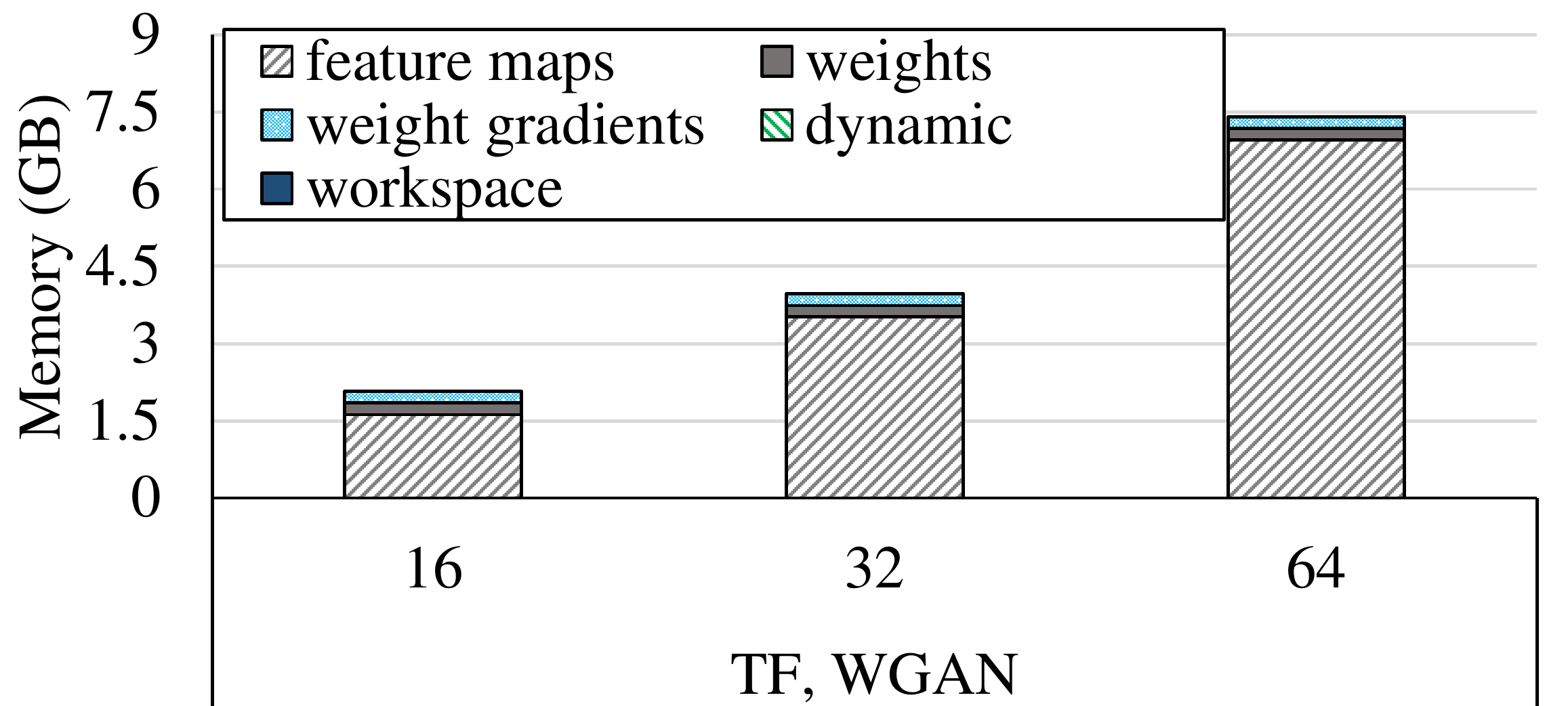}
        \caption{WGAN}
        \label{fig:memory_wgan}
    \end{subfigure}

    \begin{subfigure}[t]{0.54\textwidth}
        \centering
        \includegraphics[width=\textwidth]{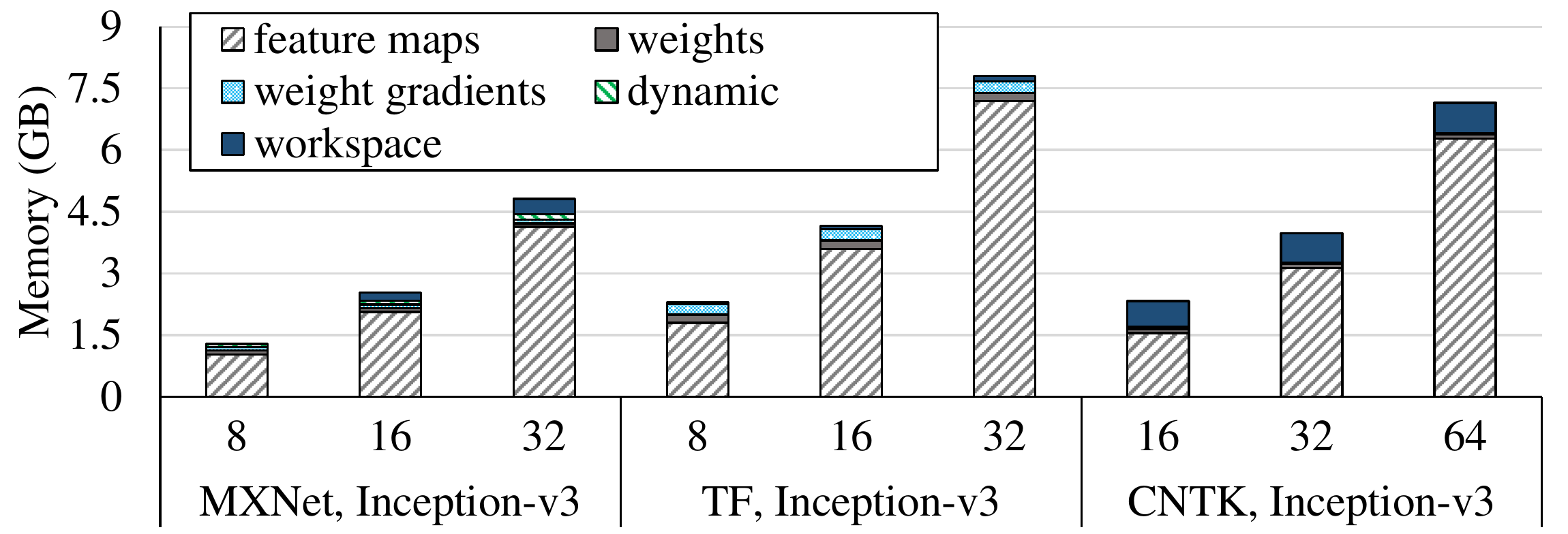}
        \caption{Inception-v3}
        \label{fig:memory_inception}
    \end{subfigure}%
    \hspace{0.1cm}
    \begin{subfigure}[t]{0.4\textwidth}
        \centering
        \includegraphics[width=\textwidth]{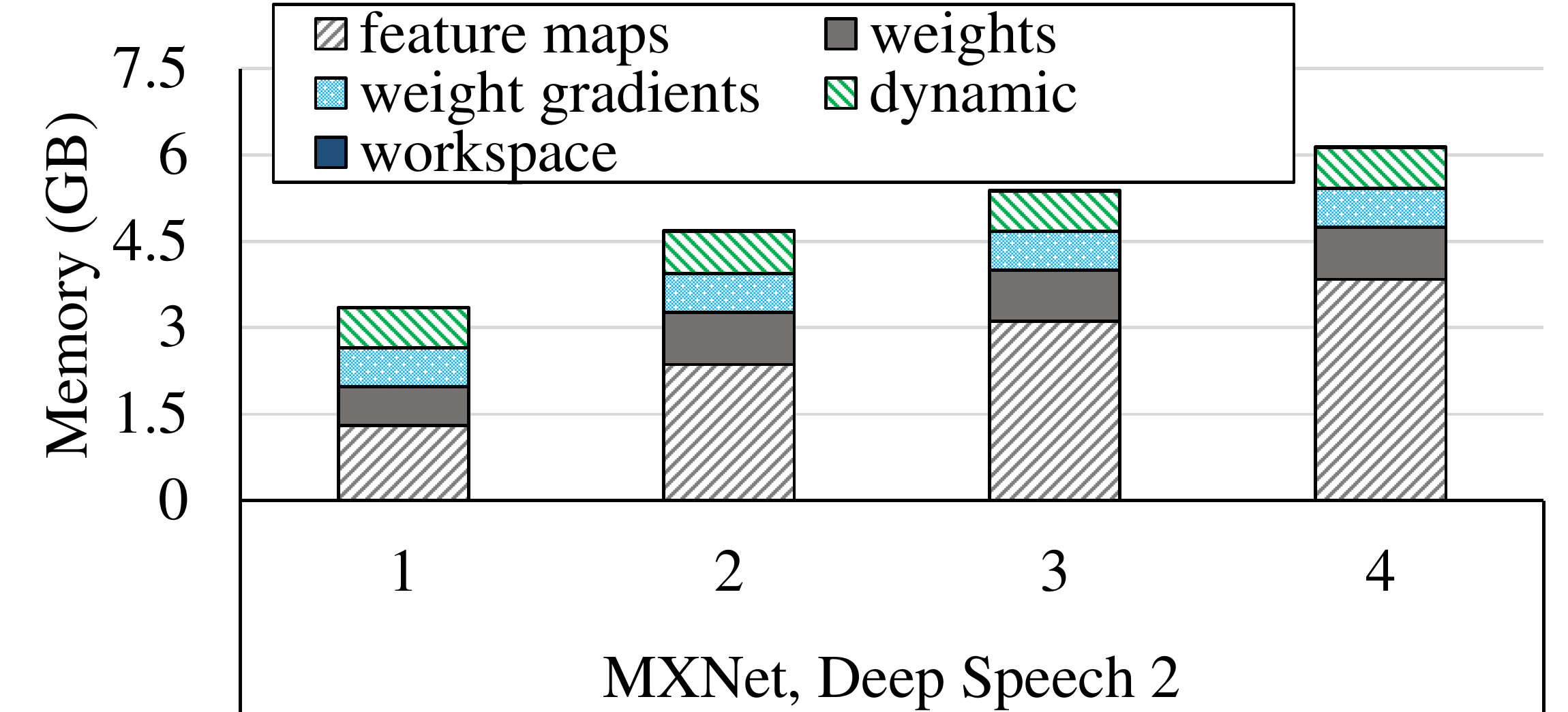}
        \caption{Deep Speech 2}
        \label{fig:memory_seq2seq}
    \end{subfigure}

    \begin{subfigure}[t]{0.32\textwidth}
        \centering
        \includegraphics[width=\textwidth]{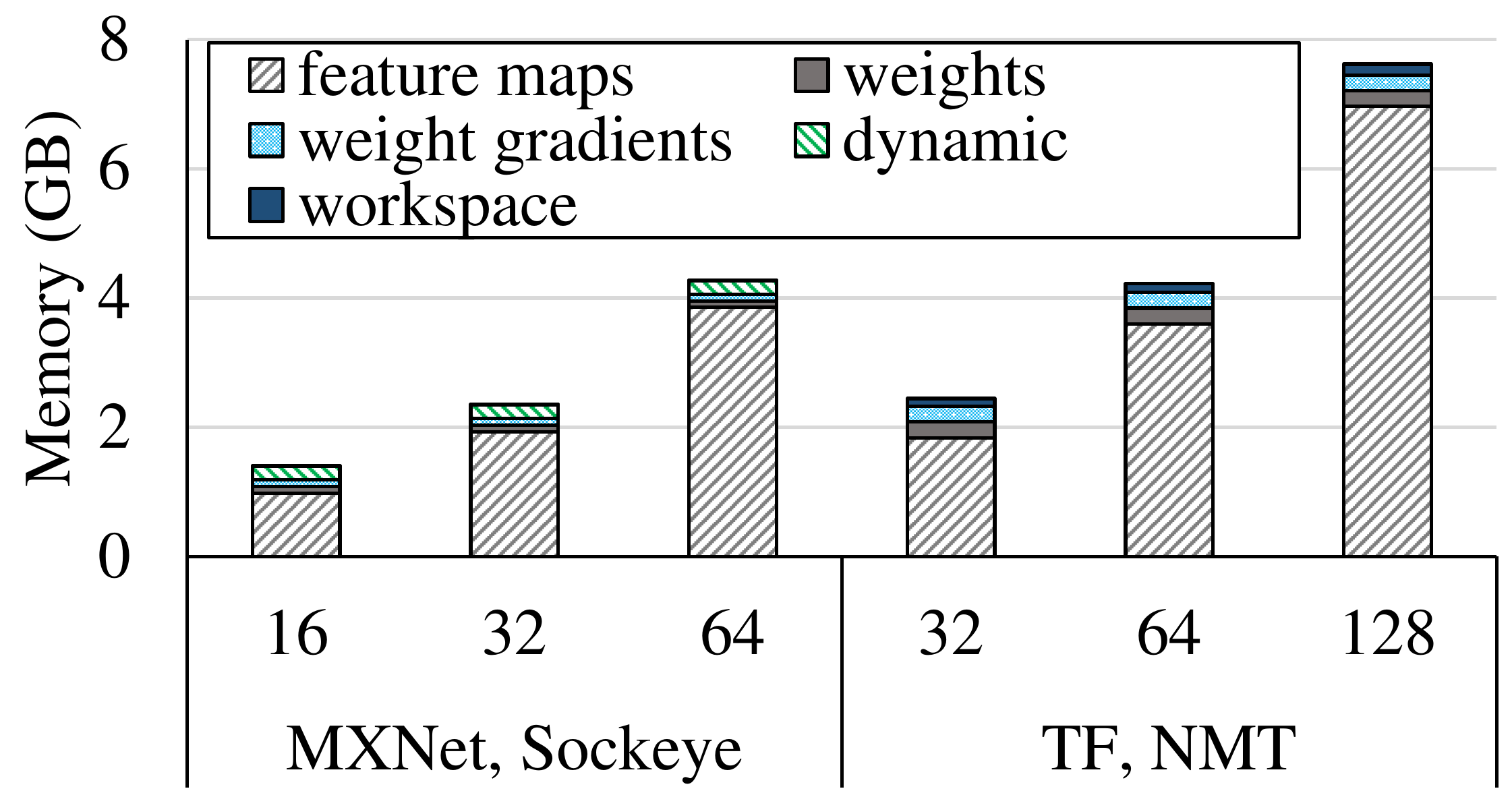}
        \caption{Seq2Seq}
        \label{fig:memory_ds2}
    \end{subfigure}
    \hspace{0.05cm}
    \begin{subfigure}[t]{0.30\textwidth}
        \centering
        \includegraphics[width=\textwidth]{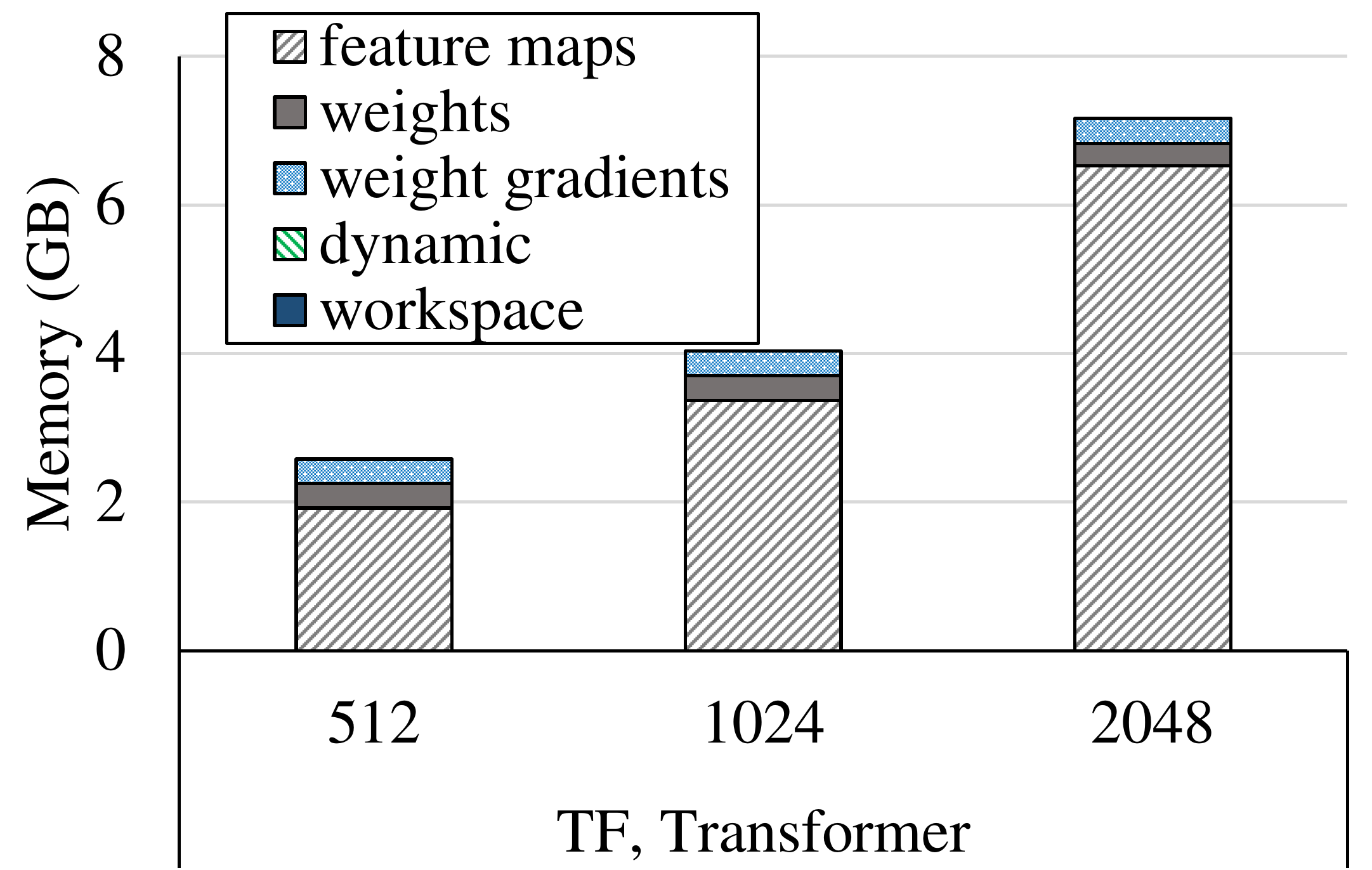}
        \caption{Transformer}
        \label{fig:memory_transformer}
    \end{subfigure}
    \hspace{0.05cm}
    \begin{subfigure}[t]{0.31\textwidth}
        \centering
        \includegraphics[width=\textwidth]{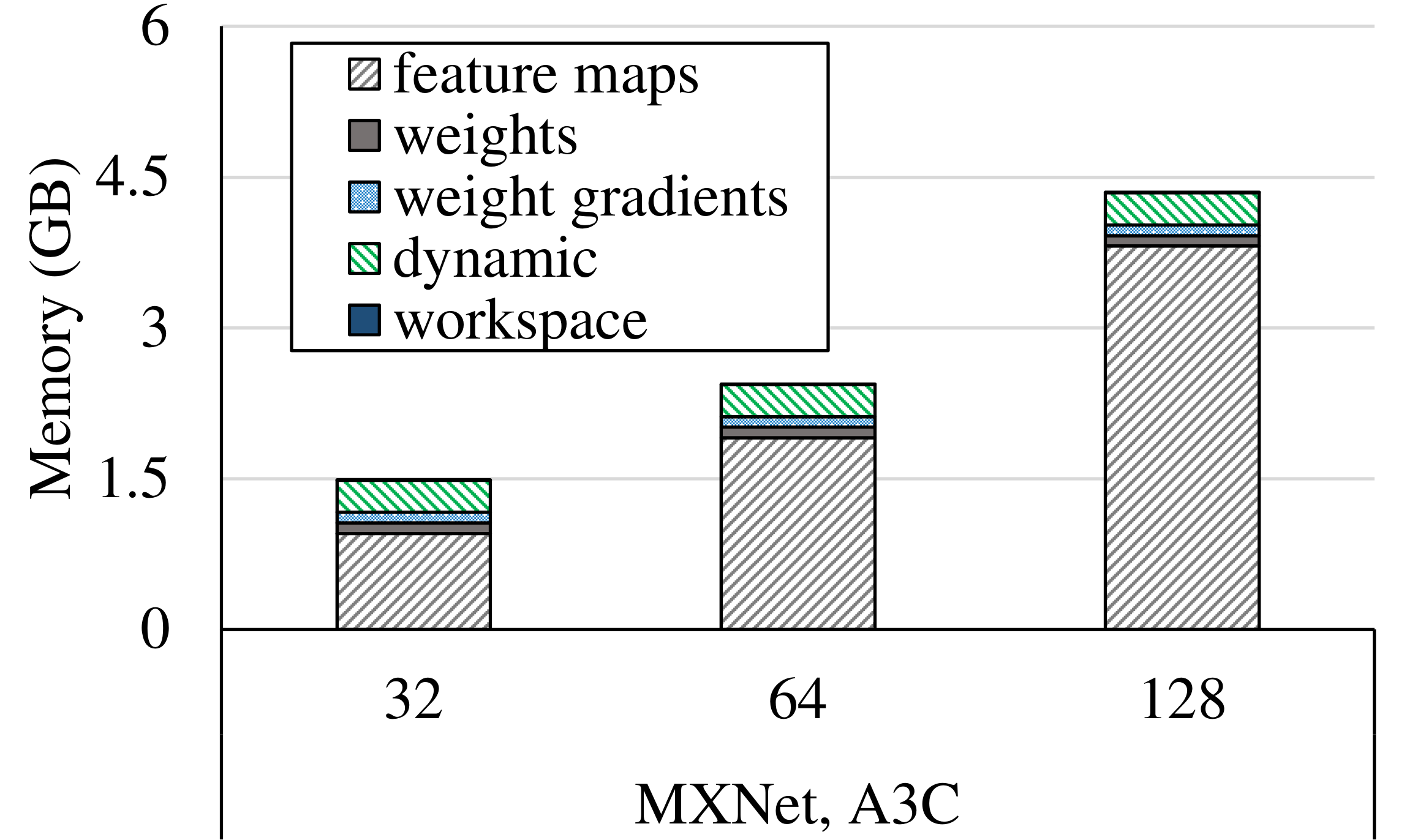}
        \caption{A3C}
        \label{fig:memory_a3c}
    \end{subfigure}
    \caption{GPU memory usage breakdown for different models on multiple mini-batch sizes.}
    \label{fig:memory}
\vspace{-0.1cm}
\end{figure*}

As we have previously shown, the throughput (and hence the performance) of DNN
training can be significantly bottlenecked by the available GPU memory.
Figure~\ref{fig:memory} shows the result of our analysis where the memory is
separated in five categories: weights, gradient weights, feature maps, dynamic,
and workspace. Where appropriate, we vary the size of the mini-batch (shown in
parentheses). The Faster R-CNN model results are similar to image classification models,
but only support one batch size (hence we do not plot them in a separate graph).

\emph{Observation 11: Feature maps are the dominant consumers of memory.} It turns out that \emph{feature maps} (intermediate layer outpus)
are the dominant part of the memory consumption, rather than
weights, which are usually the primary focus of memory optimization for inference.
The total amount of memory
consumed by feature maps ranges from 62\% in \emph{Deep Speech 2} to 89\% in
\emph{ResNet-50} and \emph{Sockeye}. Hence any optimization that wants to
reduce the memory footprint of training should, first of all, focus on feature maps.
This is an interesting observation also because it expands on the results reported in
the only prior work reporting on memory consumption breakdown for DNN training
by Rhu et al.~\cite{rhu2016vdnn}. They look at CNN training only and
find that weights are only responsible for a very small portion of the total memory
footprint. We extend this observation outside of CNNs, but also observe that there are
models (e.g., Deep Speech 2) where weights are equally important.

\emph{Observation 12: Simply exhausting GPU memory with large mini-batch size
might be inefficient.} The memory consumption of feature maps scales
almost linearly with the mini-batch size. From observation 11,
we know that reducing the mini-batch size can dramatically reduce the overall memory consumption needed
for training. Based on observation 1, we also know that the side-effect of
throughput loss while reducing the mini-batch size can be acceptable (for non-RNN models) until
you do not go below saturation point. One can use the
additional GPU memory for larger workspace (can be used for faster implementation of matrix multiplications or convolutions)
and deeper models (e.g., ResNet-102 vs. ResNet-50).
%It is also possible to design
%deeper and more powerful DNNs with more memory available.

%\begin{itemize}
%\item We build a tool for TF, MXNet, cntk to show the memory breakdown.
%\item Memory consumption can be dominated by different parts. (figs: 1 figure of memory breakdown for all benchmarks)
%\item Changing mini-batch size or layer number affect only certain parts. (figs: 1 figure shows mini-batch size sensitivity, 1 figure shows layer sensitivity for ResNet)
%\end{itemize}

\subsection{Multi-GPU and Multi-Machine Training}
\label{sec:performance-distributed}

\begin{figure} [h!]
    \centering
    \includegraphics[width=0.64\textwidth]{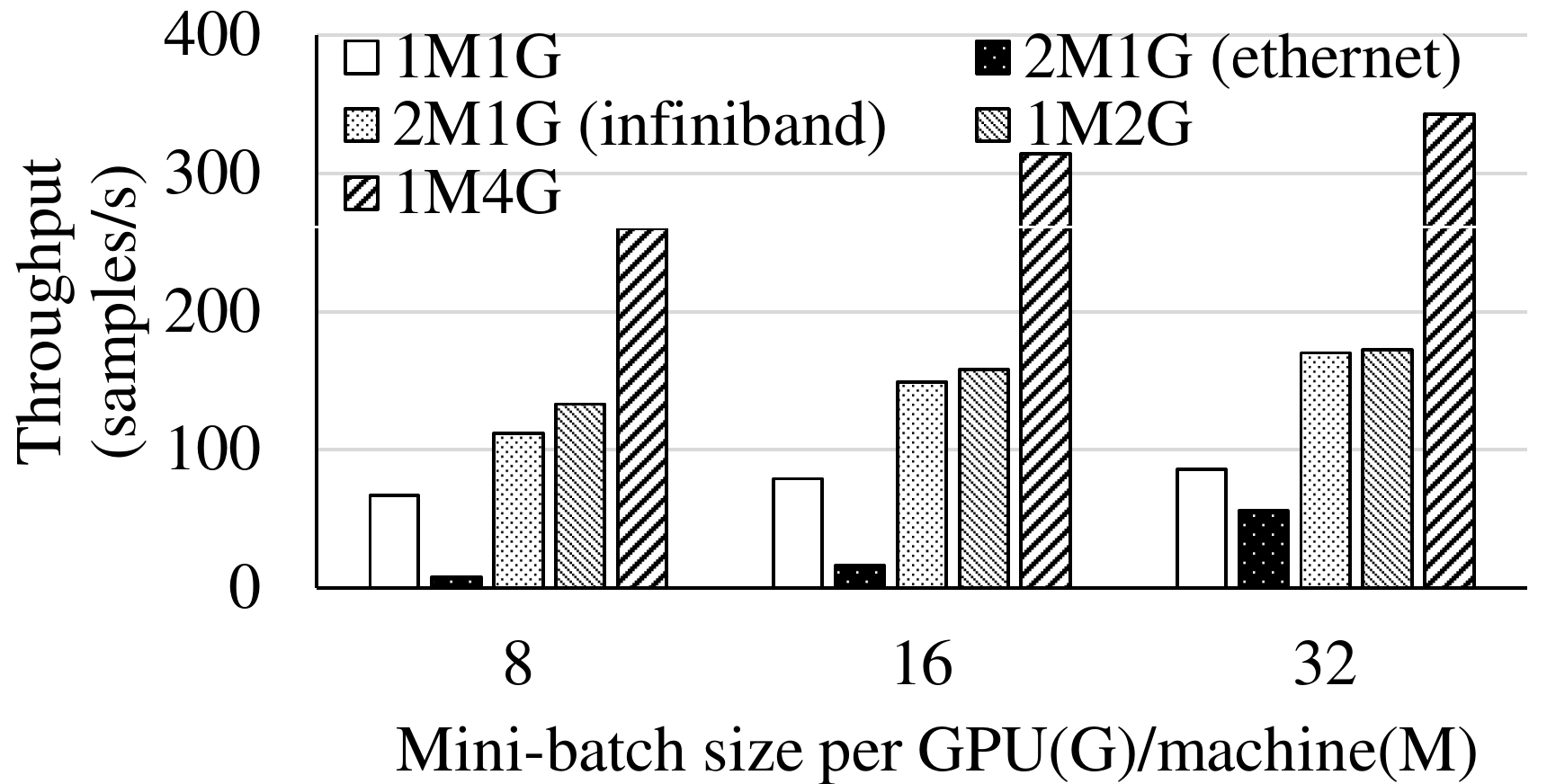}
    \caption{ResNet-50 on MXNet with multiple GPUs/machines.}
    \label{fig:distributed}
    \vspace{-0.3cm}
\end{figure}

Training large DNNs can be done faster when multiple GPUs and/or multiple machines are used. This is
usually achieved by using \emph{data parallelism}, where mini-batches are split between individual
GPUs and the results are then merged, for example, using the \emph{parameter server} approach~\cite{li2014scaling}.
But in order to realize the computational potential of multiple GPUs the comminication channels between them need to have sufficient
bandwidth to exchange proper weight updates. In our work, we decide to analyze the performance scalability
of DNN training using multiple GPUs and multiple machines. We use the \emph{ResNet-50} model on MXNet to perform this analysis.
Figure~\ref{fig:distributed} shows the results of our experiment.
*M stands for the number of machines, and *G for the number of GPUs.

\emph{Observation 13: Network bandwidth must be large enough for good scalability.} We observe
that going from the one machine (\emph{1M1G}) to the two machine (\emph{2M1G
(ethernet)}) coniguration the performance degrades significantly.
This is because DNN training requires constant synchronization
between GPUs in distributed training. Hence faster networking is
required to improve the situation (the 2M1G (infiniband) coniguration has 100Gb/s
IniniBand Mellanox networking). In contrast, DNN training on a single machine with multiple GPUs
(\emph{1M1G, 1M2G, 1M4G}) scales reasonably well as PCIe 3.0 gives enough bandwidth (16 GB/s)s.
In summary, this suggests that networking bandwidth is critical for performance of distributed
training and different techniques (in both software and hardware) should be applied to either
reduce the amount of data sent or increase the available bandwidth.

%\begin{itemize}
%\item Bad scalability comes from two sources: decreasing the mini-batch size, and communication overhead.
%\item To achieve good enough scalability, high communication bandwidth is required.
%\end{itemize}

\section{Related Work} \label{sec:related}

%\subsection{DNN Benchmarks}

There are only a handful of existing open-source DNN benchmark projects, each with a focus that is very different from our work.
ConvNet~\cite{convnet}, CNN-benchmarks~\cite{cnnbenchmark} and Shaohuai et
al.~\cite{shi2016benchmarking} focus exclusively on convolutional network models
mainly for image classification based on the ImageNet data, with the only exception
being one LSTM network in ~\cite{shi2016benchmarking}. In contrast the goal of our work is
a benchmark that covers a wide range of models and applications, beyond just CNNs and image
classification.

%compare the training throughput of various CNNs
%on multiple frameworks, with different underlying libraries and hardware
%environments. Their major focus is on convolutional network models mainly from the winners of
%ImageNet competition (Shaohuai et al.~\cite{shi2016benchmarking} uses
%one LSTM network in their benchmark pool).

DeepBench~\cite{deepbench} is an open-source project from Baidu Research, which is targeted
at a lower level in the deep learning stack than our work: rather than working with
implementations of deep learning models and frameworks, it instead benchmarks
the performance of individual lower level operations (e.g. matrix multiplication) as implemented
in libraries used by various frameworks and directly executed against the underlying hardware.

The Eyeriss project~\cite{eyerissBench} presents
evaluations of a few DNN processors~\cite{chen2017eyeriss,han2016eie} on
hardware metrics for several convolutional networks, but their work is focused
on inference, while ours targets training.

Among existing work, Fathom~\cite{adolf2016fathom} is probably the one
closest to our own, as it also focuses on training and more than a single application
(machine translation, speech recognition and reinforcement learning).
However, their focus is on micro-architectural aspects of execution, breaking down training time into time spent on the various
operation types (e.g. matrix multiplication).
In contrast, our benchmark pool focuses on \emph{system level} aspects of execution such as throughput,
hardware utilization, and memory consumption profiling.
Moreover, Fathom is based on only one framework (TensorFlow), does not consider
distributed training and uses models that are somewhat out-dated by now.

%and compare the performance similarity across
%the models from its benchmark pool, which is a different focus than this work.

%\subsection{DNN Frameworks}

Driven by the achievements of DNN algorithms and the availability of large
datasets, many
frameworks~\cite{abadi2016tensorflow,bergstra2010theano,chen2015mxnet,yu2014introduction,jia2014caffe,tokui2015chainer,collobert2011torch7,chollet2015keras,paszke2017automatic}
were recently proposed for users to easily develop new DNN models for different application
domains. These frameworks provide high-level numpy-like APIs or layer-wise configuration files
to promote programmability. They are also able to make use of modern hardware
accelerators (especially GPUs) to speed up the DNN training process.
In our work, we use three of these frameworks: TensorFlow~\cite{abadi2016tensorflow}, MXNet~\cite{chen2015mxnet},
and CNTK~\cite{yu2014introduction}, but we envision to add more frameworks with more models as we keep
developing our benchmark pool and based on the feedback we expect from both academia and industry.

%\subsection{Profiling Tools}
Due to the complexity of DNN computations, ML developers usually need
additional profiling tools to understand the training performance characteristics. Some
frameworks have their own profiling tools embedded. For example, MXNet allows
users to see the timeline for individual layers. However, this layer-level representation hides
details inside a layer computation, and provide little information about efficiency
of GPU computation. We therefore use the NVIDIA profiling tool called nvprof~\cite{nvprof}, which enables
detailed performance information for each GPU kernel. It also
shows a timeline of both CPU and GPU activities at the function/kernel level.
Another end-to-end profiling tool we use is the Intel VTune~\cite{reinders2005vtune} Amplifier.
It provides detailed analysis of CPU performance (e.g., identifies
the ``hotspots'', where the application spends a lot of time) and also supports the metrics we need.
At the same time, we can not use these tools directly on the full training process due to memory
and time limits, and has to adopt these tools for our need as described in Section~\ref{sec:method}.

%\subsection{Libraries}

To exploit the GPU computation power and to reduce the programming effort,
there are several GPU libraries that provide efficient implementation of basic
operations on vectors and multi-dimension matrices. The most notable examples include
cuDNN~\cite{chetlur2014cudnn} and cuBLAS~\cite{cublas} by NVIDIA,
MKL~\cite{wang2014intel} by Intel, and Eigen~\cite{eigen}. NVIDIA also provides the NCCL library~\cite{nccl}
which implements multi-GPU and multi-node communication primitives to reduce communication
overhead among GPUs. These libraries are widely employed by most mainstream deep
learning frameworks and the three frameworks we use in this work (TensorFlow, MXNet, and CNTK).
These libraries greatly affect the overall training performance, and hence are an important
target of performance analysis and tuning.

%\subsection{GPU Memory Optimization for DNNs}

Multiple techniques are proposed to reduce the memory usage of DNN, including
network pruning~\cite{hanson1989comparing,lecun1990optimal,hassibi1993second,han2015deep,han2015learning},
parameter quantizing or precision
reducing~\cite{han2015deep,anwar2015fixed,esser2015backpropagation,courbariaux2015binaryconnect,rastegari2016xnor}.

%{\bf I am removing the comment on how Rhu's work finds weights only responsible for small portion of total memory
%and the summary of our more differentiated results. This must go to Section 5 results.}

These are extremely useful for inference since they are very effective in
reducing the memory footprint of weights, which enables DNN inference in
mobile environment. Rhu et al.~\cite{rhu2016vdnn} has shown that for the state-of-the-art CNN
training, weights are only responsible for a very small portion of total memory
footprint. We extend this observation outside of CNNs, but also show that there are
models (e.g., Deep Speech 2) where weights are equally important.
Algorithms designed for training DNN with quantized
values~\cite{courbariaux2015binaryconnect,han2015learning,esser2015backpropagation}
suffer from accuracy loss for state-of-the-art models on large datasets. Rhu et
al.~\cite{rhu2016vdnn} develops a new mechanism which uses CPU memory as a
temporary container for \emph{feature maps}, which greatly reduces the memory
footprint of DNN training.

They also developed a tool to show the GPU memory usage for different data structures,
but, unfortunately, this tool is not available outside of NVIDIA.
In this work, we aim to build efficient memory profilers that will be available to a wider
community as we open source them.

\section{Conclusion} 
\label{sec:summary}

In this work, we proposed a new benchmark suite for DNN training, called \texttt{TBD},
that covers a wide range of machine applications from image classification and 
machine translation to reinforcement learning. \texttt{TBD} consists of
eight state-of-the-art DNN models implemented on major deep learning 
frameworks such as TensorFlow, MXNet, and CNTK. We used these models to perform
extensive performance analysis and profiling to shed light on efficiency of 
DNN training for different hardware configurations (single-/multi-GPU and 
multi-machine). We developed a new tool chain for end-to-end analysis of DNN training
that includes (i) piecewise profiling
of specific parts of training using existing performance
analysis tools, and (ii) merging and analyzing the results from these tools using
the domain-specific knowledge of DNN training. Additionally, we built
new memory profiling tools specifically for DNN training for all three
major frameworks. These useful tools can precisely characterize where the memory
consumption (one of the major bottlenecks in training DNNs) goes and 
how much memory is consumed by key data structures (weights, activations, gradients, workspace). 
By using our tools and methodologies, we made several important observations and
recommendations on where the future research and optimization of DNN training should be
focused. We hope that our \texttt{TBD} benchmark suite, tools, methodologies, and
observations will be useful for a large number of ML developers and systems designers in making
their DNN training process efficient.

\bibliographystyle{plain}
\bibliography{references.bib}

\end{document}